# Structural Learning for Template-free Protein Folding

By

Feng Zhao

Submitted to:
**Toyota Technological Institute at Chicago**
6045 S. Kenwood Ave, Chicago, IL, 60637

For the degree of Doctor of Philosophy in Computer Science

Thesis Committee:

Jinbo Xu (Thesis Supervisor)
David McAllester
Jie Liang
Tobin Sosnick

# Structural Learning for Template-free Protein Folding

By

Feng Zhao

Submitted to:
Toyota Technological Institute at Chicago
6045 S. Kenwood Ave, Chicago, IL, 60637

August 2013

For the degree of Doctor of Philosophy in Computer Science

Thesis Committee:

Jinbo Xu (Thesis Supervisor)   Signature: 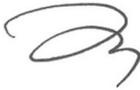   Date: 10/25/2013

David McAllester   Signature: 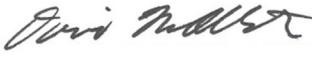   Date: 10/25/2013

Jie Liang   Signature: 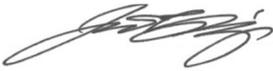   Date: 10/25/2013

Tobin Sosnick   Signature: 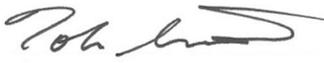   Date: 10/25/2013

# Abstract

The thesis is aimed to solve the template-free protein folding problem by tackling two important components: efficient sampling in vast conformation space, and design of knowledge-based potentials with high accuracy.

We have proposed the first-order and second-order CRF-Sampler to sample structures from the continuous local dihedral angles space by modeling the lower and higher order conditional dependency between neighboring dihedral angles given the primary sequence information. A framework combining the Conditional Random Fields and the energy function is introduced to guide the local conformation sampling using long range constraints with the energy function.

The relationship between the sequence profile and the local dihedral angle distribution is nonlinear. Hence we proposed the CNF-Folder to model this complex relationship by applying a novel machine learning model Conditional Neural Fields which utilizes the structural graphical model with the neural network. CRF-Samplers and CNF-Folder perform very well in CASP8 and CASP9.

Further, a novel pairwise distance statistical potential (EPAD) is designed to capture the dependency of the energy profile on the positions of the interacting amino acids as well as the types of those amino acids, opposing the common assumption that this energy profile depends only on the types of amino acids. EPAD has also been successfully applied in the CASP 10 Free Modeling experiment with CNF-Folder, especially outstanding on some uncommon structured targets.

# Acknowledgements

It was when I started to write this acknowledgement that I realize I owe my gratitude to many people who have made this thesis possible.

My deepest gratitude is to my advisor, Jinbo Xu, whose guidance, support, advice and long discussions have enabled me to gain a profound understanding throughout the PhD studies. It was from him that I have learned the spirit of research, which I believe will be beneficiary to my whole life, and that I will cherish forever.

I would like to thank my committee members David McAllester, Tobin Sosnick and Jie Liang for all their support and suggestions. My heartfelt thanks are also dedicated to all the professors from TTI-C and University of Chicago who have taught me, David McAllester, Nati Srebro, Tamir Hazan, Raquel Urtasun, Julia Chuzhoy, Mattias Blume, Karen Livescu, Greg Shakhnanovich, Anastasios Sidiropoulos, Yury Makarychev, Yang Shen, Laszlo Babai, Tobin Sosnick, Paul J. Sally, and Gregory F. Lawler for their encouraging and sharing their plethora of knowledge on various subjects and courses.

It has been a pleasure to study at TTI-C during the years. My fellow students are cordially thanked for their friendship: Jian Peng, Karthik Sridharan, Andrew Cotter, Zhiyong Wang, Avleen Bijral, Payman Yadollahpour, Jianzhu Ma, Sheng Wang, Joe DeBartolo, Aashish Adhikari, Raman Arora, Qingming Tang, Taewhan Kim, Jian Yao, Hao Tang, Somaye Hashemifar and Behnam Tavakoli Neyshabur. Thank you for your good company and support.

Research reported in this thesis was supported by the National Science Foundation grant DBI-0960390 and National Institutes of Health grant R01GM089753. The content is solely the responsibility of the authors and does not necessarily represent the official views of the NSF or NIH.

This work was made possible by the computing resources and facilities of TeraGrid, Beagle, SHARCNET and Open Science Grid (OSG).

Finally, special thanks to my wife Wentao, my parents and my beloved Ayla. It was because of them that my work becomes meaningful.

# Contents

# List of Figures

# List of Tables

# Chapter 1. Introduction

As shown in Figure 1, the number of solved 3D protein structures in RCSB Protein Data Bank (PDB) increases to 84223 as of Aug. 28th, 2012. On the other hand, the genome sequencing projects have led to the identification of tens of millions of protein sequences publicly available in NCBI Non-redundant database. The high quality annotated and non-redundant protein sequence database SwissProt contains 536789 sequences at Release 2012_07 (Figure 2). The fact that so many protein sequences are part of the public knowledge, and that so many of them have structures that we have not yet solved, means that we have a long way to go to understand the way that proteins work and function within the body. Many computational methods have been developed to predict the structure of a protein from its primary sequence, based on the famous Anfinsen dogma that all of the information necessary for a protein to fold to the native structure resides in the protein's amino acid sequence(Anfinsen, 1973). These methods can be roughly classified into two categories: template-based and template-free modeling.

Despite considerable progress taken place over the past decade, template free modeling remains one of the unsolved mysteries in the field of computational structural biology. The primary challenges still remain in two areas: the vast conformation space to be searched and limited accuracy of the current energy functions designed. The purpose of this thesis is to explore these topics in two sections: first, protein conformation sampling, that is, the exploration of conformational space that corresponds with a particular protein sequence; second, Design of energy function, that is, an accurate physics-based or knowledge-based potential to quantify interactions among residues or atoms.

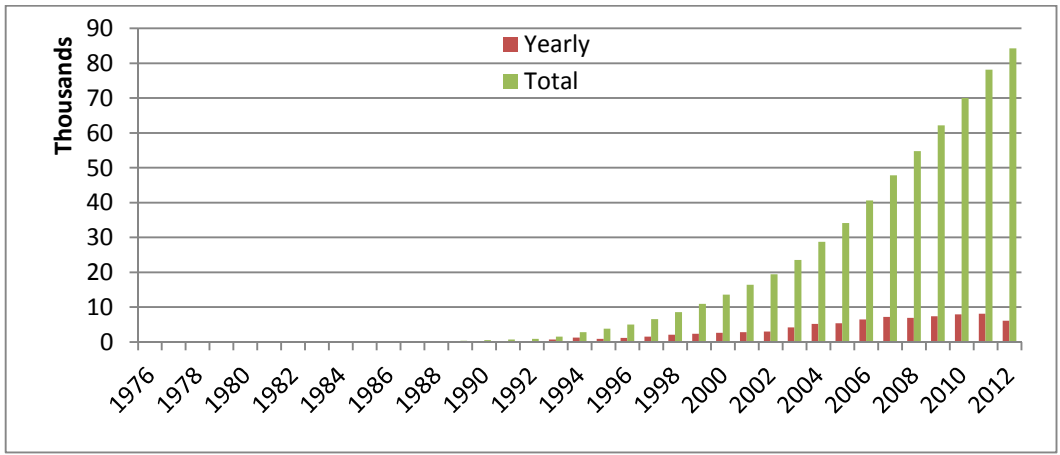

Figure 1 Yearly growth of total structures in RCSB Protein Data Bank (PDB).

Adapted from http://www.rcsb.org/pdb/statistics/contentGrowthChart.do?content=total

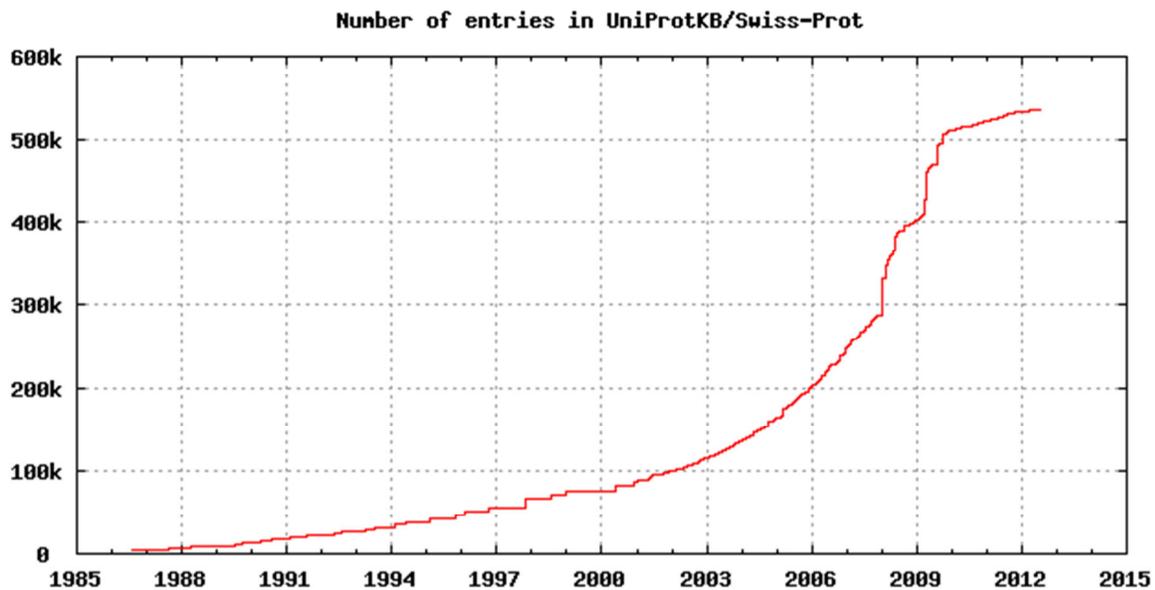

Figure 2 The growth of the SwissProt database.

Adapted from http://web.expasy.org/docs/relnotes/relstat.html

## 1.1 Conformation Sampling

Template-free modeling that comes from fragment assembly (Bowie and Eisenberg, 1994; Claessens et al., 1989; Jones and Thirup, 1986; Levitt, 1992; Simon et al., 1991; Sippl, 1993; Unger et al., 1989) and lattice-models (Kihara et al., 2001; Wendoloski and Salemme, 1992; Xia et al., 2000; Zhang et al., 2003) have been studied extensively. These two common methods and their combined use in template-free modeling have brought impressive results in CASP (Critical Assessment of Structure Prediction) competitions (Moult, 2005; Moult et al., 2007; Moult et al., 2005; Moult et al., 2003). The popular fragment assembly program Robetta (Misura et al., 2006; Simons et al., 1997) has the best track records in all of the template-free modeling programs. Both TASSER (Zhang and Skolnick, 2005a) and its derivative Zhang-Server (Wu et al., 2007a) have garnered recognition for stellar results in both CASP7 and CASP8 through the combination of threading-generated fragments, distance restraints, and lattice modeling.

Fragment based protein structure prediction takes place in two distinct steps. First, it is necessary to cut a protein sequence into minuscule segments and then pick out literally scores of possible fragments that might construct each segment. Then, one uses a simulation or search algorithm to build the protein structure. While it is important to note that the results attained by existing template-free modeling methods are exciting, there are several areas of concern that have not yet been resolved. One of these has to do with the small pool of proteins with solved experimental structures within the PDB. This makes it almost impossible to put together a collection of fragments that can match the number of ways a protein can locally conform – particularly in the loop regions, where the possibilities grow. It is true that a new fold may entirely consist of rarely occurring super-secondary structure motif that is found nowhere else in the protein data bank. Second, because the conformation space that a lattice model or fragment library defines is by nature discrete, it may even keep the original fold from being searched since a slight change in backbone angles, even in the most minuscule degree, can result in a totally different fold.

It is possible to make the conformational space continuous, as Bystroff et al. have shown with HMMSTR (Bystroff et al., 2000), which is a Hidden Markov Model (HMM) model trained from a fragment library, to generate local conformations for a given sequence. Both Colubri et al. (Colubri et al., 2006) and Gong et al. (Gong et al., 2005) have have analyzed the skeletal structure of protein, to see if that structure can be reconstructed – under the assumption that the angles are limited to the degrees allowed in their original Ramachandran basins ( a native Ramachandran basin is the region that holds the values of the native backbone (φ, ψ) torsion angle pairs). Sosnick and coworkers (Colubri et al., 2006) divided the torsion angle space into six distinct sections (see Figure 3) while Gong et al. (Gong et al., 2005) split the space into 36 distinct regions (see Figure 4). The upshot from both pieces of research is that it is possible to reconstruct the backbone structure of many of the smaller proteins with good accuracy, if the angles are limited to those native basins. These studies do not consist of pure ab initio folding; instead, they show that if researchers can come up with a reasonably close guess as to each angle's native basin, then it should also be possible to forecast with accuracy a protein's backbone structure.

Fragment-HMM (Li et al., 2008), which is a variant of Robetta, has the ability to sample conformations within a continuous space. However, because Hidden Markov model is constructed from 9-mer fragments, it still has the issue of coverage that plagues many of the other analytical machines. The TOUCHSTONE programs (Kihara et al., 2001; Zhang et al., 2003) , which uses the lattice model, does not have the coverage issue. but its sampling of protein conformations comes from a three-dimensional lattice that has only finite resolution. More importantly, these conformations that get sampled might not even have a local structure that resembles those of native proteins, because TOUCHSTONE does not sample a conformation based upon the primary sequence of a protein. Rather, these applications take several statistical potentials over short range in the energy function to manage the creation of a local structure that closely resembles the desired protein(s).

It is worth mentioning that some other methods exist for trying to sample conformations of protein in a continuous space using the probability. For each conformation, the probability is an approximation to its overall stability. Sequence information is used to estimate those probabilities. In response, Feldman and Hogue developed the FOLDTRAJ (Feldman and Hogue, 2002) program, which implements a probabilistic all-atom conformation sampling algorithm. FOLDTRAJ is tested on three small proteins 1VII, 1ENH, and 1PMC, with 100,000 decoys for each protein. The best models achieve 3.95, 5.12 and 5.95 Angstroms in term of RMSD from the native structures. However, because FOLDTRAJ does not use the nearest neighbor effects or the sequence profile in modeling the relationship between sequence and structure, it is not able to generate models that match the quality of the models from the popular fragment assembly Rosetta application.

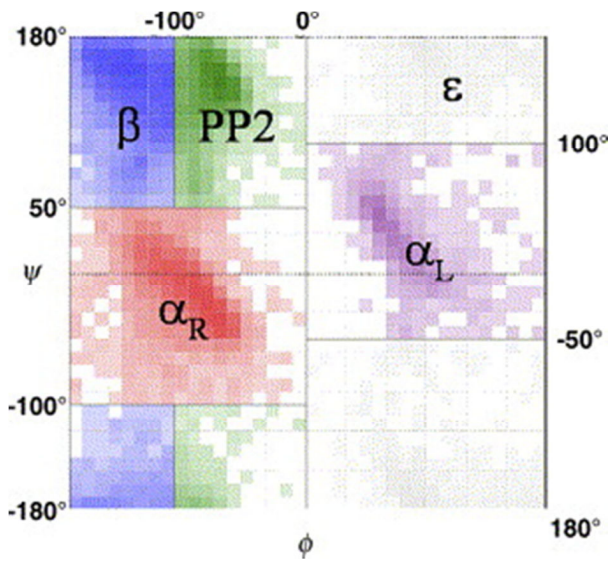

Figure 3 The 6 Ramanchandran basins: β (blue), poly-proline II, PPII (green), αR (red), αL (magenta) and ε (grey). Color intensity reflects the (φ, ψ) occupancy. Adapted from (Colubri et al., 2006)

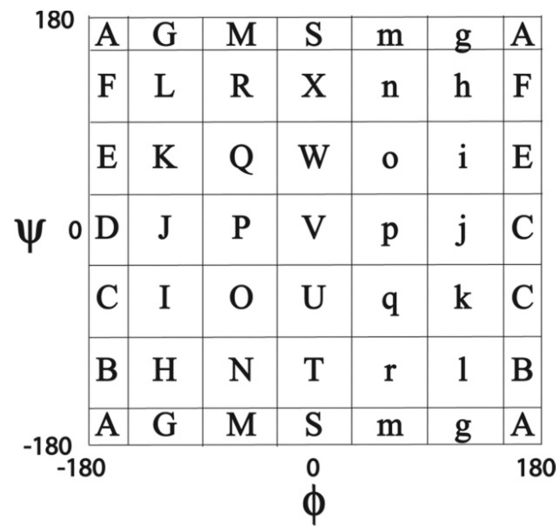

Figure 4 Backbone (φ, ψ) -space subdivided into 36 alphabetically labeled, 60° × 60° grid regions. Adapted from (Gong et al., 2005)

Hamelryck et.al have developed two different HMM models FB5-HMM (Hamelryck et al., 2006) and TorusDBN (Boomsma et al., 2008) to sample protein conformations from a continuous space. These models not only capture the relationship between backbone angles and primary sequence, but also consider the angle-dependency between two adjacent residues. FB5-HMM uses a Hidden Markov Model to discover the native basins of the local $(\theta, \tau)$ pseudo backbone dihedral angles of a protein sequence by learning the local sequence-structure relationship; and TorusDBN learns the local $(\phi, \psi)$ backbone dihedral angle dependency using a dynamic Bayesian network, which is a generalization of the Hidden Markov Model. There are two particular advantages of these methods: first, they model the backbone angles using a directional statistics distribution (Kent, 1982; Mardia et al., 2007; Singh et al., 2002) so that they can be sampled from a continuous space; second, they can find how two adjacent backbone angles are dependent on one another. The experiments shows that by modeling the dependency between two adjacent positions, FB5-HMM can generate more conformations that are close to the native structure than the applications that do not model the correlation between neighboring residue positions.

There are quite a few fragment assembly methods (Kent, 1982; Mardia et al., 2007; Singh et al., 2002; Tuffery and Derreumaux, 2005) that also exploit the dependency between the two adjacent fragments. As a result, researchers have learned that this dependency is one of the most important things of conformation sampling. They also demonstrated that Torus-DBN can generate local conformations as accurately as the fragment assembly method (Boomsma et al., 2008). However, it is important to note that these models only consider angle-dependency among two neighboring residues. Because of this restriction, these models cannot make use of more enriched forms of the sequence information, such as PSI-BLAST sequence profile or the threading-generated constraints. Hence their sampling efficiency is quite limited. Furthermore, these models have not been considered in the context of template-free modeling in the real world yet.

Due to the expressivity of the HMM model, FB5-HMM assumed that each residue is independent of its secondary structure type, and each backbone angle only depends on its corresponding residue (monomer, or 1-mer) and secondary structure at the same position, though it actually depends on at least three neighborhood residues (3-mer). Zhao et al (2008, 2009) implemented an extensible protein conformation sampler, CRF-Sampler (Zhao et al., 2008; Zhao et al., 2009), based on a probabilistic graphical model Conditional Random Fields. CRF-Sampler incorporates the dependence among up to 9 neighboring residues and sequence profile information with continuous distribution into the backbone angle model.

The application CRF-Sampler is a protein conformation sampler that is extensible and has been built using a discriminative learning method by modeling the dependence of the unobserved variables upon observed variables in the forms of the conditional probability distribution

P(backbone angle | protein sequence) to predicting (θ, τ) from the sequence information. CRF-Sampler can learn over 10,000 different parameters that quantify relationships among the backbone angles, the primary sequence, and the secondary structure. By using just the self-avoiding constraints and compactness, CRF-Sampler can simulate conformations from the existing primary sequence and the predicted secondary structure. CRF-Sampler also exhibits high flexibility in that a variety of model topologies and feature sets can be defined to model the sequence-structure relationship without worrying about parameter estimation. Experimental results show that the first-order CRF-Sampler using a small collection of features can generate decoys with much higher quality than the FB5-HMM model.

The second-order Conditional Random Fields (CRFs) model has the capability of portraying more complicated levels of dependency among the dihedral angles in the local sequences. When coupled with a simple energy function, this probabilistic method compares favorably with the fragment assembly method in the blind CASP8 evaluation, especially on alpha or small beta proteins. When the second-order CRF-Sampler is combined with a simple energy function with 3 terms including a distance potential, a hydrogen potential and a hydrophobic potential, it generates superior outcomes that compare favorably with the fragment assembly method in the blind CASP8 evaluation, especially on alpha or small beta proteins. This is the first probabilistic method that can search conformations in a continuous space and achieves favorable performance. In fact, this is the very first method with the ability to search within a continuous conformation space and also generate favorable decoys. In addition, this method also created three dimensional models that were more accurate than template-based methods for several of the hard targets in CASP8. The second-order CRF-Sampler can also be applied to protein loop modeling, model refinement, and even RNA tertiary structure prediction.

The second-order CRF-Sampler is used in RAPTOR++ in CASP8 to predict the 3D structures of the hard targets. RAPTOR++ has four different modules: threading, model quality assessment, multiple protein alignment and template-free modeling, i.e. the second-order CRF-Sampler. RAPTOR++ first tries to connect a target protein to the templates using the first three modules. Next, it predicts the quality of the 3D model implied by each alignment using a model quality assessment method. If all the alignments come back with insufficient quality, RAPTOR++ uses the second-order CRF-Sampler. CRF-Sampler is also used to refine template-based models.

The CRF models describe the relationships among input features and outputs by using linear potential functions. The difficulty emerges because this relationship is frequently nonlinear, complex beyond the capacity of the model. To take advantage of both the structured graphical models and non-linear classifiers such as SVM and neural networks, Zhao et al. proposed CNF-Folder, a fragment-free approach to protein folding based on Conditional Neural Fields (CNF) (Peng et al., 2009; Zhao et al., 2010). CNF-Folder extends CRF-Sampler by adding a middle layer between input features and outputs. The middle layer consists of a series of hidden gates, each with its own parameter, and each acts as a local neuron (i.e. feature extractor) to capture the non-linear relationship between input features and outputs. Based on experimental results, CNF-Folder can create superior decoys on a variety of test proteins including the CASP8 free-modeling targets when it is used with replica exchange simulation and same energy function as used by CRF-Sampler. The most impressive results that CNF-Folder can forecast include the correct fold for T0496_D1, which is one of the two CASP8 targets that have a truly new fold. For T0496, the predicted model is also considerably superior to all of the CASP8 models.

## 1.2 Energy function

To express the function and structure of a protein, one often needs the ability to give a quantifiable value to evaluate the interactions among residues or atoms. This quantifiable value is called potential which is created either based on physics rules or extracted from knowledge base. There are studies (Bradley et al., 2005; Skolnick, 2006) indicating that knowledge-based statistical potentials (Li and Liang, 2007; Lu et al., 2008; Miyazawa and Jernigan, 1985; Shen and Sali, 2006; Simons et al., 1997; Sippl, 1990; Tanaka and Scheraga, 1976; Zhang and Zhang, 2010; Zhou and Zhou, 2002) compare favorably to physics-based potentials (Brooks et al., 2009; Bryngelson et al., 1995; Case et al., 2005; Dill, 1985, 1997; Dobson et al., 1998; Schuler et al., 2001; Shakhnovich, 2006) in many applications including ab initio folding (Jones and Thirup, 1986; Kachalo et al., 2006; Kihara et al., 2001; Levitt, 1992; Simons et al., 1997; Wu et al., 2007a; Zhao et al., 2010), docking (Zhang et al., 1997), binding (Kortemme and Baker, 2002; Laurie and Jackson, 2005), mutation study (Gilis and Rooman, 1996, 1997), decoy ranking (Bauer and Beyer, 1994; Casari and Sippl, 1992; Gatchell et al., 2000; Hendlich et al., 1990; Samudrala and Moult, 1998; Simons et al., 1999; Vendruscolo et al., 2000) and protein model quality assessment (Jones and Thornton, 1996; Panchenko et al., 2000; Peng and Xu, 2010; Reva et al., 1997; Sippl, 1993). Knowledge-based statistical potentials extract interactions from the solved protein structures in the Protein Data Bank (PDB) (Berman et al., 2000). They are more user-friendly than physics-based potentials in terms of user API and calculation complexity. Many Knowledge-based statistical potentials have been created, including the popular DOPE (Shen and Sali, 2006) and DFIRE (Zhou and Zhou, 2002). Some statistical potential quantify local atomic interactions (e.g., torsion angle potential) while others capture non-local atomic interactions (e.g., distance-dependent potential).

Even after extensive study though, designing protein potential with significant precision remains highly challenging. A lot of knowledge-based statistical potentials are derived from the inverse of the Boltzmann law. They have two primary components: reference state and observed atomic interacting probability. The observed atomic interacting probability is usually assumed to correlate with only atom types, and is estimated through a similar simple counting method. The reference state can be derived theoretically based on physical rules or be estimated statistically from the empirical dataset. Many potentials are distinguished in the construction of the reference state.

Zhao et al. (2012) designed the statistical potential EPAD (Evolutionary PAirwise Distance potential) by making use of protein evolutionary information, which has not been used by currently popular statistical potentials (e.g., DOPE and DFIRE). EPAD is unique in that is has different energy profiles for the same type of atom pairs, depending on their sequence positions. Experiments confirm that this position-specific statistical potential EPAD outperforms several popular statistical potentials in both decoy discrimination and ab initio folding. Overall, the results suggest several implications: 1. statistical potentials that are protein-specific and position-specific work more effectively; 2. evolutionary information makes energy potentials perform more effectively; 3. observed probability and reference state both make energy potentials different; 4. context-specific information improves the estimation of observed probability.

## 1.3 Contribution

The thesis has several contributions to science discovery as following.

First, in protein folding, the local dihedral angles are from a continuous space. A first-order CRF-Sampler is proposed to model the conditional dependency between two neighboring dihedral angles given the primary sequence information.

Second, the 3D structure of protein depends on higher order relationships among the constructing amino-acids (residues) in every neighborhood, as well as the long range interactions among the residues at the positions that are far from each other in the primary sequence. A framework combining the second-order Conditional Random Fields and the energy function is introduced to guide the local conformation sampling using long range constraints with the energy function.

Third, it is a nonlinear relationship between the sequence profile and the local dihedral angle distribution. A novel machine learning model Conditional Neural Fields which utilizes the structural graphical model with the neural network is applied to model this complex relationship in CNF-Folder, a powerful fragment-free protein conformation sampler.

Fourth, the energy profiles of the pairwise distance potential in the proteins depend on the positions of the interacting amino acids in the primary sequence as well as the types of those amino acids opposing the common assumption that this energy profile depends only on the types of amino acids. A new probabilistic neural network model is proposed to estimate the observed probability of the pairwise distances, and a novel statistical distance potential (EPAD) is designed for protein structure evaluation and functional studies.

The demand for using these tools already exists. We have developed software packages of the CNF-Folder and EPAD for the users to download.

## 1.4 Organization

This thesis is essentially a collection of original published papers in the area of template-free protein folding and ranking. Each chapter is written in such way that it can be read independently.

Chapter 2 introduces the basic concepts on continuous representation of protein conformation. The notations are used in the subsequent chapters.

Chapter 3 presents the first order Conditional Random Fields model for continuous protein conformation sampling. With the help of the continuous representations defined in Chapter 2, Chapter 3 presents the CRF model to learn the relationship between the sequence information and the local backbone dihedral angles. An efficient forward back-track conformation sampling algorithm is also presented. The performance of the first order CRF-Sampler is evaluated against several protein simulation approaches including the most popular Rosetta method.

Chapter 4 describes the higher dependency among the dihedral angles in the local sequences, as well as the importance of long range interactions in protein folding. Hence we proposes a framework combining the second order Conditional Random Fields to model the local interactions with the energy function to add long range constraints in the simulated annealing sampling algorithm. The second order CRF-Sampler is built upon this framework, and applied in CASP8.

Chapter 5 introduces the nonlinear relationship between the sequence information and the local conformations. The Conditional Neural Fields model is applied to learn this complex relationship and explained in detail in this chapter. Moreover, a different simulation method parallel tempering is implemented in the CNF-Folder application to give out more high quality decoys in comparison to CRF-Samplers as well as other popular free-modeling methods.

Chapter 6 briefly introduces the statistical potentials and then presents in detail the novel Evolutionary PAirwise Distance potential (EPAD). A complete comparison of EPAD against other knowledge-based distance potentials on all the previously used datasets is provided, which demonstrates the superior performance of EPAD. A new large scale dataset is also proposed to avoid the statistical potentials' over-tuning over the small datasets.

In Appendix A, we have proposed a contact capacity potential (CCP). The experimental results imply that the contact capacity potential helps significantly improve decoy discrimination when combined with distance-dependent potentials. The results also support the hypothesis that the wrapping of each residue by surrounding amino acids is guided by maximizing the local static electric field, so that the core of the residue is protected against the water molecules.

# Part 1.

# Probabilistic and Continuous Models of Protein Conformational Space for Fragment-free and Template-Free Modeling

# Chapter 2. Continuous representation of protein conformations

Evaluating a full-atom energy function takes a great deal of time. However, a residue-level energy function simply lacks the accuracy of an atom-level energy function. In this thesis, we use a continuous yet simplified representation of a protein model. We only consider the main chain of C-beta atoms in the folding simulations discussed in the subsequent chapters.

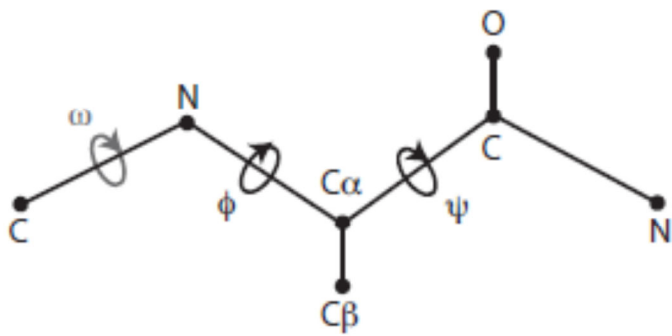

Figure 5 The $\phi, \psi, \omega$ dihedral angles in one residue of the protein backbone. $\omega$ can be assumed to be fixed at 180° (trans) or 0° (cis).

## 2.1 C$_\alpha$-trace representation

A protein backbone conformation can be described by angle triples ($\phi, \psi, \omega$), as well as a set of bond lengths. $\phi$ is the dihedral angle around $N-C_\alpha$ bond and $\psi$ is the dihedral angle around $C_\alpha-C$ bond. Because it is possible to approximate $\omega$ and the bond lengths as constants, we can represent a protein backbone as a set of ($\phi, \psi$) angles. With the exception of both the two terminal residues of a protein chain, each residue has a pair of $\phi$ and $\psi$ angles. These ($\phi, \psi$) angles give us what we need to calculate the coordinates for all the nonhydrogen atoms of a protein backbone. However, for some proteins, even if we have all of their native $\varphi$ and $\psi$ angles, we cannot accurately rebuild their backbone conformations because of slight variation of other angles.

This article utilizes a different way to represent a protein backbone instead of the $(\phi, \psi)$ representation. Because the virtual bond length between two adjacent $C_\alpha$ atoms can be approximated as a constant, i.e., 3.8 Å, the protein backbones can be represented as a set of pseudo angles $(\theta, \tau)$ (Levitt, 1976). One rare exception is when the second residue is *cis* proline, the virtual bond length is approximately 3.2 Å. In this representation, all the other atoms are ignored except the $C_\alpha$ atoms. For any position *i* in a backbone, $\theta$ is defined as the pseudo-bond angle formed by the $C_\alpha$ atoms at positions *i*–1, *i* and *i*+1; $\tau$ is a pseudo dihedral angle formed by the $C_\alpha$ atoms at positions *i*–2, *i*–1 *i* and *i*+1. Given the coordinates of the $C_\alpha$ atoms at positions *i*–2, *i*–1, and *i*, the coordinates of the $C_\alpha$ atom at position *i*+1 can be calculated from $(\theta, \tau)$ at position *i*.

Therefore, for a protein with N amino acid residues, given the positions of the first three $C_\alpha$ and N–2 $(\theta, \tau)$ pairs, we can build the $C_\alpha$ trace of a protein. The relative positions of the first three $C_\alpha$ atoms are determined by the $\theta$ angle of the second residue. Using $(\theta, \tau)$ representation, only the coordinates of the $C_\alpha$ atoms can be recovered. The coordinates of other backbone atoms and $C_\beta$ atom can be built using programs such as MaxSprout (Holm and Sander, 1991), BBQ (Gront et al., 2007), and SABBAC (Maupetit et al., 2006), which can build backbone positions with RMSD less than 0.5 Å.

## 2.2 Distribution of bond angles

The preferred conformations of an amino acid in the protein backbone can be described as a probabilistic distribution of the $(\theta, \tau)$ angle pair. Each $(\theta, \tau)$ corresponds to a unit vector in the three-dimensional space (i.e., a point on a unit sphere surface). We can use the 5-parameter Fisher-Bingham (FB5) distribution (Hamelryck et al., 2006; Kent, 1982) to model the probability distributions over unit vectors. FB5 is the analogue on the unit sphere of the bivariate normal distribution with an unconstrained covariance matrix. The probability density function of the FB5 distribution is given by

$$f(u) = \frac{1}{c(\kappa, \beta)} \exp\left(\kappa \gamma_1 . u + \beta((\gamma_2 . u)^2 - (\gamma_3 . u)^2)\right)$$

where *u* is a unit vector variable and *c*(κ,β) is a normalizing constant (Kent, 1982). The parameters $\kappa (> 0)$ and $\beta (0 < 2\beta \leq \kappa)$ determine the concentration of the distribution and the ellipticity of the contours of equal probability, respectively. The higher the $\kappa$ and $\beta$ parameters, the more concentrated and elliptical the distribution is, respectively. The three vectors $\gamma_1$, $\gamma_2$, and $\gamma_3$ are the mean direction, the major and minor axes, respectively. The latter two vectors determine the orientation of the equal probability contours on the sphere, while the first vector determines the common center of the contours.

We cluster the whole space of ($θ$, $τ$) into 100 groups, each of which can be described by an FB5 distribution. We calculate the ($θ$, $τ$) distribution for each group from a set of ~3000 non-redundant proteins with high-resolution x-ray structures using KentEstimator (Hamelryck et al., 2006). More detailed description of how to calculate the FB5 distributions is shown in Chapter 3 (Zhao et al., 2008). Once we know the distribution of ($θ$, $τ$) at one residue, we can sample a pair of real-valued ($θ$, $τ$) angles in a probabilistic way and thus, explore protein conformations in a continuous space.

## 2.3 Building backbone atoms

Using ($θ$, $τ$) representation, only the coordinates of the C$_α$ atoms can be built. To use an atom-level energy function, we also need to build the coordinates of other atoms. Given a C$_α$ trace, there are many methods that can build the coordinates for the main chain and C$_β$ atoms (Gront et al., 2007; Holm and Sander, 1991; Maupetit et al., 2006). To save computing time, we want a method that is both accurate and efficient. We choose to use a method similar to BBQ (Gront et al., 2007). The original BBQ method can only build coordinates for the backbone N, C, and O atoms. We extend the method to build coordinates for the C$_β$ atom. Experimental results (data not shown) indicate that RMSD of this method is approximately 0.5 Å supposing the native C$_α$-trace is available. This level of accuracy is good enough for our folding simulation.

To employ the KMB hydrogen-bonding energy (Morozov et al., 2004) for $β$-containing proteins, we also need to build the backbone hydrogen atoms. We use a quick and dirty method to build coordinates for the hydrogen atom HN (Tooze and Branden, 1999)). Let $N_i$ denote the position of the main chain N atom in the same residue as the HN atom. Let $N_i C_{i-1}$ denote the normalized bond vector from the N atom to the C atom in the previous residue. Let $N_i C_α$ denote the normalized bond vector from the N atom to the C$_α$ atom in the same residue. Then the position of the hydrogen atom HN can be estimated by $N_i - \frac{N_i C_{i-1} + N_i C_α}{|N_i C_{i-1} + N_i C_α|}$. The average RMSD of this method is approximately 0.2Å (data not shown) supposing the native coordinates of other main chain atoms are available.

## 2.4 Mathematical Symbols

Unless specifically clarified, we will use the mathematical symbols for the rest chapters in in Part 1 as listed in Table 1. In the context of the Conditional Random Fields (CRFs) and Conditional Neural Fields (CNFs) models we use, the primary sequence (or sequence profile) and predicted secondary structure are viewed as observations; the backbone angles and their FB5 distributions are treated as hidden states or labels.

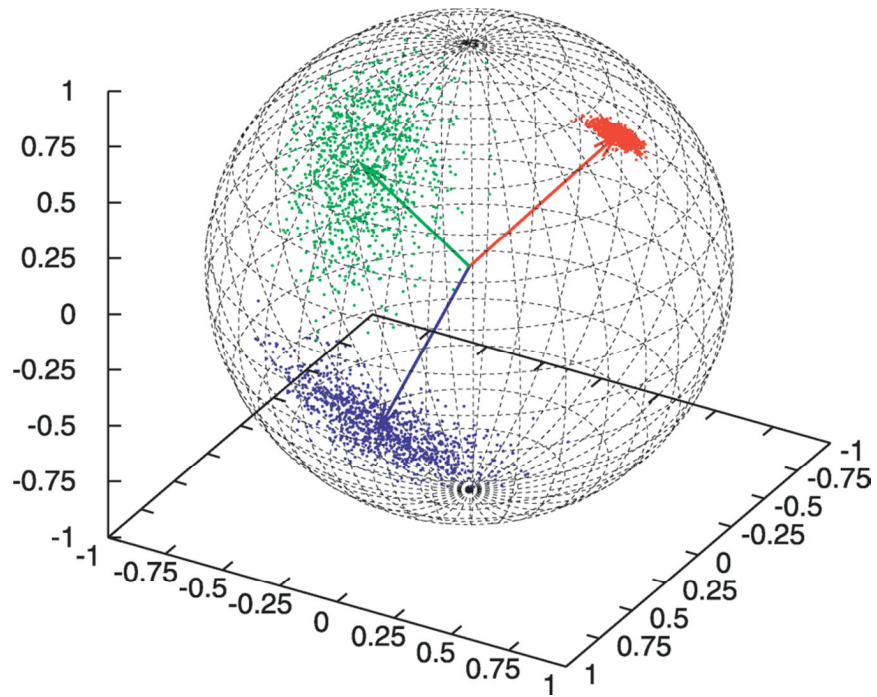

Figure 6 Three point sets sampled from the FB5 distribution on the unit sphere.

The three node values are typical representatives of coil (blue), α-helix (red), and β-strand (green). The arrows points to the mean directions of the three point sets. Adapted from (Hamelryck et al., 2006).

Table 1 Some Mathematical Symbols Used in the CRF Models

| Symbols | Annotations |
|---|---|
| $X$ | The PSIPRED-predicted secondary structure likelihood scores. A matrix with $3 \times N$ elements where $N$ is the number of residues in a protein. |
| $X_i$ | The predicted likelihood of three secondary structure types at position $i$. It is a vector of three values, indicating the likelihood of helix, beta and loop, respectively. |
| $X_i(x)$ | The predicted likelihood of secondary structure type $x$ at position $i$. |
| $M$ | The position-specific frequency matrix with $20 \times N$ entries, each being the occurring frequency of one amino acid at a given position. |
| $M_i$ | A vector of 20 elements, denoting the occurring frequency of 20 amino acids at position $i$. |
| $M_i(aa)$ | The occurring frequency of amino acid $aa$ at position $i$. |
| $H$ | $H = \{h_1, h_2, \ldots, h_{100}\}$, the set of 100 backbone angle states, each representing an FB5 distribution. |
| $\Lambda$ | $\Lambda = \{\lambda_1, \lambda_2, \ldots, \lambda_p\}$, the set of model parameters used by CRF/CNF model. |

# Chapter 3. First-order CRF-Sampler

## 3.1 Introduction

The FB5-HMM and TorusDBN models developed by Hamelryck et.al (Boomsma et al., 2008; Hamelryck et al., 2006) sample the protein conformations from a continuous space. These models are able to capture the relationship between backbone angles and primary sequence, and the angle-dependency between two adjacent residues. FB5-HMM uses a Hidden Markov Model to discover the native basins of the local (θ, τ) pseudo backbone dihedral angles of a protein sequence by learning the local sequence-structure relationship; and TorusDBN learns the local (ϕ, ψ) backbone dihedral angle dependency using a dynamic Bayesian network, which is a generalization of the Hidden Markov Model. There are two particular advantages of these methods: first, they model the backbone angles using a directional statistics distribution (Kent, 1982; Mardia et al., 2007; Singh et al., 2002) so that they can be sampled from a continuous space; second, they can find how two adjacent backbone angles are dependent on one another.

Although experimental result shows that the sampler based on Hidden Markov Model(HMM) method is promising in generating decoys with good quality, the model has a couple of assumptions that may be relaxed.

The **1st assumption** is that the residue at any position is independent of its type of secondary structure. This would appear to be against the fact that one type of amino acid may prefer some type of secondary structure to the others. The **2nd assumption** is that the hidden state (i.e., the distribution of backbone angles) at position *i* is only dependent on the type of amino acid residue and the type of secondary structure at that particular position. This contradicts with the finding in (Jha et al., 2005) that the angles at any particular position *i* depends on at least the three residues at positions *i−1*, *i* and *i+1*.

These assumptions are required in (Hamelryck et al., 2006) because the parameters in the HMM model are estimated by maximizing the joint probability $P(A, X, S)$ of a primary sequence *A*, secondary structure *X* and hidden states *S* (i.e., angles). For a reasonable estimate of the joint probability $P(A, X, S)$, these assumptions are necessary because of the sparsity in the training data, as well as for the purpose to avoid overfitting. In FB5-HMM (Hamelryck et al., 2006), $P(A, X, S)$ is estimated using Law of total probability $\sum_H P(A|H)P(S|H)P(X|H)P(H)$ where $H$ is a possible hidden node sequence.

Due to the sparsity in the training data, it is also very difficult to incorporate more complexity into the HMM model without elevating the risk of overfitting, which restricts the expressive power of the HMM model. The problem is not merely limited to the HMM model, it is actually a common problem of all the generative learning methods (e.g., HMM) that relies on optimizing the joint probability of states and observations, where the primary sequence and secondary structure are both observations. As an alternative, discriminative learning methods such as CRFs (Conditional Random Fields) can be more expressive and at the same time keeping the risk of overfitting under control. Discriminative learning is different from generative learning in that the former one optimizes the conditional probability of states on the observations while the latter one the joint probability of states and observations.

This chapter presents the first-order CRF-Sampler, an extensible and fully automatic framework, for effective protein conformation sampling, based on a probabilistic graphical model Conditional Random Fields (Lafferty et al., 2001; Sha and Pereira, 2003). Similar to the HMM model, CRF-Sampler samples backbone angles from a continuous space using sequence and secondary structure information. CRF-Sampler also models the dependency between the angles at two adjacent positions. CRF-Sampler differs from the HMM model in the following aspects. First, CRF-Sampler is more expressive than the HMM model. The backbone angles at position $i$ can depend on residues and secondary structures at many positions instead of only one. In CRF-Sampler, a sophisticated model topology and feature set can be defined to describe the dependency between sequence and structure without worrying about learning of model parameters. Different from the HMM model, in which the model complexity (hence risk of overfitting) roughly equals to the number of parameters in the model, the effective complexity of CRF-Sampler is regularized by a Gaussian prior of its parameters, allowing the user to achieve a balance between model complexity and expressivity. Second, CRF-Sampler does not assume that primary sequence is independent of secondary structure in determining backbone angles. Instead, CRF-Sampler can automatically learn the relative importance of primary sequence and secondary structure. Finally, CRF-Sampler can easily incorporate sequence profile (i.e., position-specific frequency matrix) and predicted secondary structure likelihood scores into the model to further improve sampling performance. Our experimental results demonstrate that, using only compactness and self-avoiding constraints, CRF-Sampler can quickly generate more native-like conformations than the HMM model and best decoys closer to their natives.

## 3.2 CRF model for sequence-structure relationship

Conditional random fields (CRFs) are probabilistic graphical models that have been extensively used in modeling sequence data. Please refer to (Lafferty et al., 2001) and (Sha and Pereira, 2003) for a complete description of CRFs. Here, we describe how to predict the backbone angles of a protein from its primary sequence and secondary structure using CRFs. In this context, the primary sequence and secondary structure of a protein are called observations and its backbone angles are hidden states or labels.

Let $o = \{o_1, o_2, \ldots, o_N\}$ denote an observation sequence of length $N$ where $o_i$ is an observed object. Each observed object can be a residue or a secondary structure type or their combination. Let $H = \{h_1, h_2, \ldots, h_c\}$ be a finite set of labels (also called states), each representing a distribution of backbone angles. Let $S = \{S_1, S_2, \ldots, S_N | S_i \in H\}$ be a sequence of labels corresponding to the observation $o$. As opposed to the HMM model defining a joint probability of the label sequence $S$ and the observation $o$, our CRF model defines the conditional probability of $s$ given $o$ as follows.

$$P_\Lambda(S|o) = \frac{\exp(\sum_{i=1}^{N} F(S, o, i))}{Z(o)} \tag{3.1}$$

where $\Lambda = (\lambda_1, \lambda_2, \ldots \lambda_p)$ is the model parameter and $Z(o) = \sum_{S'} \exp(\sum_{i=1}^{N} F(S', o, i))$ is a normalization factor summing over all the possible label sequences for a given observation sequence. $F(S,o,i)$ is the sum of the CRF features at sequence position $i$:

$$F(S, o, i) = \sum_k \lambda_k e_k(S_{i-1}, S_i) + \sum_l \lambda_l v_l(o, S_i) \tag{3.2}$$

where $e_k(S_{i-1}, S_i)$ and $v_l(o, S_i)$ are called edge and label feature functions, respectively.

The edge and label functions are defined as

$$e_k(S_{i-1}, S_i) = [S_{i-1} = h_1][S_i = h_2] \tag{3.3}$$

and

$$v_l(o, S_i) = [x_l(o, i)][S_i = h] \tag{3.4}$$

where $S_i = h$ indicates that the label (or state) at position $i$ is $h$. And $x_l(o, i)$ is a logical context predicate indicating whether or not the context of the observation sequence $o$ at position $i$ holds a particular property or fact of empirical data. [$f$] is equal to 1 if the logical expression $f$ is true, and zero otherwise. Note that we can also define the edge feature function $e_k(S_{i-1}, S_i)$ as $[x_l(o, i)][S_{i-1} = h_1][S_i = h_2]$, to capture relationship between two adjacent labels and observations. By expanding Equation 3.1 using Equation 3.2, 3.3, and 3.4 and merging the same items, the conditional probability can also be reformulated as follows.

$$P_\Lambda(S|o) = \frac{\exp(\sum_{i=1}^{N} F(S,o,i))}{Z(o)} = \frac{\exp(\sum_k \lambda_k C_k(S,o))}{\sum_{S'} \exp(\sum_k \lambda_k C_k(S',o))} \tag{3.5}$$

where $C_k(S,o)$ represents the occurring times of the $k^{th}$ feature in a pair of label sequence $s$ and observation sequence $o$ and the model parameter $\lambda_k$ is the weight of this feature. Here, the parameter $\lambda_k$ does not correspond to the log probability of an event (as in the HMM model) Instead, it is a real-valued weight that either raises or lowers the "probability mass" of $s$ relative to other possible label sequences. The parameter $\lambda_k$ can be negative, positive, or zero.

The CRF model is more expressive than the HMM model. First, we do not have to interpret the parameter $\lambda_k$ as the log probability of an event. Second, CRFs do not have to assume that the observation object at one position is independent of other objects. That is, for any $v_l(o, S_i)$, the label at position $i$ can depend on many observed objects in the observation sequence or even the whole observation sequence. In addition, $S_i$ can also depend on any nonlinear combination of several observed objects. Therefore, the CRF model can accommodate complex feature sets that may be difficult to incorporate within a generative HMM model. The underlying reason is that CRFs only optimize the conditional probability $P_\Lambda(S|o)$ instead of joint probability $P_\Lambda(S,o)$, avoiding calculating the generative probability of the observation sequence.

**Model parameter estimation**

Given a set of observation sequences and their corresponding label sequences ($o^i$, $S^i$), CRFs train its parameter $\Lambda = \{\lambda_1, \lambda_2, ..., \lambda_p\}$ by maximizing the conditional log-likelihood $L$ of the data:

$$L = \sum_t \log p_\Lambda(S^t|o^t) - \frac{1}{2\sigma^2}\|\lambda\|_1 \tag{3.6}$$

This kind of training is also called discriminative training or conditional training. Different from the generative training in the HMM model, discriminative training directly optimizes the predictive ability of the model while ignoring the generative probability of the observation. The $L_1$ norm in the last term in Equation 3.6 is a regularization item to deal with the sparsity in the training data. When the complexity of the model is high (i.e., the model has many features and parameters) and the training data is sparse, overfitting may occur and it is possible that many models can fit the training data. To prevent this, we place a Gaussian prior, $\exp\left(\frac{1}{2\sigma^2}\sum_k \lambda_k\right)$, on the model parameter to choose the model with a "small" parameter. This regularization can improve the generalization capability of the model in both theory and practice.

The objective function in Equation 3.6 is convex and hence theoretically a globally optimal solution can be found using any efficient gradient-based optimization technique. There is no analytical solution to the above equation for a real-world application. Quasi-Newton methods such as L-BFGS (Liu and Nocedal, 1989) can be used to solve the above equation and usually can converge to a good solution within a couple of hundred iterations. The log-likelihood gradient component of $\lambda_k$ is

$$\frac{\partial L}{\partial \lambda_k} = \sum_i C_k(S^i, \sigma^i) - \sum_i \sum_S p_\Lambda(S|o^i) C_k(S, o^i) - \frac{\lambda_k}{\sigma^2} \tag{3.7}$$

The first two items on the right of the above equation is the difference between the empirical and the model expected values of feature count $C_k$. The expected value $\sum_S p_\Lambda(S|o) C_k(S, o)$ for a given $o$ can be computed using a simple dynamic programming algorithm if the model only has edge feature functions defined in Equation 3.3.

**Model topology**

As illustrated in Figure 7, we use a CRF model to capture the relationship between a protein sequence and its (pseudo) backbone angles. Let $S_i$ denote the label at position $i$. Each label represents a distribution of backbone angles in a protein position. We use the Sine model (Mardia et al., 2007; Singh et al., 2002) to describe the distribution of the $(\phi, \psi)$ angles and the FB5 mode (Kent, 1982) for the $(\theta, \tau)$ angles, respectively. Each label depends on a window of residues in the primary sequence, their secondary structure types, and any nonlinear combinations of them. There is also interdependence between two adjacent labels. The CRF model is not necessary a linear-chain graph. It can be easily extended to model the long-range relationship between two positions. For example, if distance restraints are available from NMR or threading programs, then we can add some edges in the CRF model to capture the long-range interactions between two nonadjacent residues. In the CRF model, we do not assume that the residues in primary sequence are independent of each other and that primary sequence is independent of secondary structure. Our CRF model can easily capture this kind of interdependence in its conditional probability $p_\Lambda(S|o)$.

In this example, $S_i$ (i.e., the label at position $i$) depends on the residues and secondary structure types at positions *i–2, i–1, i, i+1,* and *i+2* and any nonlinear combinations of them. There is also interdependence between two adjacent labels. This CRF model can also be extended to incorporate long-range interdependence between two labels.

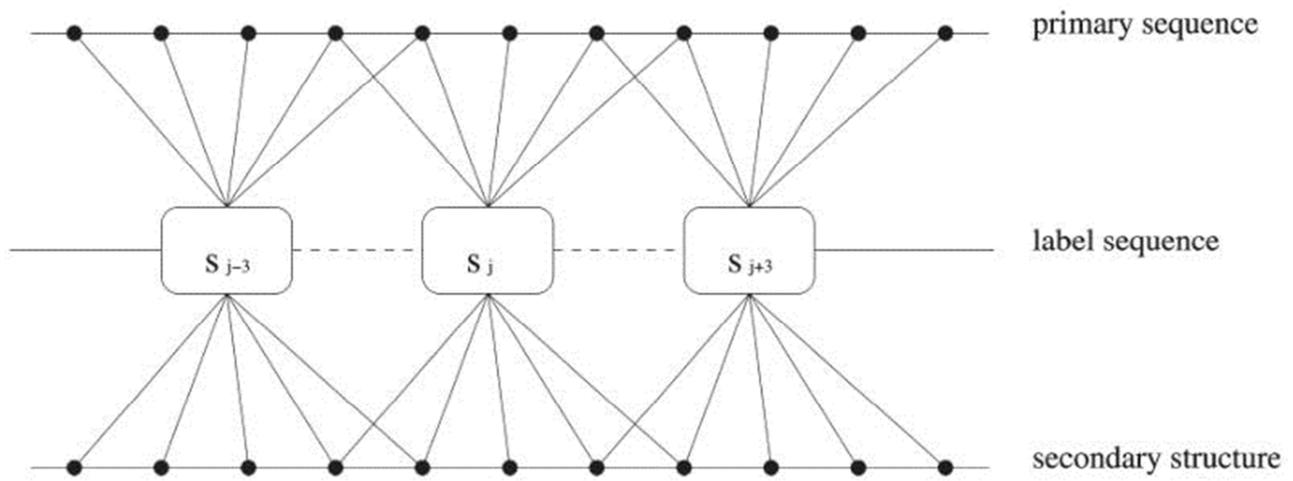

Figure 7 An example CRF model for protein conformation sampling.

**Model features**

CRF-Sampler uses two different types of feature functions. At each position $i$, CRF-Sampler uses $e(S_{i-1},S_i) = [S_{i-1} = h_1] [S_i = h_2]$ as its edge feature function. That is, currently, we only consider the first order dependence between labels. We are also investigating the second order dependence between labels. We use the following label feature functions to model the relationship among primary sequence, secondary structure, and backbone angles. Meanwhile, $w$ is half of the window size.

1. $v_{1,j}(o,S_i) = [A_{i+j} = a] [S_i = h]$. This feature set describes the interdependence between the label at position $i$ and the residue at position $i+j$, where $-w \leq j \leq w$. A feature in this set is identified by a triple $(a, h, j)$.
2. $v_{2,j}(o,S_i) = [X_{i+j} = x] [S_i = h]$. This feature set describes the interdependence between the label at position $i$ and the secondary structure type at position $i + j$, where $-w \leq j \leq w$. A feature in this set is identified by a triple $(x, h, j)$.
3. $v_{3,j}(o,S_i) = [A_{i+j} = a] [X_{i+j} = x] [S_i = h]$. This feature set describes the interdependence among the label at position $i$, the residue at position $i + j$, and the secondary structure type at position $i + j$ where $-w \leq j \leq w$. A feature in this set represents a nonlinear combination of secondary structure and primary sequence and is identified by a quadruple $(a, x, h, j)$.

We use a window size 9 (i.e., $w = 4$), which is slightly better than a window size 5 when predicted secondary structure information is used as input of CRF-Sampler (see Table 2). In total there are more than ten thousand features in CRF-Sampler quantifying protein sequence-structure relationship.

**Extension to continuous-valued observations**

We can also extend CRF-Sampler to make use of sequence profile and predicted secondary structure likelihood scores to improve sampling performance. In this case, for each protein, the observation is not two strings any more but consists of two matrices. One is the position-specific frequency matrix containing $20 \times N$ entries (where $N$ is the number of residues in a protein); each element in this matrix is the occurring frequency of one amino acid at a given position. The other matrix is the PSIPRED-predicted secondary structure likelihood matrix containing $3 \times N$ elements; each element is the predicted likelihood of one secondary structure type at a specific position. To use this kind of continuous-valued observations, we extend label feature functions as follows. In defining $v_{l,j}(o, S_i)$ ($l = 1,2,3$), instead of assigning $[A_{i+j} = a]$ to a binary value (i.e., 0 or 1), we assign $[A_{i+j} = a]$ to the frequency of amino acid $a$ appearing at position $i + j$. Similarly, we assign $[X_{i+j} = x]$ to the PSIPRED-predicted likelihood of secondary type $x$ at position $i + j$.

Table 2 F1-Values (%) of CRF-Sampler with Respect to Window Size

| Window size | a | b | c | d |
|---|---|---|---|---|
| 1 | 10.01 | 10.02 | 15.3 | 17.34 |
| 5 | 14.08 | 17.5 | 19.65 | 20.49 |
| 9 | 14.76 | 18.58 | 19.81 | 20.89 |

F1-value is an even combination of precision (*p*) and recall (*r*) and defined as $2pr/(p + r)$. a, trained and tested using primary sequence; b, trained and tested using PSI-BLAST sequence profile; c, trained and tested using primary sequence and PSIPRED-predicted secondary structure; d, trained and tested using PSI-BLAST sequence profile and PSIPRED-predicted secondary structure confidence scores.

## 3.3 Conformation sampling algorithm

**Sample one conformation for the whole protein**

Given a CRF model and its parameters, we used a forward-backward sampling algorithm to generate protein conformations. The algorithm is an extension of the sampling algorithm described for the HMM model (Hamelryck et al., 2006). The major difference is that our sampling algorithm needs to deal with many more sophisticated features and we also need to transform likelihood to probability in sampling. Let $v_l(i, h)$ denote a label feature function associated with position $i$ and label $h$. For a given position $i$ and a label $h$, we recursively define and calculate $G(i, h)$ from N-terminal to C-terminal as follows.

$$G(0, h) = \exp\left(\sum_l \lambda_l v_l(0, h)\right)$$

$$G(i, h) = \exp\left(\sum_l \lambda_l v_l(i, h)\right) \sum_{\bar{h}} G(i-1, \bar{h}) e^{\lambda_{\bar{h},h}}$$

where $\lambda_l$ is the trained parameter for the label feature $v_l(,)$ and $\lambda_{\bar{h},h}$ is the trained parameter for the edge feature $e(\bar{h}, h)$. After $G(N-1, h)$ ($N$ is the protein size) is calculated, we can sample a conformation from C-terminal to N-terminal. First, we sample the label $h$ for the last position according to probability

$$p(h) = \frac{G(N-1, h)}{\sum_{h'} G(i, h')}$$

Then we sample the label $\bar{h}$ for position $i$ according to probability

$$p(\bar{h}) = \frac{G(i, \bar{h}) e^{\lambda_{\bar{h},h}}}{\sum_{h'} G(i, h') e^{\lambda_{h',h}}}$$

assuming that the sampled label at position $i + 1$ is $h$. Note that each label corresponds to a distribution of backbone angles. Based on the sampled labels, we can sample the two backbone angles for each position and build a backbone conformation for the protein.

**Resample a small segment of the backbone conformation**

Given a backbone conformation, we generate the next conformation by resampling a small segment of the protein. First, we randomly sample the starting position of the segment and its length. The length is uniformly sampled from 1 to 15. We resample the labels of the segment including positions $i, i + 1, …, j,$ conditioned on the current labels at positions $i − 1$ and $j + 1$. Suppose that the labels at positions $i − 1$ and $j + 1$ are $h_1$ and $h_2$, respectively. We calculate $\bar{G}(k, h)$ $(i \leq k \leq j)$ from position $i$ to $j$ as follows.

$$\bar{G}(i, h) = \exp\left(\sum_l \lambda_l v_l(i, h)\right) e^{\lambda_{h_1, h}}$$

$$\bar{G}(k, h) = \exp\left(\sum_l \lambda_l v_l(k, h)\right) \sum_{\bar{h}} \bar{G}(k - 1, \bar{h}) e^{\lambda_{\bar{h}, h}}$$

After calculating $G(k, h)$ for all the $k$ between $i$ and $j$, we can sample the labels for the segment from $j$ to $i$. At position $j–1$, we sample a label $h$ according to probability

$$p(h) = \frac{\bar{G}(j - 1, h) e^{\lambda_{h, h_2}}}{\sum_h \bar{G}(j - 1, h) e^{\lambda_{h, h_2}}}$$

For any position $k$ $(i \leq k \leq j - 1)$, we sample a label $h$ according to probability

$$p(h) = \frac{\bar{G}(k, h) e^{\lambda_{h, \bar{h}}}}{\sum_h \bar{G}(k, h) e^{\lambda_{h, \bar{h}}}}$$

supposing $h$ is the sampled label at position $k+1$. After resampling the labels of this segment, we can resample the angles of this segment and then rebuild the backbone conformation.

**Folding simulation**

Since the focus of this chapter is the protein conformation sampling algorithm, we use only compactness and self-avoiding constraints to drive conformation search during the folding simulation process. We start with sampling the whole backbone conformation of a given protein and then optimize its conformation by minimizing the radius of gyration. Given a conformation, we generate its next potential conformation by resampling the local conformation of a small segment. If this potential conformation has no serious steric clashes among atoms, then we compare its radius with that of current conformation. If this potential conformation has a smaller radius, then we accept this conformation, otherwise reject it. This process is terminated if no better conformations can be found within 1000 consecutive resamplings. There is a steric clash if the distance between two $C_\alpha$ atoms is less than 4 Å.

# 3.4 Experiments and Results

**Data set**

The first-order CRF-Sampler is tested on the following proteins: 1FC2, 1ENH, 2GB1, 2CRO, 1CTF, 4ICB, 1AA2, 1BEO, 1DKT, 1FCA, 1FGP, 1JER, 1NKL, 1PGB, 1SRO, 1TRL, T0052 (PDB code: 2EZM), T0056 (1JWE), T0059 (1D3B), T0061 (1BG8), T0064 (1B0N), and T0074 (1EH2). The first six proteins have been studied in (Simons et al., 1997) and (Hamelryck et al., 2006); and the last 18 in (Xia et al., 2000); the last six proteins are also CASP3 targets. We obtained a set of non-redundant protein structures using the PISCES server (Wang and Dunbrack, 2003) as our training data. Each protein in this set has resolution at least 2.0 Å, $R$ factor no bigger than 0.25 and at least 30 residues. Any two proteins in this set share no more than 30% sequence identity. To avoid overlap between the training data and the test proteins, we removed the following proteins from our training data:(1) the proteins sharing at least 25% sequence identity with our test proteins; (2) the proteins in the same fold class as our test proteins according to the SCOP classification (Murzin et al., 1995); and (3) the proteins having a TM-score ≥ 0.5 with our test proteins in case some recently released proteins do not have a SCOP ID. According to (Zhang and Skolnick, 2005b), if the TM-score of two protein structures is smaller than 0.5, then a threading program such as PROSPECTOR_3 (Skolnick et al., 2004) cannot identify their similarity relationship with high confidence.

**Label assignment and distribution parameters**

To train CRF-Sampler, we also need to assign a label to each position in a protein. In this part, we only tested our algorithm on the (θ, τ) representation of a protein backbone conformation. There can be various methods to assign a label to a protein position. For example, we can cluster all the (θ, τ) angles into dozens of groups; each group corresponds to a label. Here, we just simply use the five-residue fragment libraries developed by Kolodny *et al.*(Kolodny et al., 2002) since these libraries have already been carefully designed. The library containing 100 five-residue fragments is used as the set of hidden labels; each label corresponds to a cluster in the fragment library. We calculated the (θ, τ) distribution for each cluster from the training proteins using the KentEstimator program enclosed in Mocapy(Hamelryck et al., 2006). Only the angles of the middle residue in a fragment are used to calculate the angle distribution parameters. We also tested other four-residue and five-residue fragment libraries developed by Kolodny *et al.* and it turns out that the five-residue fragment library with 100 clusters yields the best performance.

**Parameter tuning**

We randomly divided the training proteins into five sets of same size and then used them for five-fold cross validation. We trained CRF-Sampler using several different regularization factors [i.e., $\sigma^2$ in Equation 3.7]: 50, 100, 200, 400, and 800 and choose the one with the best F1-value. F1-value is a widely used measurement of the prediction capability of a machine learning model in the machine learning community. F1-value is an even combination of precision ($p$) and recall ($r$) and defined as $2pr/(p + r)$. The higher the F1-value is, the better. When both PSI-BLAST sequence profile and PSIPRED-predicted secondary structure likelihood scores are used and a window size 9 is used to define the model features, the average F1-values for regularization factors 50, 100, 200, 400, and 800 are 20.82%, 20.89%, 20.83%, 20.71%, and 20.56%, respectively. In fact, there is no big difference among these regularization factors in terms of F1-value. However, we prefer to choose a small regularization factor 100 to control the model complexity. The regularization factor is the only parameter that we need to tune manually. All the other model parameters (i.e., weights for features) can be estimated automatically in training.

In addition, we also tested the performance of our algorithm with respect to window size in defining model features. As shown in Table 2, our experimental results indicate that when 100 labels are used in CRF-Sampler, a window size 5 can yield a much higher F1-value than a window size 1. Increasing the window size to 9 can improve the F1-value, but the improvement is small when PSIPRED-predicted secondary structure is used in CRF-Sampler. This may be because the predicted secondary structure also contains partial information of neighbor residues. In our remaining experiments, we used window size 9 to define model features for CRF-Sampler.

## Comparison with the HMM model

It is not easy to fairly compare two protein conformation sampling algorithms. Many *ab initio* folding programs use a sophisticated energy function to drive conformation search and it is hard to evaluate performance of their conformation sampling algorithms alone, without considering their energy functions. The focus of this chapter lies in only protein conformation sampling algorithm. To evaluate the sampling algorithm, we drive conformation search by minimizing the radius of gyration instead of a well-designed energy function. Here we compare CRF-Sampler mainly with the HMM method described in (Hamelryck et al., 2006), which is also a protein conformation sampling algorithm and drives conformation search by minimizing the radius instead of an energy function. The major difference between CRF-Sampler and the HMM model is that the former generates conformations using a CRF model while the latter uses an HMM model. We tested CRF-Sampler on six proteins studied in (Hamelryck et al., 2006) and compared the quality of the decoys generated by CRF-Sampler with those by the HMM model.

Since for most proteins without known structures, we cannot obtain their true secondary structures, here we compare the HMM model and CRF-Sampler using only PSIPRED-predicted secondary structure and sequence information as their inputs. As shown in Table 3, when only sequence information used, CRF-Sampler can generate decoys with much higher quality than the HMM model. The only exception is on 1ENH where the HMM model has a comparable performance with CRF-Sampler when only primary sequence is used. When only primary sequence is used, the number of good decoys (RMSD ≤ 6 Å) generated by CRF-Sampler is 2 ~ 4 times of that by the HMM model. This difference comes from the fact that in sampling the angles at one position, CRF-Sampler can directly take into consideration the effects of its neighbor residues. When predicted secondary structure is used, as shown in Table 4, CRF-Sampler is much better than the HMM model in generating good decoys and the best decoys. Among the six test proteins, CRF-Sampler is slightly worse than the HMM model on the Calbidin protein (PDB code: 4ICB) when predicted secondary structure type is used. For the other five proteins, CRF-Sampler can generate many more good decoys. CRF-Sampler can also generate the best decoys with much smaller RMSDs, although only 20,000 decoys are generated by CRF-Sampler for each test protein while 100,000 decoys by the HMM model for each test protein. By using PSI-BLAST sequence profile and predicted secondary structure likelihood scores as input, CRF-Sampler can achieve overall performance better than using primary sequence and predicted secondary structure types. This indicates the importance of using continuous-valued observations in a conformation sampling algorithm. By contrast, it is not easy for the HMM model to incorporate these kind of continuous-valued observations as input. Figure 8 visualizes the native structures and the best decoys (generated by CRF-Sampler) of the six test proteins.

Table 5 lists the percentage of correct secondary structure (i.e., Q3-value) of all the good decoys generated by CRF-Sampler for each test protein. A software P-SEA(Labesse et al., 1997) is used to calculate the secondary structure of a decoy. As shown in this table, even with primary sequence only, CRF-Sampler can generate decoys with pretty good Q3-values, better than the HMM model (see Table II in (Hamelryck et al., 2006)). This confirms that CRF-Sampler can capture well the relationship between a sequence stretch and its local conformation.

Table 3 Decoy Quality Comparison between the HMM Model and CRF-Sampler

Note: Only sequence information is used in both training and testing. In total, 100,000 decoys are generated by the HMM model.

Column "L" lists the length of the test proteins; column "α,β" lists the number of α-helices and β-strands of the test proteins; columns "Good" and "Best" list the percentage of good decoys (with RMSD ≤ 6 Å) and the RMSD of the best decoy, respectively.

[a]Trained and tested using primary sequence and the results are taken from (Hamelryck et al., 2006).
[b]Trained and tested using primary sequence and 40,000 decoys are generated.
[c]Trained and tested using PSI-BLAST sequence profile. Only 20,000 decoys are generated.

| Test proteins | | | HMM[a] | | CRF-Sampler[b] | | CRF-Sampler[c] | |
|---|---|---|---|---|---|---|---|---|
| Name, PDB code | L | α,β | Good (%) | Best (Å) | Good (%) | Best (Å) | Good (%) | Best (Å) |
| Protein A, 1FC2 | 43 | 2,0 | 9.59 | 2.7 | 20.9 | 2.08 | 24.8 | 2.09 |
| Homeodomain, 1ENH | 54 | 2,0 | 6.6 | 2.5 | 6.23 | 2.68 | 14 | 1.98 |
| Protein G, 2GB1 | 56 | 1,4 | 0.04 | 4.9 | 0.16 | 4.67 | 10.1 | 3.36 |
| Cro repressor, 2CRO | 65 | 5,0 | 0.46 | 3.9 | 1.94 | 4.05 | 13.3 | 2.37 |
| Protein L7/L12, 1CTF | 68 | 3,1 | 0.01 | 5.4 | 0.04 | 4.94 | 0.15 | 4.49 |
| Calbidin, 4ICB | 76 | 4,0 | 0.09 | 4.3 | 0.17 | 4.57 | 0.42 | 4.72 |

Table 4 Decoy Quality Comparison between the HMM Model17 and CRF-Sampler

Note: Both sequence and secondary structure information are used. In total, 100,000 decoys are generated by the HMM model while only 20,000 decoys by each CRF-Sampler.

Please refer to Table 3 for details of the Column definitions.

[a]Trained using true secondary structure and primary sequence while tested using predicted secondary structure (by PSIPRED(McGuffin et al., 2000)) and primary sequence and the results are taken from (Hamelryck et al., 2006).
[b]Trained and tested using predicted secondary structure (by PSIPRED) and primary sequence.
[c]Trained and tested using predicted secondary structure likelihood scores (by PSIPRED) and PSI-BLAST sequence profile.

| Test proteins | | | HMM[a] | | CRF-Sampler[b] | | CRF-Sampler[c] | |
|---|---|---|---|---|---|---|---|---|
| Name, PDB code | L | $\alpha, \beta$ | Good (%) | Best (Å) | Good (%) | Best (Å) | Good (%) | Best (Å) |
| Protein A, 1FC2 | 43 | 2,0 | 17.1 | 2.6 | 26.8 | 2.13 | 49.1 | 1.94 |
| Homeodomain, 1ENH | 54 | 2,0 | 12.2 | 3.8 | 16.7 | 2.29 | 22.4 | 2.32 |
| Protein G, 2GB1 | 56 | 1,4 | 0 | 5.9 | 26.4 | 3.05 | 23.3 | 2.91 |
| Cro repressor, 2CRO | 65 | 5,0 | 1.1 | 4.1 | 18.3 | 2.76 | 16.8 | 2.79 |
| Protein L7/L12, 1CTF | 68 | 3,1 | 0.35 | 4.1 | 3 | 4.04 | 2.4 | 3.7 |
| Calbidin, 4ICB | 76 | 4,0 | 0.38 | 4.5 | 0.24 | 4.45 | 0.51 | 4.63 |

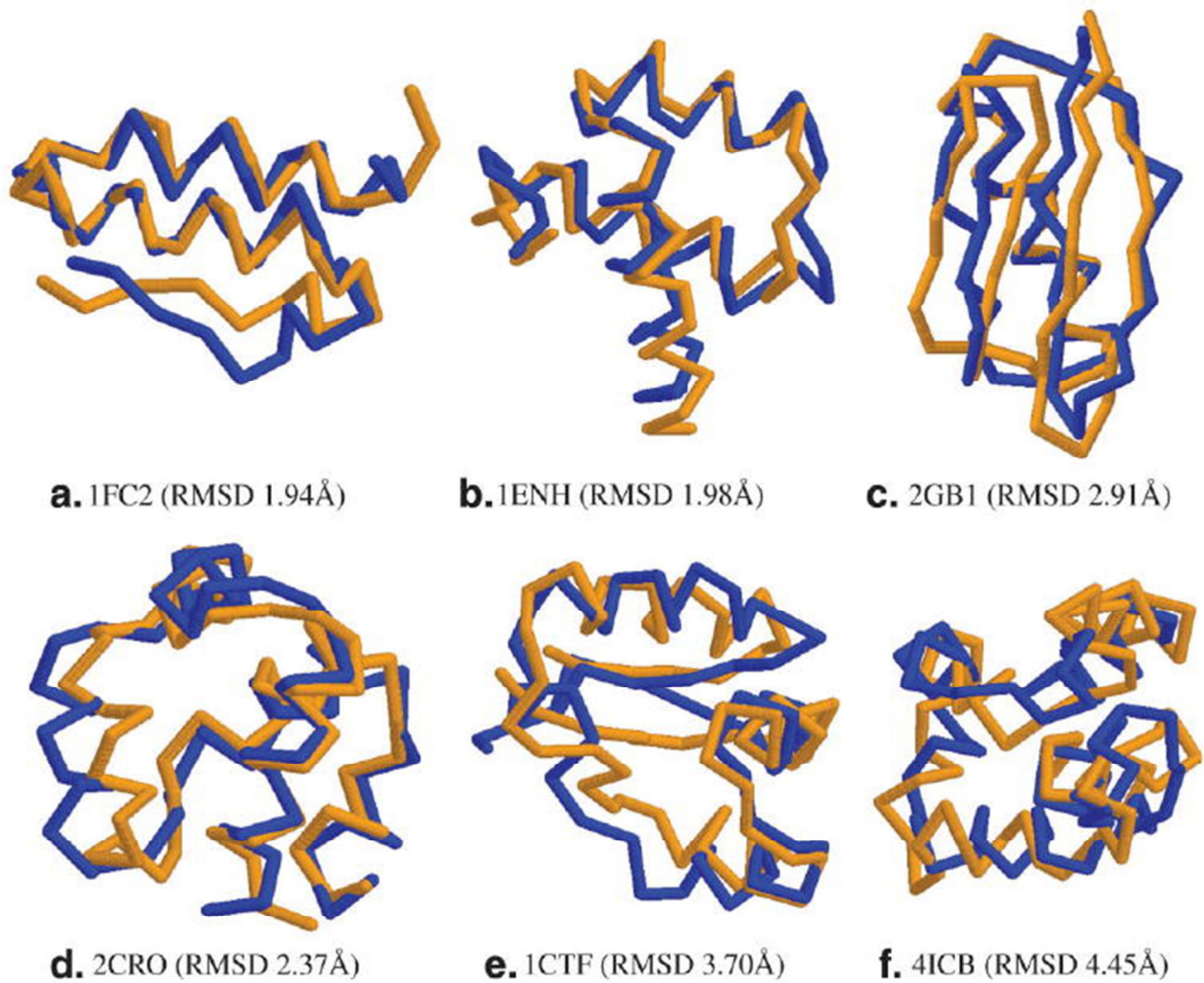

Figure 8 Native structures (in orange) and the best decoys of 1FC2, 1ENH, 2GB1, 2CRO, 1CTF, and 4ICB

Table 5 Secondary Structure Content of Good Decoys

Note:

[a] Percentage of correct secondary structure (Q3-value) and secondary structure content of good decoys (RMSD ≤ 6 Å) generated using primary sequence.

[b] Q3-value and secondary structure content of good decoys generated using PSI-BLAST sequence profile.

[c] Q3-value and secondary structure content of good decoys generated using primary sequence and PSIPRED-predicted predicted secondary structure.

[d] Q3-value and secondary structure content of good decoys generated using PSI-BLAST sequence profile and PSIPRED-predicted secondary structure likelihood scores.

|         | CRF-Sampler[a] |      |      |      | CRF-Sampler[b] |      |      |      |
|---------|------|------|------|------|------|------|------|------|
| Protein | Q3   | H    | E    | C    | Q3   | H    | E    | C    |
| 1FC2    | 66.5 | 61.3 | 1.8  | 36.9 | 75.2 | 49.4 | 1.9  | 48.7 |
| 1ENH    | 79.3 | 69.2 | 2.9  | 27.9 | 79.2 | 64.9 | 7.1  | 28   |
| 2GB1    | 65.2 | 29.5 | 24.9 | 45.7 | 61   | 24.1 | 27.6 | 48.2 |
| 2CRO    | 80   | 71.6 | 2.3  | 26.2 | 85.2 | 66   | 1.8  | 32.2 |
| 1CTF    | 67.1 | 50.3 | 9.5  | 40.1 | 62.6 | 41.8 | 19.7 | 38.5 |
| 4ICB    | 65   | 59.6 | 2.9  | 37.5 | 66.6 | 59   | 5    | 36   |

|         | CRF-Sampler[c] |      |      |      | CRF-Sampler[d] |      |      |      |
|---------|------|------|------|------|------|------|------|------|
| Protein | Q3   | H    | E    | C    | Q3   | H    | E    | C    |
| 1FC2    | 85.5 | 55.1 | 1    | 43.9 | 80.7 | 61.9 | 0.2  | 37.9 |
| 1ENH    | 86   | 64.3 | 2.9  | 32.8 | 85.1 | 67.3 | 3    | 29.7 |
| 2GB1    | 72   | 24.8 | 32.9 | 42.3 | 71.5 | 25.9 | 29.7 | 44.4 |
| 2CRO    | 87.1 | 69.3 | 1.5  | 29.1 | 85.9 | 67.5 | 1.6  | 31   |
| 1CTF    | 77.5 | 56.9 | 10.6 | 32.5 | 77.2 | 59.5 | 8.6  | 31.9 |
| 4ICB    | 65.7 | 62   | 3.81 | 34.2 | 67.1 | 63.1 | 2.5  | 34.4 |

## Comparison with Xia *et al*

In their paper, Xia *et al.* (Xia et al., 2000) have developed a hierarchical method to generate decoys. This method first exhaustively enumerates all the possible conformations on a lattice model for a given protein sequence and then builds conformations with increasing detail. At each step, this method chooses a good subset of conformations using hydrophobic compactness constraint and empirical energy functions such as RAPDF(Samudrala and Moult, 1998) and Shell(Park et al., 1997) and finally generate 10,000 or 40,000 decoys for a protein sequence. Table 6 lists the RMSD ranges of all the decoys generated by this method and CRF-Sampler for 18 test proteins. As shown in this table, CRF-Sampler can generate decoys with smaller RMSD on all the test proteins with less than 100 residues. CRF-Sampler is significantly better than this hierarchical method on 1CTF, 1NKL, 1PGB, 1SRO, 1TRL-A, T0052, T0059 and T0074 in terms of the best decoys. For those test proteins with more than 100 residues, CRF-Sampler is worse than the method of Xia *et al.* on two proteins (1AA2 and T0056) and better one on protein (T0064), and has comparable performance on 1JER. This may indicate that we need to improve CRF-Sampler further to search conformation space more effectively for a protein with more than 100 residues. We also used the Wilcoxon signed-rank test(Wilcoxon, 1945), a nonparametric alternative to the paired Student's *t*-test, to calculate the significance level at which CRF-Sampler is better than the method of Xia *et al.* Since we only had the RMSD ranges of the decoys generated by Xia *et al.*, we only considered the best decoys in calculating the statistical test. For the first 12 proteins in Table 6, we used the best decoys in a set of randomly chosen CRF-Sampler 10,000 decoys since Xia *et al.* only generated 10,000 final decoys for these proteins. For the last six proteins, we used their best decoys listed in Table 6. When the absolute RMSD difference between the best decoys is used to calculate the statistical test, CRF-Sampler is better than the method of Xia *et al.* at significance level 0.01(In fact the significance level is very close to 0.005). When the relative RMSD difference is used, CRF-Sampler is better than the method of Xia *et al.* at significance level 0.005. Finally, CRF-Sampler tends to generate decoys with larger RMSD variance because CRF-Sampler does not use any empirical energy functions to filter those bad conformations.

CRF-Sampler generated 20,000 decoys for each test protein using PSI-BLAST sequence profile and predicted secondary structure likelihood scores. Xia et al. (Xia et al., 2000) conducted a complete enumeration on a lattice model for each test protein and then generated 10,000 decoys for each of the first 12 proteins and 40,000 decoys for each of the six CASP3 targets, respectively, using predicted secondary structure and empirical energy functions as filters.

Table 6 Decoy Quality Comparison Between Xia et al. (Xia et al., 2000) and CRF-Sampler

| Test proteins | | | Xia *et al.* | CRF-Sampler |
|---|---|---|---|---|
| PDB code | $L$ | Class | All RMSD range | All RMSD range |
| 1aa2 | 108 | $\alpha$ | 6.18–15.28 | 7.34–17.06 |
| 1beo | 98 | $\alpha$ | 6.96–15.94 | 6.41–16.95 |
| 1ctf | 68 | $\alpha + \beta$ | 5.45–13.54 | 3.70–13.37 |
| 1dktA | 72 | $\beta$ | 6.68–14.79 | 6.14–15.51 |
| 1fca | 55 | $\beta$ | 5.09–12.06 | 4.98–12.90 |
| 1fgp | 67 | $\beta$ | 7.80–14.40 | 7.39–15.20 |
| 1jer | 110 | $\beta$ | 9.55–17.53 | 9.63–19.64 |
| 1nkl | 78 | $\alpha$ | 5.26–14.23 | 3.63–13.76 |
| 1pgb | 56 | $\alpha + \beta$ | 5.60–13.30 | 3.15–12.75 |
| 1sro | 76 | $\beta$ | 7.30–15.42 | 6.22–15.70 |
| 1trlA | 62 | $\alpha$ | 5.30–13.16 | 3.53–12.43 |
| 4icb | 76 | $\alpha$ | 4.74–13.28 | 4.63–13.93 |
| T0052 | 98 | $\beta$ | 10.6–16.3 | 7.58–19.17 |
| T0056 | 114 | $\alpha$ | 6.2–17.8 | 7.77–18.17 |
| T0059 | 71 | $\beta$ | 7.4–15.7 | 6.29–15.54 |
| T0061 | 76 | $\alpha$ | 6.0–14.0 | 5.35–14.84 |
| T0064 | 103 | $\alpha$ | 8.0–18.8 | 7.23–18.85 |
| T0074 | 98 | $\alpha$ | 6.3–16.5 | 4.85–15.72 |

We also calculated the secondary structure content of all the decoys using P-SEA (Labesse et al., 1997) and compared CRF-Sampler with PSIPRED(McGuffin et al., 2000) in terms of Q3-value. As shown in Table 7, CRF-Sampler can generate decoys with reasonable level of secondary structure accuracy. The average Q3-value of all the decoys generated by CRF-Sampler is 70.3% while the average PSIPRED Q3-value of these test proteins is 74.4%. The Wilcoxon signed-rank test indicates that in terms of Q3-value, CRF-Sampler is worse than PSIPRED at significance level 0.025.

Table 7 Percentage of correct secondary structure (Q3-value) and secondary structure scntent of all the decoys generated by CRF-Sampler, compared with PSIPRED Predictions

| Test proteins | | | CRF-Sampler | | | | PSIPRED | | | |
|---|---|---|---|---|---|---|---|---|---|---|
| PDB code | L | Class | Q3 | H | E | C | Q3 | H | E | C |
| 1aa2 | 108 | α | 78.5 | 53.9 | 5.4 | 40.7 | 88 | 56.5 | 1.9 | 41.7 |
| 1beo | 98 | α | 68.4 | 51.3 | 9.9 | 38.7 | 65.3 | 42.9 | 6.1 | 51 |
| 1ctf | 68 | α + β | 76.9 | 59.8 | 8.6 | 31.6 | 77.9 | 55.9 | 13.2 | 30.9 |
| 1dktA | 72 | β | 53.7 | 16.3 | 28.4 | 55.3 | 50 | 15.3 | 30.6 | 54.2 |
| 1fca | 55 | β | 71.9 | 4.7 | 16.8 | 78.5 | 83.6 | 0 | 16.4 | 83.6 |
| 1fgp | 67 | β | 57.3 | 4.5 | 28.5 | 66.9 | 64.3 | 0 | 35.7 | 64.2 |
| 1jer | 110 | β | 63.2 | 14.1 | 34.7 | 51.2 | 74.3 | 8.3 | 37.6 | 54.1 |
| 1nkl | 78 | α | 87.7 | 75.3 | 0.4 | 24.3 | 92.3 | 73.1 | 0 | 26.9 |
| 1pgb | 56 | α + β | 67.3 | 25.6 | 27.8 | 46.7 | 80.4 | 21.4 | 50 | 28.6 |
| 1sro | 76 | β | 59.2 | 7.7 | 28.8 | 63.5 | 56.6 | 10.5 | 43.4 | 46.1 |
| 1trlA | 62 | α | 80.9 | 70 | 0.4 | 29.6 | 83.9 | 67.7 | 0 | 32.3 |
| 4icb | 76 | α | 67.2 | 61.8 | 3.5 | 34.8 | 67.1 | 60.5 | 2.6 | 36.8 |
| T0052 | 98 | β | 54.1 | 6 | 36.1 | 57.9 | 54.5 | 8.9 | 42.6 | 48.5 |
| T0056 | 114 | α | 83.3 | 71.2 | 1.4 | 27.5 | 81.6 | 74.6 | 0 | 25.4 |
| T0059 | 71 | β | 64.9 | 10.4 | 41.3 | 48.3 | 72.2 | 8.3 | 52.8 | 38.9 |
| T0061 | 76 | α | 59.9 | 42 | 13.7 | 44.3 | 71.1 | 40.8 | 14.5 | 44.7 |
| T0064 | 103 | α | 87.2 | 66.2 | 2.7 | 31 | 94.2 | 67 | 0 | 33 |
| T0074 | 98 | α | 84.9 | 53.6 | 3.2 | 43.2 | 83.2 | 55.8 | 0 | 44.2 |

The average Q3-value of all the decoys is 70.3% while the average PSIPRED Q3-value is 74.4%. CRF-Sampler is worse than PSIPRED at significance level 0.025.

## Comparison with Rosetta

Here, we compare CRF-Sampler with the well-known fragment-assembly-based program Rosetta (Simons et al., 1997). Rosetta uses multiple sequence alignment information to choose 25 fragments for each sequence segment of nine residues and then assembles them into decoys using a time-consuming simulated annealing procedure. Rosetta drives conformation search using a well-developed energy function and generates a few hundred decoys, while CRF-Sampler generates 20,000 decoys without using any energy function. Although this comparison is interesting, we want to point out it is also unfair to both CRF-Sampler and Rosetta. On one hand, CRF-Sampler does not use an energy function to drive conformation search. On the other hand, the result for Rosetta is taken from an article published in 1997, which may not be the performance of the state-of-the-art Rosetta. As indicated in Table 8, by quickly generating a large number of decoys, CRF-Sampler can obtain decoys with much smaller RMSD than Rosetta. However, it is also not surprising that Rosetta can generate higher percentage of good decoys for five out of six test proteins by using a time-consuming energy minimization procedure.

**Computational efficiency**

Although we have not optimized the C++ code of CRF-Sampler, CRF-Sampler can quickly generate a decoy within seconds for a test protein. Table 9 the approximate running time in minutes spent by CRF-Sampler generating 100 decoys for each test protein, on a single 2.2 GHz CPU. As indicated in this table, CRF-Sampler can generate decoys for these test proteins very quickly. It takes CRF-Sampler approximately 1 h to generate 100 decoys for protein G (2GB1) and no more than ten minutes for 1FC2. It does not increase the running time of CRF-Sampler by using more information as input such as PSI-BLAST sequence profile and secondary structure likelihood scores. Instead, using them tend to reduce the running time of CRF-Sampler, maybe because of the reduction in the entropy of conformation search space.

Table 8  Decoy Quality Comparison Between ROSETTA2 and CRF-Sampler

Note: Only sequence information is used in both training and testing. ROSETTA generates a few hundred decoys using energy function optimization while CRF-Sampler generates 20,000 decoys by minimizing radius of gyration.

Refer to Table 3 for details of the Column definitions.

[a]Multiple sequence alignment information is used and the results are taken from (Simons et al., 1997).
[b]Trained and tested using primary sequence.
[c]Trained and tested using PSI-BLAST sequence profile.

| Test proteins | | | ROSETTA[a] | | CRF-Sampler[b] | | CRF-Sampler[c] | |
|---|---|---|---|---|---|---|---|---|
| Name, PDB code | L | α, β | Good (%) | Best(Å) | Good (%) | Best (Å) | Good (%) | Best (Å) |
| Protein A, 1FC2 | 43 | 2,0 | 95 | 3.3 | 20.9 | 2.08 | 24.8 | 2.09 |
| Homeodomain, 1ENH | 54 | 2,0 | 47 | 2.7 | 6.23 | 2.68 | 14 | 1.98 |
| Protein G, 2GB1 | 56 | 1,4 | 0 | 6.3 | 0.16 | 4.67 | 10.1 | 3.36 |
| Cro repressor, 2CRO | 65 | 5,0 | 18 | 4.2 | 1.94 | 4.05 | 13.3 | 2.37 |
| Protein L7/L12, 1CTF | 68 | 3,1 | 6 | 5.3 | 0.04 | 4.94 | 0.15 | 4.49 |
| Calbidin, 4ICB | 76 | 4,0 | 17 | 4.7 | 0.17 | 4.57 | 0.42 | 4.72 |

Table 9 Approximate Running Time in Minutes Spent by CRF-Sampler in Generating 100 Decoys

Note: The features are used in both the training and testing data set. The unit of time in the grids is Minute.

| Features | 1FC2 | 1ENH | 2GB1 | 2CRO | 1CTF | 4ICB |
|---|---|---|---|---|---|---|
| primary sequence only | 7 | 13 | 45.5 | 29.5 | 46 | 37.5 |
| PSI-BLAST sequence profile only | 8 | 12 | 63 | 14.5 | 78 | 43.5 |
| PSIPRED-predicted secondary structure and primary sequence | 6 | 7.5 | 67.5 | 7.5 | 16 | 29.5 |
| PSIPRED-predicted secondary structure confidence scores and PSI-BLAST sequence profile | 4 | 8 | 56 | 7 | 13 | 27.5 |

# 3.5 Conclusion

This chapter presented an extensible and fully automatic framework CRF-Sampler that can be used to effectively sample conformations of a protein from its sequence information and predicted secondary structure. CRF-Sampler uses thousands of parameters to quantify the relationship among backbone angles, primary sequence and secondary structure without worrying about risk of overfitting. Experimental results demonstrate that CRF-Sampler is more effective in sampling conformations than the HMM model(Hamelryck et al., 2006). CRF-Sampler is quite flexible. Using CRF-Sampler, the user only needs to choose a set of appropriate features describing the relationship between protein sequence and structure. CRF-Sampler can take care of the remaining tasks such as parameter estimation and conformation sampling.

The first order CRF-Sampler only takes into consideration the dependency between two adjacent positions. It may explore the conformation space of medium-sized proteins ($\geq$ 100 residues) more effectively by incorporating the interdependency among more residues. For example, given the labels at positions $i$ and $i + 1$, there are on average only 10 possible labels at position $i + 2$. Given the labels at positions $i$ and $i + 2$, there are on average only 16 possible labels at position $i + 4$. If this kind of constraint information is incorporated into CRF-Sampler, it can greatly reduce the entropy of conformation search space and is very likely to scale CRF-Sampler up to proteins with more than 100 residues.

In this chapter, we drove the conformation optimization by minimizing the radius of gyration. Our next step is to couple CRF-Sampler with a good energy function such as DOPE (Shen and Sali, 2006) and DFIRE (Zhou and Zhou, 2002), to do real protein structure prediction. The decoys generated by CRF-Sampler can also be used to benchmark energy functions since CRF-Sampler does not employ any energy function to generate these decoys and thus no energy-bias is introduced into these decoys.

We can also incorporate other predicted information such as solvent accessibility and contact capacity (i.e., the number of contacts for a residue) into CRF-Sampler, which may improve sampling performance. In addition, if some distance restraints can be obtained from NMR data or comparative modeling, then it is also possible to extend CRF-Sampler to incorporate long-range interdependency. CRF-Sampler can also be extended to make use of other NMR data sources such as chemical shifts (Neal et al., 2006) and residue dipolar coupling (RDC) data (Meiler et al., 2000).

We will address the above issues in the discussion of the second-order CRF-Sampler in the next chapter. Since the experimental results indicate that using predicted secondary structure we can dramatically improve the sampling performance of CRF-Sampler, compared to using

PSI-BLAST sequence profile only, we will employ those features in the second order CRF-Sampler.

# Chapter 4. Second-order CRF-Sampler with Energy

## 4.1 Introduction

In Chapter 3, we have described a protein conformation sampling algorithm based on the 1st-order conditional random fields (CRF) and directional statistics. The CRF model is a generalization of the HMM models and is much more powerful. Our CRF model can accurately describe the complex sequence-angle relationship and estimate the probability of a conformation, by incorporating various sequence and structure features and directly taking into consideration the nearest neighbor effects. We have shown that by using the 1st-order CRF model, we can sample conformations with better quality than Hamelryck et al.'s FB5-HMM (Zhao et al., 2008). All these studies have demonstrated that it is promising to do template-free modeling without using discrete representations of protein conformational space.

This chapter presents the first template-free modeling method that can search conformations in a continuous space and at the same time achieves performance comparable to the popular fragment assembly methods. This method differs from our previous work discussed in chapter 3 (Zhao et al., 2008) and FB5-HMM (Hamelryck et al., 2006) in that the latter two only describe a method for conformation sampling in a continuous space, but did not demonstrate that this sampling technique actually lead to a template-free modeling method with comparable performance as the fragment assembly method. By contrast, in this chapter we describe a 2nd-order CRF model of protein conformational space and show that with a simple energy function, the 2nd-order CRF model works well for template-free modeling. We will show that it is necessary to use the 2nd-order model instead of the 1st-order model described in our previous work since the former can dramatically improve sampling efficiency over the latter, which makes the 2nd-order model feasible for real-world template-free modeling. Blindly tested in the CASP8 evaluation, our CRF method compares favorably with the Robetta server (Misura et al., 2006; Simons et al., 1997), especially on alpha and small beta proteins. Our method also generated 3D models better than template-based methods for a couple of CAP8 hard targets.

## 4.2 Methods

### A 2nd-order CRF model of protein conformation space

We have described a 1st-order CRF model for protein conformation sampling in Chapter 3. Here we extend our 1st-order CRF model to a 2nd-order model to more accurately capture local sequence-angle relationship.

Given a protein with solved structure, we can calculate its backbone angles at each position and determine one of the 100 groups (i.e., states or labels) in which the angles at each position belong. Each group is described by an FB5 distribution. Let $S = \{S_1, S_2, ..., S_{100}\}$ ($S_i \in H$) denote such a sequence of states/labels (i.e., FB5 distributions) for this protein. We also denote the sequence profile of this protein as $M$ and its secondary structure as $X$. As shown in Figure 9, our CRF model defines the conditional probability of $S$ given $M$ and $X$ as follows.

$$P_\Lambda(S|M,X) = \frac{\exp(\sum_{i=1}^{N} F(S,M,X,i))}{Z(M,X)} \tag{4.1}$$

where $\Lambda = (\lambda_1, \lambda_2, ..., \lambda_p)$ is the model parameter and

$$Z(M,X) = \sum_S \exp\left(\sum_{i=1}^{N} F(S,M,X,i)\right)$$

is a normalization factor summing over all the possible labels for the given $M$ and $X$.

Comparing to Equation 3.1, we will use both the sequence profile $X$ and predicted secondary structure $M$ as our observation features, since we have learned from the results in Chapter 3 that this combination contains information for generating higher percentage of good decoys.

$F(S, M, X, i)$ consists of two edge features and two label features at position $i$. It is given by

$$F(S,M,X,i) = e_1(S_{i-1}, S_i) + e_2(S_{i-1}, S_i, S_{i+1}) + \sum_{j=l-w}^{i+w} v_1(S_i, M_j, X_j)$$
$$+ \sum_{j=l-w}^{i+w} v_2(S_{i-1}, S_i, M_j, X_j) \tag{4.2}$$

where $e_1(s_{i-1}, s_i)$ and $e_2(s_{i-1}, s_i, s_{i+1})$ are the 1st-order and 2nd-order edge feature functions, respectively. $v_1(s_i, M_j, X_j)$ and $v_2(s_{i-1}, s_i, M_j, X_j)$ are the 1st-order and 2nd-order label feature functions, respectively. If we remove $e_2(s_{i-1}, s_i, s_{i+1})$ and $v_2(s_{i-1}, s_i, M_j, X_j)$, then we can get a 1st-order CRF model.

The two edge functions model local conformation dependency, given by

$$e_1(S_{i-1}, S_i) = \lambda(h', h'')[S_{i-1} = h'][S_i = h''] \tag{4.3}$$

$$e_2(S_{i-1}, S_i, S_{i+1}) = \lambda(h', h'', h''')[S_{i-1} = h'][S_i = h''][S_{i+1} = h'''] \tag{4.4}$$

Meanwhile, $[s_i = h]$ is an indicator function, which is equal to 1 if the state at position $i$ is $h \in H$, otherwise 0; $\lambda(h', h'')$ is a model parameter identified by two states $h'$ and $h''$; and $\lambda(h', h'', h''')$ is a model parameter identified by three states. The two label feature functions are given by

$$v_1(S_i, M_j, X_j) = \sum_S \sum_{aa} \lambda(j-1, s, aa, h) X_j(S) M_j(aa)[S_i = h] + \sum_S \lambda(j-i, s, h) X_j(S)[S_i = h]$$
$$+ \sum_{aa} \lambda(j-i, aa, h) M_j(aa)[S_i = h] \tag{4.5}$$

$$v_2(S_{i-1}, S_i, M_j, X_j)$$
$$= \sum_S \sum_{aa} \lambda(j-i, s, aa, h', h'') X_j(S) M_j(aa)[S_{i-1} = h'][S_i = h'']$$
$$+ \sum_S \lambda(j-1, s, h', h'') X_j(S)[S_{i-1} = h'][S_i = h'']$$
$$+ \sum_{aa} \lambda(j-1, aa, h', h'') M_j(aa)[S_{i-1} = h'][S_i = h''] \tag{4.6}$$

The label feature functions model the dependency of backbone angles on protein sequence profiles and predicted secondary structure. Equation 4.5 and 4.6 indicate that not only the state (i.e., angle distribution) itself but also the state transition depend on sequence profiles and predicted secondary structure. As shown in the third and fourth items in the right hand side of Equation 4.2, the state (or state transition) at one position depends on sequence profile and secondary structure in a window of width $2w + 1$ where $w$ is set to 4 in our experiments. It will slightly improve sampling performance by setting the window size larger. Since secondary structure is predicted from sequence profiles, the former is not independent of the latter. Therefore, we need to consider the correlation between sequence profiles and predicted secondary structure, as shown in the first items of the right hand sides of Equation 4.5 and 4.6. The model parameters for the label features are identified by one or two states, secondary structure type, amino acid identity, and the relative position of the observations.

The 2nd-order CRF model has millions of features, each of which corresponds to a model parameter to be trained. Once this model is trained, we can use it to sample protein conformations in a continuous space. Coupled with an energy function and a folding simulation method, we can also use it for template-free modeling.

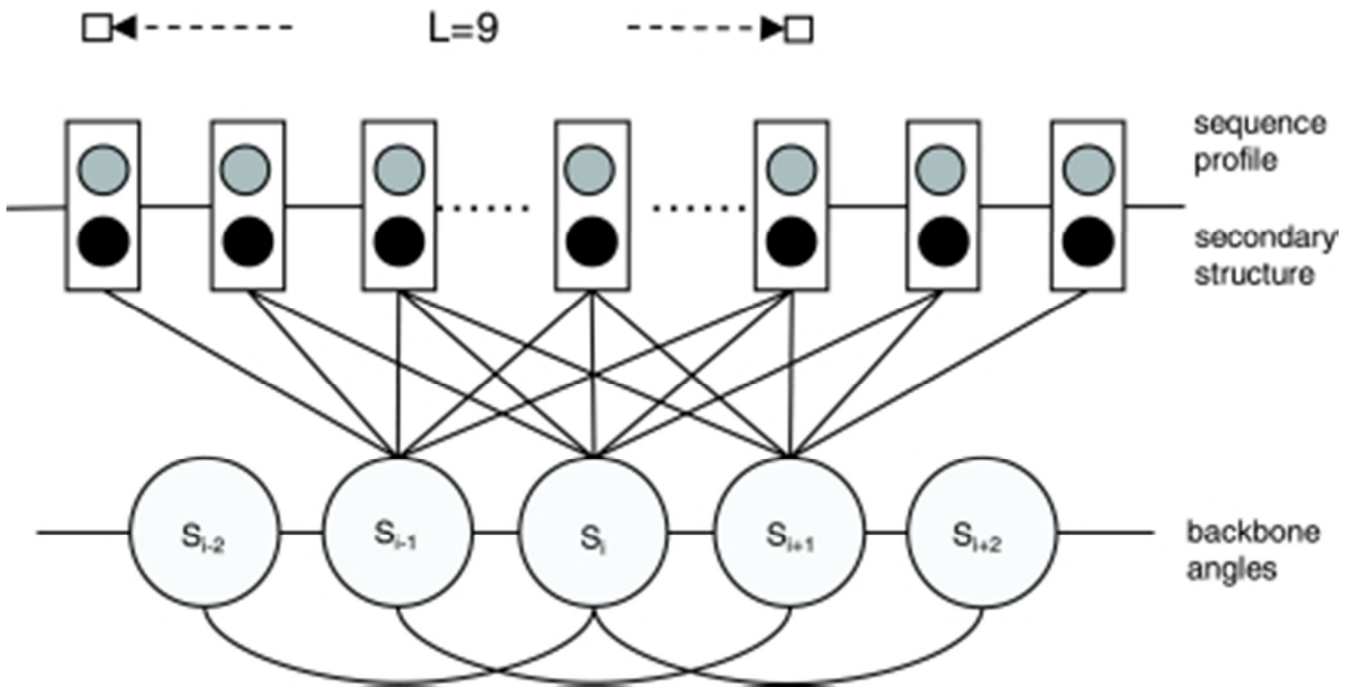

Figure 9  A second-order Conditional Random Fields (CRF) model of protein conformation space. Each backbone angle state depends on a window (size 9) of sequence profiles and secondary structure and also the states in its local neighborhood.

## Model parameter training

Given a set of $m$ proteins with sequence profile $M^i$, predicted secondary structure $X^i$ and corresponding backbone angles $S^i$ $(i = 1,2,...,m)$, our CRF model trains its parameter $\Lambda = \{\lambda_1, \lambda_2, ..., \lambda_p\}$ by maximizing the conditional log-likelihood $L$ of the data:

$$L = \sum_t \log p_\Lambda(s^t|M^t, X^t) - \frac{1}{2\sigma^2}\|\Lambda\|_1 \tag{4.7}$$

The second item in Equation 4.7 is a regularization factor to deal with the sparsity in the training data. When the complexity of the model is high (i.e., the model has many features and parameters) and the training data is sparse, overfitting may occur, and it is possible that many models can fit the training data. Our 2nd-order CRF model has around one million of parameters;, we place a Gaussian prior $\exp\left(-\frac{1}{2\sigma^2}\sum_k \lambda_k\right)$ on the model parameter to choose the model with a "small" parameter in order to avoid overfitting. This regularization can improve the generalization capability of the model in both theory and practice (Vapnik, 1998). This kind of training is also called discriminative training or conditional training. Different from the generative training in the FB5-HMM model, discriminative training directly optimizes the predictive ability of the model while ignoring the generative probability of the observation.

The objective function in Equation 4.7 is convex and hence theoretically a globally optimal solution can be found using any efficient gradient-based optimization technique. There is no analytical solution to the above equation for a real-world application. Quasi-Newton methods such as L-BFGS (Liu and Nocedal, 1989) can be used to solve the above equation and usually can converge to a good solution within a couple of hundred iterations. For a detailed description of how to train a CRF model, see elsewhere (Lafferty et al., 2001; Sha and Pereira, 2003). We revised the FlexCRFs program (Phan et al., 2005) to train our CRF model, and it takes approximately 24 hours to train a single model on a cluster of 150 2GHz CPUs.

We used a set of ~3000 non-redundant proteins to train the parameters in our CRF model. Any two proteins in the training set share no more than 30% sequence identity, and the resolution of a training protein is at least 2.0Å. To avoid overlap between the training data and the test proteins (i.e., the proteins in Table 10 - 12), we removed the following proteins from our training set: (1) the proteins sharing at least 25% sequence identity with our test proteins; (2) the proteins in the same fold class as our test proteins according to the SCOP classification; and (3) the proteins having a TM-score 0.5 with our test proteins in case some recently released proteins do not have a SCOP ID. If the TM-score of two protein structures is smaller than 0.5, then a threading program such as PROSPECTOR_3 cannot identify their similarity relationship with high confidence.

The training set is randomly divided into five sets of same size and then used for five-fold cross validation. We trained our CRF model using three different regularization factors (i.e., $\sigma^2$ in Equation 4.7): 25, 125, and 625, and chose the one with the best F1-value. F1-value is a widely used measurement of the prediction capability of a machine learning model. F1-value is an even combination of precision $p$ and recall $r$ and calculated as $\frac{2pr}{p+r}$. The higher the F1-value is, the better the CRF model. The average F1-values for regularization factors 25, 125, and 625 are 21.73%, 21.55%, and 22.03%, respectively. In terms of F1-value, the difference among these regularization factors is small. Therefore, we choose a small regularization factor 25 to control the model complexity since a model with lower complexity usually generalizes better to the test data. The regularization factor is the only parameter that we need to tune manually. All the other model parameters (i.e., weights for features) can be estimated automatically in the training process.

## Conformation sampling and resampling

**Initial conformation sampling**

Once the CRF model is trained, we can sample a protein conformation or resample the local conformation of a segment by probability using a forward-backward algorithm. We first sample labels (i.e., angle distribution) by probability estimated from our CRF model and then sample real-valued angles from the labels. Let $V(i, S_i, S_{i+1})$ denote the sum of all the edge features associated with edge ($s_i$, $s_{i+1}$) and all the label features associated with labels $S_i$, $S_{i+1}$ and ($S_i$, $S_{i+1}$). Let $G(i, S_i, S_{i+1})$ denote the marginal probability of a label pair ($S_i$, $S_{i+1}$). We can recursively calculate $G(i, S_i, S_{i+1})$ from N-terminal to C-terminal as follows.

$$G(0, S_0, S_1) = e^{V(0, S_0, S_1)}$$

$$G(i, S_i, S_{i+1}) = e^{V(i, S_i, S_{i+1})} \sum_{S_{i-1}} G(i-1, S_{i-1}, S_i) e^{\lambda(S_{i-1}, S_i, S_{i+1})}$$

where $\lambda$ ($s_{i-1}$, $s_i$, $s_{i+1}$) can be interpreted as state transition log-likelihood. Once $G(N-1, s_{N-1}, s_N)$ is calculated where $N$ is the protein size, we can sample a conformation from C-terminal to N-terminal. First, we sample a label pair ($s_{N-1}$, $s_N$) for the last two positions by probability

$$\frac{G(N-1, S_{N-1}, S_N)}{\sum_{S_{N-1}, S_N} G(N-1, S_{N-1}, S_N)}.$$

Then we sample the label $s_i$ for position $i$ by probability

$$\frac{G(i, S_i, S_{i+1})e^{\lambda(S_i, S_{i+1}, S_{i+2})}}{\sum_{S_i} G(i, S_i, S_{i+1})e^{\lambda(S_i, S_{i+1}, S_{i+2})}},$$

supposing that the sampled labels at position $i$+1 and $i$+2 are $s_{i+1}$ and $S_{i+2}$, respectively.

**Conformation resampling**

The algorithm for resampling the local conformation of a randomly chosen segment is similar. We first randomly determine a segment for which we are going to resample backbone angles. Then we resample the angles for this segment using a forward-backward algorithm similar to the initial conformation sampling algorithm. The major difference is that in this scenario we calculate $G(i, S_i, S_{i+1})$ for a segment conditioning on the labels of the two residues flanking this segment at the left and when do resampling we also have to consider the two residues flanking this segment at the right.

**Biased sampling**

Our sampling method works well in alpha regions but not in beta regions. We decided that we should do more frequent sampling in beta and loop regions than in alpha regions. This is because both beta and loop regions are more varied than alpha regions. By sampling in the beta and loop regions more frequently, we can generate decoys with better quality. We achieve this goal by empirically assigning different weights to each position depending on its predicted secondary structure type. The weights for alpha, beta and loop regions are 1, 5, and 3 respectively. These weights are empirically determined using a simple enumeration method on the test proteins in Table 10. To determine which segment with angles to be resampled, we first uniformly sample the segment length $l$ between 1 and 15. Then we sample the starting position of this segment using biased sampling. We calculate the weight of a segment as the sum of the weights of all the positions in this segment. Then we randomly sample a segment with the length $l$ by probability proportional to the weight of this segment.

Biased sampling is employed only when we do folding simulations using the energy function described in this work. In the case that the energy function is not used, we still use uniform sampling.

**Energy function**

The energy function we used for folding simulation consists of three items: DOPE, KMBhbond, and ESP. The weight factors combining these three energy items are trained on the proteins in Table 10 using grid search in a progressive way. First, we fix the weight factor of DOPE to 1 and determine the weight factor for ESP by minimizing the average RMSDs of generated decoys. Then we fix the weight factors of both DOPE and ESP and determine the weight factor for KMBhbond using the same way.

**DOPE**

DOPE is a full-atom, distance-dependent pairwise statistical potential originally designed by Shen and Sali and then improved by the Sosnick group (Fitzgerald et al., 2007; Shen and Sali, 2006). DOPE performs as well or better than many other statistical potentials and force fields in differentiating a native structure from decoys. The statistical potential in DOPE distinguishes the amino acid identity and atomic identity of two interacting particles. In our folding simulation, we only build coordinates for main chain and $C_\beta$ atoms, so only the statistical potentials related to main-chain and $C_\beta$ atoms are used to calculate the energy of a conformation. We denote this revised DOPE as DOPE-$C_\beta$. According to (Fitzgerald et al., 2007), DOPE-$C_\beta$ is highly correlated with the full-atom DOPE. DOPE-$C_\beta$ also performs favorably in applications to intra-basin protein folding (Colubri et al., 2006).

**Hydrogen bonding**

KMBhbond is a statistical potential for hydrogen bonding developed by the Baker group (Morozov et al., 2004). It depends on the distance between the geometric centers of the N–H bond vector and the C=O bond vector, the bond angle between the N–H bond vector and the hydrogen bond, the bond angle between the C=O bond vector and the hydrogen bond, and the dihedral angle about the acceptor-acceptor base bond. The three angles describe the relative orientation of the bond vectors in the hydrogen bond.

**ESP**

ESP is an approximation to the Ooi-Scheraga solvent-accessible surface area (SASA) potential (Li et al., 2008). Since our conformation representation does not contain side-chain atoms, which are necessary for the calculation of the solvent-accessible surface area potential, we employ a simple ESP that assigns each residue with an environmental energy score. ESP is a function of the protein size and the number of $C_\alpha$ atoms contained within an 8.5-Å sphere centered on the residue's $C_\alpha$ atom (Fernandez et al., 2002). Explicitly, the ESP statistical potential has the form given by

$$\text{ESP}(aa, n) = -\ln\frac{P(n|R, aa)}{P(n|R)}$$

where n is the number of $C_\alpha$ atoms in an 8.5-Å sphere centered on the $C_\alpha$ atom of the residue, R is the radius of gyration of the protein, aa is the amino acid identity of the residue, P(n|R) is the number of $C_\alpha$ atoms in an 8.5-Å sphere for a given protein radius regardless of amino acid

identity, and P(n|R,aa) is the number of C$_\alpha$ atoms in an 8.5-Å sphere for a given protein radius and amino acid identity. We calculate ESP(aa, n) from a set of ~3000 non-redundant experimental structures chosen by the PISCES server(Wang and Dunbrack, 2003). Each protein in this set has resolution at least 2.0 Å, R factor no bigger than 0.25, and at least 30 residues. Any two proteins in this set share no more than 30% sequence identity.

To parameterize the ESP potential, we need to discretize the radius of gyration *R*, which ranges from 7Å to 39Å in our training set. We tested the following three discretization schemes: (1) *R* is discretized into 65 bins with equal width 0.5Å; (2) *R* is discretized into 33 bins with equal width 1Å; and (3) *R* is first discretized into 33 bins with equal width 1Å. Then we merge [7, 9), [34, 36) and [37, 39] into a single bin, respectively, to guarantee sufficient statistics for these intervals. We calculated the Pearson correlation coefficient between the resultant ESP energies and TM-score of the decoys. The third scheme yields the best correlation and thus is used in our energy function.

## Energy minimization

We employ a simulated annealing (SA) algorithm to minimize the energy function for a given protein. The SA routine is based on the algorithm proposed by Aarts and Korst in 1991 (Aarts and Korst, 1991). We start with sampling an initial conformation and then search for a better one by minimizing the energy function. Given a conformation, we propose a new conformation by resampling the local conformation of a randomly-chosen small segment using the CRF model. The new conformation is rejected if there are serious steric clashes among atoms; otherwise, it is accepted with probability $\min(1, \rho^{-\Delta E/t})$ where $\Delta E$ is the energy increment and t is the annealing temperature.

The initial annealing temperature is chosen so that at the beginning of the annealing process an energy increase is accepted with a given probability $p_0$ (=0.8). The initial temperature $t_0$ is determined by $t_0 = -\frac{\Delta E}{\ln p_0}$ where $\Delta E$ is the average energy increase. To determine $\Delta E$, we first conduct a series of trial conformation samplings and accept all the generated conformations. Then we estimate $\Delta E$ by calculating the average energy increase observed in our trial samplings.

During the folding simulation process, we decrease the annealing temperature gradually using an exponential cooling schedule. The temperature is updated by $t_{k+1} = 0.9 t_k$. At each annealing temperature, the number of sampled conformations is set to (100+N) where N is the number of residues in the protein. This number is set to achieve thermal equilibrium. The termination of the SA process is triggered when any of the following two conditions is satisfied: (1) either the temperature is low enough such that almost no energy increase is accepted and the annealing process is trapped at local minima; or (2) or the number of conformations generated in a single simulation process reaches a threshold (say 10,000).

## 4.3 RESULTS

### The 2nd-order CRF model is much better than the 1st-order model

We compare our 2nd-order CRF model with the 1st-order model described in Chapter 3 to see how much improvement we can achieve by considering the interdependency among three adjacent residues. To exclude the impact of an energy function in this comparison, we guide conformation search using only compactness and self-avoiding constraints but not an energy function (see Chapter 3 for more details). In total, we tested our models on a set of 22 proteins with different structure properties. We generated ~20,000 decoys for each test protein using each CRF model and then calculated the percentage of decoys with RMSD smaller than a given threshold, as shown in Table 10.

In terms of the best decoys, the 2nd-order model is better on 13 out of 22 test proteins and worse on seven proteins. The best decoys may be generated by chance, so they cannot be reliably used to evaluate the performance of the two CRF models. We further examine their difference in terms of the percentage of decoys with RMSD smaller than a given threshold. A general trend we observed is that when the proteins under consideration are not big (<100 residues), the 2nd-order model generally outperforms the 1st-order model by a large margin. The only exception is 4icbA. The performance difference between these two CRF models is small on relatively large proteins such as 1aa2, 1jer, and T056. This may be because a large protein tends to have a large conformation space and neither CRF model can search a very large conformation space efficiently. The reason that the 2nd-order model performs worse on 4icb is because there is a *cis*-proline in 4icb, and the length of the virtual $C_\alpha$-bond ending at this proline is approximately 3.2Å instead of our assumption 3.8Å. Therefore, the more accurately can our CRF models predict the backbone angles, the more the decoys deviate from the native structure of 4icb. It is not very difficult to resolve this issue since from PSI-BLAST sequence profile we can predict with accuracy 92% if a residue is a *cis*-proline or not (data not shown). This comparison result indicates that we can dramatically improve sampling efficiency by using the 2nd-order CRF model.

Table 10  Quality Comparison of the Decoys Generated by the 1st-Order and 2ND-Order CRF Models

Note: Columns 1–3 list the PDB code, protein size and the type of the test proteins. Columns "best" list the RMSD (Å) of the best decoys; the other columns list the percentage of decoys with RMSD smaller than a given threshold. "O1" and "O2" denote the 1st-order and the 2nd-order CRF models, respectively. In total, ~20,000 decoys are generated for each protein without an energy function.

|  | Size | C |  | Best | ≤6Å | ≤7Å | ≤8Å | ≤9Å | ≤10Å | ≤11Å | ≤12Å |
|---|---|---|---|---|---|---|---|---|---|---|---|
| 1aa2 | 108 | α | O1 | 7.34 | 0 | 0 | 0.035 | 0.245 | 1.05 | 4.27 | 13.5 |
|  |  |  | O2 | 7.31 | 0 | 0 | 0.0145 | 0.116 | 0.97 | 4.99 | 17.7 |
| 1beo | 98 | α | O1 | 6.42 | 0 | 0.02 | 0.2 | 0.99 | 3.27 | 9.7 | 23 |
|  |  |  | O2 | 5.84 | 0.0096 | 0.048 | 0.385 | 1.37 | 4.5 | 12.5 | 29.3 |
| 1ctfA | 68 | αβ | O1 | 3.7 | 2.41 | 7.18 | 16.2 | 31.4 | 51.7 | 77.2 | 95.9 |
|  |  |  | O2 | 3.67 | 6.62 | 22.3 | 47.1 | 64.7 | 78.9 | 94.8 | 99.5 |
| 1dktA | 72 | β | O1 | 6.15 | 0 | 0.1 | 0.87 | 3.81 | 12.5 | 33.4 | 62.8 |
|  |  |  | O2 | 5.07 | 0.121 | 1.48 | 5.94 | 16.2 | 34.3 | 59.3 | 82.6 |
| 1enhA | 54 | α | O1 | 2.32 | 22.4 | 32.4 | 44.7 | 61.2 | 85.4 | 98.5 | 100 |
|  |  |  | O2 | 2.21 | 69.3 | 72 | 77.7 | 87.1 | 97.4 | 99.9 | 100 |
| 1fc2C | 43 | α | O1 | 1.94 | 49.1 | 64.1 | 85 | 97.6 | 99.7 | 100 | 100 |
|  |  |  | O2 | 2.28 | 85.4 | 91.7 | 97.3 | 99.8 | 100 | 100 | 100 |
| 1fca | 55 | β | O1 | 4.99 | 0.145 | 1.3 | 6 | 19.9 | 49.7 | 85.5 | 99.3 |
|  |  |  | O2 | 4.96 | 0.207 | 2.65 | 13.5 | 36.3 | 68.2 | 94 | 99.8 |
| 1fgp | 67 | β | O1 | 7.4 | 0 | 0 | 0.035 | 0.46 | 4.21 | 20.2 | 54.2 |
|  |  |  | O2 | 5.94 | 0.0048 | 0.043 | 0.582 | 4.25 | 18.2 | 48.2 | 81.8 |
| 1jer | 110 | β | O1 | 9.64 | 0 | 0 | 0 | 0 | 0.005 | 0.12 | 0.91 |
|  |  |  | O2 | 10.2 | 0 | 0 | 0 | 0 | 0 | 0.185 | 1.11 |
| 1nkl | 78 | α | O1 | 3.64 | 5.91 | 14.1 | 25.1 | 45.2 | 66.6 | 86.8 | 97.7 |
|  |  |  | O2 | 3.06 | 20.3 | 30.1 | 44.2 | 65.3 | 84 | 96.7 | 99.8 |
| 1pgb | 56 | αβ | O1 | 3.15 | 22.3 | 45 | 65.1 | 81 | 93 | 98.6 | 99.9 |
|  |  |  | O2 | 2.6 | 63.2 | 85.8 | 93.9 | 97.7 | 99.5 | 99.9 | 100 |
| 1sro | 76 | β | O1 | 6.22 | 0 | 0.07 | 0.525 | 2.75 | 10 | 27.1 | 54.7 |
|  |  |  | O2 | 5.39 | 0.0193 | 0.289 | 1.54 | 5.09 | 14.5 | 32.9 | 60 |
| 1trlA | 62 | α | O1 | 3.53 | 13.5 | 25.3 | 38.5 | 57.4 | 85.1 | 99.1 | 99.9 |
|  |  |  | O2 | 3.72 | 34.1 | 45.3 | 53.5 | 68 | 94.8 | 100 | 100 |
| 2croA | 65 | α | O1 | 2.8 | 16.8 | 31.4 | 47.9 | 63.3 | 79.9 | 93 | 99.6 |

|  |  |  | O2 | 2.58 | 35.4 | 52.9 | 67.4 | 79.3 | 88.9 | 95.6 | 99.9 |
| --- | --- | --- | --- | --- | --- | --- | --- | --- | --- | --- | --- |
| 2gb1A | 56 | β | O1 | 2.91 | 23.3 | 45.6 | 65.8 | 81.8 | 93.2 | 98.8 | 99.9 |
|  |  |  | O2 | 2.04 | 65.2 | 86.1 | 93.8 | 97.8 | 99.5 | 100 | 100 |
| 4icbA | 76 | α | O1 | 4.63 | 0.515 | 2.65 | 7.64 | 16.8 | 33.6 | 59.1 | 84.3 |
|  |  |  | O2 | 4.4 | 0.125 | 0.6 | 2.85 | 12.3 | 31 | 58.9 | 88.5 |
| T052 | 98 | β | O1 | 7.58 | 0 | 0 | 0.01 | 0.035 | 0.135 | 0.8 | 3.52 |
|  |  |  | O2 | 8.37 | 0 | 0 | 0 | 0.025 | 0.296 | 1.78 | 6.58 |
| T056 | 114 | α | O1 | 7.78 | 0 | 0 | 0.0198 | 0.084 | 0.51 | 2.26 | 7.07 |
|  |  |  | O2 | 7.57 | 0 | 0 | 0.005 | 0.094 | 1.07 | 3.83 | 8.38 |
| T059 | 71 | β | O1 | 6.3 | 0 | 0.01 | 0.135 | 1.14 | 7.2 | 26.8 | 61.1 |
|  |  |  | O2 | 6.21 | 0 | 0.025 | 0.421 | 3.85 | 17.4 | 45.1 | 77.3 |
| T061 | 76 | α | O1 | 5.36 | 0.01 | 0.37 | 2.89 | 10.7 | 27 | 50.9 | 77.6 |
|  |  |  | O2 | 6.04 | 0 | 0.282 | 4.73 | 19.9 | 40.5 | 62.7 | 82.6 |
| T064 | 103 | α | O1 | 7.23 | 0 | 0 | 0.035 | 0.2 | 0.91 | 2.79 | 7.3 |
|  |  |  | O2 | 7.47 | 0 | 0 | 0.032 | 0.412 | 1.53 | 3.62 | 9.05 |
| T074 | 98 | α | O1 | 4.86 | 0.015 | 0.235 | 1.23 | 4 | 10.4 | 22 | 41 |
|  |  |  | O2 | 4.22 | 0.098 | 0.835 | 3.56 | 9.62 | 18.7 | 30.4 | 49.3 |

## Comparison with FB5-HMM and fragment assembly

We further compare our two CRF models with the FB5-HMM model in (Hamelryck et al., 2006), as shown in Table 11. Here we compare FB5-HMM and our CRF models using PSIPRED-predicted secondary structure and sequence information as their input. For each test protein, FB5-HMM generates 100,000 decoys, and we generated only ~20,000 decoys. As shown in Table 11, our CRF models can generate decoys with significantly better quality than FB5-HMM on five out of six proteins tested in the FB5-HMM article. The only exception is 4icb, which has been explained in the above section. The result of FB5-HMM in Table 11 is taken from (Hamelryck et al., 2006). The significant improvement of the $2^{nd}$-order CRF models over the FB5-HMM model lies in that in estimating the probability of the angles at one residue, the $2^{nd}$-order CRF model can directly take into consideration the effects of its neighbor residues. Our $2^{nd}$-order CRF also models the relationship among three adjacent residues. By contrast, FB5-HMM only takes into consideration the relationship between two adjacent residues. Furthermore, FB5-HMM does not consider the effects of the neighbor residues when estimate the probability of angles at one residue.

We also compare our $2^{nd}$-order CRF model with the fragment assembly method without using energy function. We revised the Rosetta code to do conformation optimization using the compactness and self-avoiding constraints instead of the Rosetta energy function. As shown in Table 11, the advantage of our $1^{st}$-order CRF model over Rosetta is not obvious. However, our $2^{nd}$-order model can generate a much larger percentage of good decoys than Rosetta for five out of six proteins. The protein 4icb is an exception, which has been explained in previous sections. This comparison result further indicates that it is essential to use the $2^{nd}$-order model instead of the $1^{st}$-order model for template-free modeling. In terms of the quality of the best decoys, Rosetta is slightly better. One of the major differences between these two methods is that our CRF model uses a more simplified representation of protein conformation than Rosetta. That is, we use the pseudo backbone angles to represent a protein conformation while Rosetta uses the true backbone angles (i.e., phi/psi). The phi/psi representation has almost twice the degree of freedom as that of the pseudo backbone angle representation. This may explain why our method tends to generate more decoys with RMSD smaller than 6 Å and the best decoys generated by Rosetta tend to have smaller RMSD.

Table 11  Quality Comparison of the Decoys Generated by FB5-HMM, the 1st-Order CRF, the 2nd-Order CRF, and Rosetta

For each protein, 100,000 decoys are generated by FB5-HMM, while only ~20,000 decoys by each CRF model and Rosetta. No energy function is used in this comparison. Columns 1–3 list name and PDB code, size and number of $\alpha$-helices, and $\beta$-strands of the test proteins. Columns "Good" and "Best" list the percentage of good decoys (with RMSD ≤ 6 Å) and the RMSD of the best decoys, respectively.

| Test proteins | | | FB5-HMM | | 1st-order CRF | | 2nd-order CRF | | Rosetta | |
|---|---|---|---|---|---|---|---|---|---|---|
| PDB | L | α, β | Good | Best | Good | Best | Good | Best | Good | Best |
| 1FC2 | 43 | 2,0 | 17.1 | 2.6 | 49.1 | 1.94 | 85.4 | 2.28 | 36.5 | 2.72 |
| 1ENH | 54 | 2,0 | 12.2 | 3.8 | 22.4 | 2.32 | 69.3 | 2.21 | 44.8 | 1.23 |
| 2GB1 | 56 | 1,4 | 0 | 5.9 | 23.3 | 2.91 | 65.2 | 2.04 | 5.82 | 2.26 |
| 2CRO | 65 | 5,0 | 1.1 | 4.1 | 16.8 | 2.79 | 35.4 | 2.58 | 17.2 | 2.38 |
| 1CTF | 68 | 3,1 | 0.35 | 4.1 | 2.4 | 3.7 | 6.62 | 3.67 | 2.35 | 1.3 |
| 4ICB | 76 | 4,0 | 0.38 | 4.5 | 0.51 | 4.63 | 0.125 | 4.4 | 4.51 | 3.9 |

## Comparison with lattice model

By combining a simple energy function and our 2nd-order CRF model, we build a program, denoted as CRFFolder, for template-free modeling. We compare CRFFolder with TOUCHSTONE-II, a representative lattice model method developed by Skolnick group. TOUCHSTONE-II is an excellent template-free modeling program, and its two derivatives TASSER (Zhang and Skolnick, 2007) and I-TASSER (Wu et al., 2007a) perform very well in both CASP7 and CASP8. We do not compare CRFFolder with the two derivatives because both TASSER and I-TASSER use threading-generated constraints to guide conformation search, while CRFFolder does not. Due to the limitations of computational power, we tested CRFFolder on a set of 15 test proteins with various structure properties, which were also tested by TOUCHSTONE-II. These test proteins have very different secondary structures and sizes ranging from 47 to 157. We generated approximately 3000 decoys for each alpha protein, 7000 decoys for each alpha-beta protein, and 10,000 decoys for each beta protein. By contrast, TOUCHSTONE-II used a complex energy function consisting of 21 items and generated 24,000 decoys for each test protein (Zhang et al., 2003). As shown in Table 12, CRFFolder performs much better than TOUCHSTONE-II on all the alpha proteins except one. CRFFolder also has comparable performance on beta and alpha-beta proteins. On larger proteins, CRFFolder is slightly worse than TOUCHSTONE-II. This may be because the replica exchange Monte Carlo algorithm used by TOUCHSTONE-II for energy minimization is better than the simulated annealing algorithm used in CRFFolder. Note that since two programs use very different clustering methods, it is not easy to compare these two programs fairly. TOUCHSTONE-II used a program SCAR to do decoy clustering while we use MaxCluster (http://www.sbg.bio.ic.ac.uk/maxcluster/). For the purpose of comparison, we also show the RMSD of the best decoys and the average RMSDs of the top 1% and 2% decoys generated by CRFFolder.

## Performance in the blind CASP8 evaluation

We tested the performance of our method by participating in the blind CASP8 evaluation with RAPTOR++, a new protein structure prediction method. Our 2nd-order CRF model was trained before CASP8 started (in May 2008), so it is unlikely for us to overfit our model for the CASP8 targets.

### Multi-domain proteins

Some of the CASP target proteins are large and contain multiple domains. For those targets, we parse them into several possible domains by searching through the Pfam database (Finn et al., 2008) using HMMER (Eddy, 1998; Krogh et al., 1994). If one target protein can be aligned to a single template, then domain parsing is skipped. In the case that there is a big chunk of the target not aligned to any top templates, we will treat this unaligned chunk as a single target and do protein modeling separately. Except the last several CASP8 targets, the models for multiple domains are not assembled into a single coordinate system.

**Comparison with Robetta**

We first examine the performance of our method by comparing it with Baker's Robetta server on some CASP8 hard targets, on which both Robetta and CRFFolder did template-free modeling before their experimental structures were released. These hard targets have no good templates in the PDB. It is unclear how many decoys Robetta generated for each target, but the top five models generated by Robetta for each target are available for download from the official website of CASP8 (http://predictioncenter.org/download_area/CASP8/server_predictions/). Using our template-free modeling program CRFFolder, we generated ~7000 decoys for each target and then chose the top five models. Note that the first models chosen by CRFFolder are not exactly the same as our CASP8 submissions since we submitted template-based models for some of these targets.

Table 13 compares CRFFolder and Robetta in terms of the quality of the first-ranked models. The model quality is evaluated by a program TM-score (Zhang and Skolnick, 2005b), which generates a real number between 0 and 1 to indicate the quality of a structure model. Roughly, the higher the TM-score is, the better the model quality. Note that in this table the domain definition of T0510_D3 is from Zhang's CASP8 assessment page (http://zhanglab.ccmb.med.umich.edu/casp8/), while others are from Robetta CASP8 web site.

Table 12  Performance Comparison between CRFFolder and Skolnick's TOUCHSTONE-II

Note: Columns 1–3 list the PDB code, size, and type of the test proteins. Column "Best Cluster" lists the RMSDs of the representative decoys of the best clusters. Column "best" lists the RMSDs of the best decoys. The first number in parentheses denotes the rank of the best cluster and the second is the total number of clusters. Columns "1%" and "2%" list the average RMSDs of the top 1% and 2% decoys, respectively. The results of TOUCHSTONE-II are from (Zhang et al., 2003).

|  |  |  | *TOUCHSTONE* | *CRFFolder* |  |  |  |
|---|---|---|---|---|---|---|---|
| *Target* | *Size* | *Class* | *BestCluster* | *BestCluster* | *Best* | *1%* | *2%* |
| 1bw6A | 56 | α | 4.79 (2/3) | **3.82 (3/3)** | 2.75 | 3.38 | 3.54 |
| 1lea | 72 | α | 5.69 (5/5) | **4.10 (5/7)** | 3.41 | 4.19 | 4.48 |
| 2af8 | 86 | α | 11.07 (5/6) | **8.9 (12/19)** | 7.07 | 8.53 | 8.97 |
| 256bA | 106 | α | 3.61 (2/3) | **2.75 (6/11)** | 2.5 | 3.45 | 3.7 |
| 1sra | 151 | α | 10.71 (3/12) | 13.95 (17/25) | 10.82 | 13.8 | 14.24 |
| 1gpt | 47 | α β | 6.30 (1/25) | **5.55 (42/67)** | 4.34 | 5.2 | 5.47 |
| 1kp6A | 79 | α β | 10.01 (8/14) | **7.99 (1/7)** | 6.29 | 7.51 | 7.81 |
| 1poh | 85 | α β | 9.10 (5/9) | **8.84 (5/10)** | 7.49 | 8.7 | 9.04 |
| 1npsA | 88 | α β | **6.89 (33/34)** | 9.91 (41/57) | 7.87 | 9.19 | 9.66 |
| 1t1dA | 100 | α β | **8.96 (7/13)** | 9.22 (10/13) | 6.51 | 9.51 | 9.94 |
| 1msi | 66 | β | 7.72 (19/28) | 7.77 (12/15) | 6.24 | 7.55 | 7.89 |
| 1hoe | 74 | β | **9.39 (5/13)** | 9.87 (16/35) | 7.96 | 10 | 10.37 |
| 1ezgA | 82 | β | 11.03 (40/44) | **10.42 (42/66)** | 9.66 | 10.4 | 10.62 |
| 1sfp | 111 | β | **7.48 (2/18)** | 11.07 (5/11) | 9.32 | 11.1 | 11.59 |
| 1b2pA | 119 | β | 12.52 (31/56) | **10.01 (18/25)** | 8.76 | 10.9 | 11.32 |

As shown in Table 13, overall CRFFolder is better than Robetta by ~8%. Compared to the Robetta server, our method performs very well on mainly alpha proteins, e.g., T0460, T0496_D1 and T0496_D2. This could be expected since our CRF model can capture well the local sequence-structure relationship and alpha helices are stabilized by local interactions between neighbor residues. Our method also works well on small, mainly beta proteins. For example, our method is better than Robetta on two small beta proteins T0480 and T0510_D3. (Please note that T0480 is evaluated without removing the disordered regions at the two ends. If the disorder regions are removed, CRFFolder is still better than Robetta by about 12 in GDT score.) However, our method does not perform very well on some relatively large proteins (>100 residues) with a few beta strands, e.g., T0482 and T0513_D2. This is probably because our CRF method can only model local sequence-structure relationship while a beta sheet is stabilized by non-local hydrogen bonding. For a small beta protein, our method can search more thoroughly the conformation space by sampling in a continuous space and potentially do better. However, for a large beta-containing protein, the search space is too big to be explored in a continuous space. Another possible reason is that our energy function is not as good as the Robetta energy function in guiding the formation of beta-sheets.

It is also worth noting that compared to Robetta, our method did better on the first domain of T0496, a mainly-alpha protein with120 residues. According to the study in (Shi et al., 2009), this domain target is one of the only two CASP8 targets with really new folds. Our method did as well as Robetta on another target with new fold (i.e., T0397_D1).

**Comparison with template-based methods**

Our program CRFFolder can also generate 3D models better than template-based methods for a couple of hard CASP8 targets. According to the CASP8 official assessment, if only the first-ranked models are evaluated, CRFFolder produced the best model among all the CASP8 human and server groups for T0510_D3, a small alpha/beta protein with 43 residues (http://predictioncenter.org/casp8/results.cgi). T0510_D3 is treated as a free-modeling target by CASP8 while Grishin et al. classified it as a fold recognition target (Shi et al., 2009). We also examined all the template-based models generated by our threading methods in CASP8 for this target. The best template-based model has TM-score of 0.339.

Table 13  Performance Comparison (TM-Score) between CRFFolder and Robetta on Some CASP8 Hard Targets

| Target | Size | Class | Robetta | CRFFolder |
|---|---|---|---|---|
| T0397_D1 | 70 | $\alpha\beta$ | 0.25 | 0.258 |
| T0460 | 111 | $\alpha\beta$ | 0.262 | **0.308** |
| T0465 | 157 | $\alpha\beta$ | 0.243 | 0.253 |
| T0466 | 128 | $\beta$ | **0.326** | 0.217 |
| T0467 | 97 | $\beta$ | 0.303 | **0.364** |
| T0468 | 109 | $\alpha\beta$ | 0.253 | **0.308** |
| T0476 | 108 | $\alpha\beta$ | 0.279 | 0.25 |
| T0480 | 55 | $\beta$ | 0.208 | **0.307** |
| T0482 | 120 | $\alpha\beta$ | **0.352** | 0.223 |
| T0484 | 62 | $\alpha$ | 0.253 | 0.249 |
| T0495_D2 | 65 | $\alpha\beta$ | 0.312 | **0.436** |
| T0496_D1 | 110 | $\alpha\beta$ | 0.235 | **0.293** |
| T0496_D2 | 68 | $\alpha$ | 0.291 | **0.5** |
| T0510_D3 | 43 | $\alpha\beta$ | 0.147 | **0.352** |
| T0513_D2 | 77 | $\alpha\beta$ | **0.581** | 0.367 |
| T0514 | 145 | $\alpha\beta$ | 0.283 | 0.277 |
| Average | | | 0.286 | 0.31 |

CRFFolder also produced one of the best models for T0496_D1, better than other template-based models. Both CASP8 and Grishin et al. classified T0496_D1 as a free-modeling target (Shi et al., 2009). In fact, the first-ranked model we submitted for T0496_D1 is much worse than the best decoy we generated for this target. Among the ~7000 decoys we generated, there are around 18% decoys with TM-score better than the first-ranked model. The best decoy has TM-score 0.475 and RMSD to native 6.592Å. By contrast, the first-ranked template-free model has TM-score 0.293 and RMSD 11.457Å. The best template-based model generated by our threading methods in CASP8 for this has TM-score 0.251, and RMSD 15.372Å.

We also examined all the template-based models generated by our threading methods in CASP8 for T0397_D1, another target with really new fold (Shi et al., 2009). There are only six template-based models with TM-score higher than the first-ranked template-free models generated by CRFFolder. The best template-based model generated by us in CASP8 has a TM-score of 0.338, while the best template-free model generated by CRFFolder has a TM-score of 0.364. There are 6.6% decoys generated by CRFFolder have have a TM-score better than our first-ranked template-free model.

Table 14 summarizes the results of RAPTOR++ in CASP8 free modeling targets. CASP8 defined 164 effective domains and classified them into three categories while Grishin et al (http://prodata.swmed.edu/CASP8/evaluation/DomainsAll.First.html) defined 146 domains and classified them into five categories. As shown in Columns 3-5, the best models generated by RAPTOR++ for FM targets are much better than the first models. This indicates that we still need to improve our model selection method for FM targets. As shown in Columns 5-7, for FM targets, the best models submitted by all CASP8 servers are much better than the best generated by RAPTOR++. This means that in addition to improve model selection, we also need to further improve our model generation method for FM targets. We can have similar observations when Grishin's domain definition and classification is used.

Our template-free modeling method samples protein conformations in a continuous space without using fragments in the PDB. Our method aims to overcome two major issues with current popular fragment assembly and lattice model methods. These two methods may exclude native structure from search space by sampling in a discrete space since a small change in a backbone angle can result in a totally different fold or by assembling a protein structure using even medium-sized fragments since a "new fold" seems to be composed of rarely occurring super-secondary structure motifs (Andras Fisher, CASP8 talk).

Compared to the Robetta server (see Table 13), our method performs very well on mainly-alpha proteins, e.g., T0460, T0496_D1 and T0496_D2. This is not surprising since our CRF model can capture well the local sequence-structure relationship. Our method also works well on small mainly-beta proteins. For example, our method is better than Robetta on T0480 and T0510_D3. However, our method does not fare well on a relatively large protein (>100 residues) with a few beta strands, e.g., T0482 and T0513_D2. This is probably because our CRF method can only model local sequence-structure relationship while a beta sheet is stabilized by non-local hydrogen bonding. Although sampling in a continuous space, our method can still efficiently search the conformation space of a small beta protein. However, for a large protein with a few beta sheets, the search space is too big to be explored by our continuous conformation sampling algorithm. It is also worth to note that compared to Robetta, our method works well on T0397_D1 and T0496_D1, which, according to Nick Grishin, are the only two CASP8 targets with really new folds.

Table 14 Summarized results of RAPTOR++ predictions of free modeling targets in CASP8.

Note: The upper half table contains the results of 13 CASP8 official domains and the lower half contains the results of 35 domains by Grishin's definition (http://prodata.swmed.edu/CASP8/evaluation/).

R1: GDT-TS score sum of the first-ranked models by RAPTOR.

RB: GDT-TS score sum of the best models submitted by RAPTOR.

RBAll: GDT-TS score sum of the best models generated by RAPTOR.

S1: GDT-TS score sum of the best first models submitted by all servers.

SB: GDT-TS score sum of the best models submitted by all servers.

**FR:** Top 10 First-GDT_TS avg ≥ 30

**FM:** Top 10 First-GDT_TS avg < 30

|  | Category(#) | R1 | RB | RBAll | S1 | SB |
|---|---|---|---|---|---|---|
| **CASP8 Official** | FM (13) | 393.20 | 459.24 | 511.74 | 591.34 | 646.47 |
| **Grishin's Definition** | FR (30) | 997.41 | 1125.73 | 1242.65 | 1386.72 | 1456.00 |
|  | FM (5) | 106.46 | 117.58 | 131.59 | 157.11 | 168.15 |

## 4.4 Conclusion

This chapter has presented a probabilistic and continuous model of protein conformational space for template-free modeling. By using the $2^{nd}$-order CRF model and directional statistics, we can accurately describe protein sequence-angle relationship and explore a continuous conformation space by probability, without worrying about that the native fold is excluded from our conformation space. This method overcomes the following limitations of the fragment assembly method: (1) fragment assembly samples conformations in a discrete space; and (2) fragment assembly is not really template free since it still uses short fragments (e.g., 9-mer) extracted from the PDB. Both restrictions may cause loss of prediction accuracy.

Even though we use a simple energy function to guide conformation search, our probabilistic model enables us to do template-free modeling as well as two well-developed programs TOUCHSTONE-II and Robetta. Both of them have been developed for many years and have well-tuned and sophisticated energy functions. Our template-free modeling is much better than TOUCHSTONE-II on alpha proteins and has similar performance on mainly beta proteins. Blindly tested on some CASP8 hard targets, our method is also better than the Robetta server on quite a few (mainly alpha and small beta) proteins but worse on some relatively large beta-containing proteins. Finally, our method also generated the 3D models for a couple of CASP8 targets (i.e., one mainly-alpha target T0496_D1 and one small alpha/beta target T0510_D3) better than template-based methods. The good performance on alpha proteins indicates that our $2^{nd}$-order CRF model can capture well the local sequence-structure relationship for alpha proteins. The good performance on small beta proteins indicates that by sampling in a continuous space we can explore the conformational space of small beta proteins more thoroughly. To improve the performance of our template-free modeling on relatively large beta-containing proteins, we need to further improve our probabilistic model of beta regions and develop a better hydrogen-bonding energy item for the formation of beta sheets.

# Chapter 5. CNF-Folder: Modeling with Nonlinear Features

## 5.1 Introduction

This chapter presents a new probabilistic graphical model Conditional Neural Fields (CNF) for ab initio protein folding. Please see the work of (Peng et al., 2009) for a detailed exposition. CNF is similar to but much more powerful than CRF on modeling the sequential data in that CNF can naturally model the nonlinear relationship between input and output while CRF cannot do so. Thus, CNF can model better the sophisticated relationship between backbone angles, sequence profile and predicted secondary structure, estimate the probability distribution of backbone angles more accurately and sample protein conformations more efficiently.

In addition, this work also differs from those shown in Chapter 3&4 in that (1) We developed a Replica Exchange Monte Carlo (REMC) method for folding simulation instead of using a simulated annealing (SA) method; (2) This work will use the position-specific scoring matrix (PSSM) generated by PSI-BLAST as the input of our CNF model instead of using the position-specific frequency matrix (PSFM) generated by PSI-BLAST as the input. The REMC method enables us to minimize energy function to a lower level and thus possibly produce better decoys. Also, it has been proved that PSSM contains more information than PSFM for structure prediction such as secondary structure prediction. We did not use PSSM with CRF because CRF cannot easily take PSSM as input. By contrast, we can easily feed PSSM into our CNF model. We will show that our new method is much more effective than our previous method and can dramatically improve sampling efficiency and we can generate much better decoys than before on a variety of test proteins.

## 5.2 Methods

### A 2nd-order CNF model of conformation space

The Conditional Random Fields (CRF) methods developed earlier for protein conformation sampling use a linear combination of input features (i.e., PSI-BLAST sequence profile and predicted secondary structure) to estimate the probability distribution of backbone angles. This kind of linear parameterization implicitly assumes that all the features are linearly independent, which contradicts with the fact that some input features are highly correlated.

For example, the predicted secondary structure is correlated with sequence profiles since the former is usually predicted from the latter using tools such as PSIPRED (Jones, 1999). To model the correlation between predicted secondary structure and sequence profiles, an easy way is to explicitly enumerate all the possible combinations of secondary structure type and amino acid identity in the linear CRF model. In fact, we can always combine some basic features to form a complex feature. However, explicitly defining complex features may introduce a number of serious issues. First, it will result in a combinatorial explosion in the number of complex features, and hence, in the model complexity. It is challenging to train a model with a huge number of parameters without overfitting. Second, explicit enumeration may miss some important complex features. For example, the CRF model presented in Chapter 3&4 (Zhao et al., 2008; Zhao et al., 2009) does not accurately model the correlation among sequence information at several adjacent positions. Finally, explicit enumeration of complex features may also introduce a large number of unnecessary features, which will increase the running time of probability estimation.

Instead of explicitly enumerating all the possible nonlinear combinations of the basic sequence and structure features, we can use a better graphical model to implicitly account for the nonlinear relationship between sequence and structure. Very recently, we have developed a new probabilistic graphical model Conditional Neural Fields (CNF) (Peng et al., 2009), which can implicitly model nonlinear relationship between input and output. As shown in Figure 10, CNF consists of at least three layers: one or more hidden layers, input (i.e., sequence profile and secondary structure) and output (i.e., backbone angles) while CRF consists of only two layers: input and output. The relationship between the backbone angles and the hidden layer is still linear. However, the hidden layer uses some gate functions to nonlinearly transform the input features into complex features. Here we use $G_\theta(x) = \frac{1}{1+\exp(-\theta^T x)}$ as the gate function where $\theta$ is the parameter vector and $x$ a feature vector. CNF can also be viewed as the seamless integration of CRF and neural networks (NN). The neurons in the hidden layer will automatically extract nonlinear relationship among input features. Therefore, without explicit enumeration, CNF can directly model nonlinear relationship between input and output. The training of a CNF model is similar to that of a CRF, but more complicated.

We have tested this CNF model for protein secondary structure (SS) prediction from sequence profiles. Table 15 compares the performance of various machine learning methods for SS prediction. The results are averaged on a 7-fold cross-validation on the CB513 data set, except that SPINE uses 10-fold cross-validation. As shown in Table 15, by using only one hidden layer to model nonlinear relationship between output and input, CNF achieves almost 10% relative improvement over CRF. CNF also outperforms other methods including SVMpro (Hua and Sun, 2001), SVMpsi (Kim and Park, 2003), YASSPP (Karypis, 2006), PSIPRED (Jones, 1999), SPINE (Dor and Zhou, 2007) and TreeCRFpsi (Dietterich et al., 2004). The linear CRF is the worst since it does not model nonlinear relationship between secondary structure and sequence profile. This result indicates that we can indeed benefit from modeling nonlinear sequence-structure relationship. We expect that using CNF, we are able to more accurately model sequence-angle relationship and thus, to sample conformations more efficiently.

In the context of CNF, the PSI-BLAST sequence profile (i.e., position-specific scoring matrix) and predicted secondary structure are viewed as observations; the backbone angles and their FB5 distributions are treated as hidden states or labels. Similar to what we have discussed in Chapter 4, let H denote the 100 groups (i.e., states or labels) generated from clustering of the backbone angles. Each group is described by an FB5 distribution. Given a protein with solved structure, we calculate its backbone angles at each position and determine one of the 100 groups (i.e., states or labels) to which the angles at each position belong. Let $S = \{S_1, S_2, \ldots, S_N\}(S_i \in H)$ denote such a sequence of states/labels (i.e., FB5 distributions) for this protein. We also denote the sequence profile of this protein as $M$ and its secondary structure as $X$. As shown in Figure 10, our CNF model defines the conditional probability of $S$ given $M$ and $X$ in the same form as in Equation (4.1) as follows.

$$P_\Lambda(S|M,X) = \frac{\exp(\sum_{i=1}^{N} F(S,M,X,i))}{Z(M,X)}$$

where $\Lambda = (\lambda_1, \lambda_2, \ldots, \lambda_p,)$ is the model parameter and $Z(M,X) = \sum_S \exp(\sum_{i=1}^{N} F(S,M,X,i))$ is a normalization factor summing over all the possible labels for the given M and X. $F(S,M,X,i)$ consists of two edge feature functions and one label feature function at position *i*. It is given by

$$F(S,M,X,i) = e_1(S_{i-1}, S_i) + e_2(S_{i-1}, S_i, S_{i+1}) + \sum_{j=i-w}^{i+w} v(S_{i-1}, S_i, M_j, X_j) \qquad (5.1)$$

where $e_1(S_{i-1}, S_i)$ and $e_2(S_{i-1}, S_i, S_{i+1})$ are the 1st-order and 2nd-order edge feature functions, respectively, and $v(S_{i-1}, S_i, M_j, X_j)$ is the label feature function. The edge functions describe the interdependency between two or three neighboring labels. CNF is different from CRF in the label feature function. In CRF, the label feature function is defined as a linear combination of features. In CNF, there is an extra hidden layer between the input and output, which consists of *K* gate functions (see Figure 10). The *K* gate functions extract a *K*-dimensional implicit nonlinear representation of input features. Therefore, CNF can be viewed as a CRF with its inputs being *K* homogeneous hidden feature-extractors at each position.

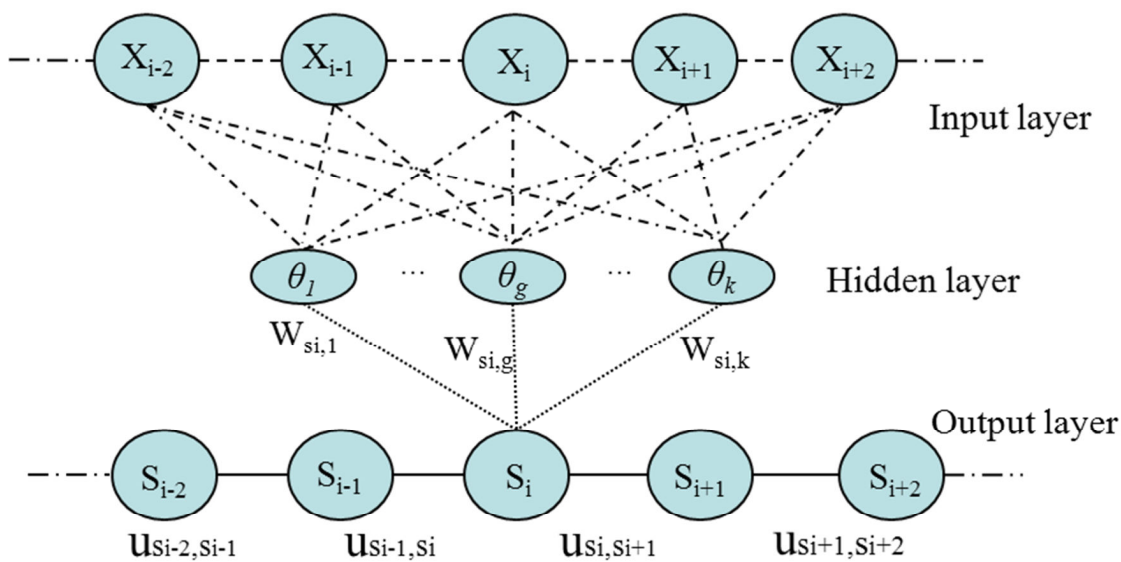

Figure 10 A 1st-order CNF model consists of three layers: input, output and hidden layer. A 2nd-order model is similar but not shown for the purpose of simplicity. By contrast, a CRF model consists of only input and output.

Table 15 Secondary structure prediction accuracy comparison with the current top predictors.

Note : Q3 denotes the Percentage of correct secondary structure;

| Methods | Q3 (%) | Methods | Q3 (%) |
|---|---|---|---|
| CRF | 72.3 | **CNF** | **80.1** |
| TreeCRFpsi | 77.6 | YASSPP | 77.8 |
| SVMpro | 73.5 | PSIPRED | 76.0 |
| SVMpsi | 76.6 | SPINE | 76.8 |

The label feature function of CNF is defined as follows.

$$v(S_{i-1}, S_i, M, X) = \sum_{g=1}^{X} w_{S_{i-1}, S_i, g} G_{\theta_g}(f(M, X, i)) \qquad (5.2)$$

That is, the label feature function is a linear combination of $K$ gate functions $G$. In the above definition, $w$ is the parameter vector and $f$ is a vector of basic features at position $i$. In our current implementation, $f$ contains 23×9 (=207) elements, corresponding to the sequence profile and secondary structure information in a window of size 9 centered at position $i$. We use PSIPRED to predict the secondary structure of a protein from its sequence profile. PSIPRED generates likelihood score of three secondary structure types for each residue, which is used as the input of our CNF model.

Similar to CRF, we use the maximum likelihood method to train the model parameters such that $P_\Lambda(S|M,X)$ is maximized. That is, we maximize the occurring probability of a set of ~3000 non-redundant high-resolution protein structures. Although both the output and hidden layers contain model parameters, all the parameters can be learned together by gradient-based optimization. We use LBFGS (Liu and Nocedal, 1989) as the optimization routine to search for the optimal model parameters. Since CNF contains a hidden layer of gate functions $G$, the log-likelihood function is not convex any more. Therefore, it is very likely that we can only obtain a local optimal solution of the model parameters. To achieve a good solution, we run the training algorithm several times and use the solution with the best objective function as the final solution of the model. See (Peng et al., 2009) for a detailed description of training CNF.

**Model parameter training**

To do a fair comparison between our previous CRF model and this CNF model, we used exactly same data to train both CRF and CNF models. That is, we use a set of ~3000 non-redundant proteins to train the parameters in our CNF and CRF models. Any two proteins in the training set share no more than 30% sequence identity and the resolution of a training protein is at least 2.0Å. To avoid overlap between the training data and the test proteins, we removed the following proteins from our training set: 1) the proteins sharing at least 25% sequence identity with our test proteins; 2) the proteins in the same fold class as our test proteins according to the SCOP classification; and 3) the proteins having a TM-score (Zhang and Skolnick, 2007) at least 0.5 with our test proteins. Finally, the training data was prepared before CASP8 started. Therefore, we can use our CRF/CNF models to test the CASP8 free-modeling targets without worrying about bias.

The training set is randomly divided into five sets of same size and then used for 5-fold cross validation. To train a CNF model, we shall determine the number of gate functions at the hidden layer. In addition, since the CNF model contains a very large number of model parameters, to avoid overfitting, we shall also control the model complexity. We achieve this by regularizing the $L_2$-norm of the model parameters using a regularization factor. We trained our CNF model by enumerating the number of gate functions (50, 100, 200, and 300) and different regularization factors: 25, 50 100, and 200 to see which one yields the best F1-value. F1-value is widely-used to measure the prediction capability of a machine learning model. F1-value is an even combination of precision $p$ and recall $r$ and defined as $2pr/(p+r)$. The higher the F1-value is, the better the CNF model. Our CNF model achieves the best F1-value (23.44%) when 200 gate functions are used with regularization factor 50. By contrast, the best F1-value achieved by our previous CRF method is 22.0%. The F1-value improvement achieved by CNF over CRF seems not to be very big, partially because in total 100 labels are used in our models. Later we will show that CNF can do conformation sampling much better than CRF.

## Conformation sampling and resampling

Using the trained CNF model, we can sample the whole conformation of a protein or propose a new conformation from an existing one by resampling the local conformation of a segment. This procedure is very similar to the conformation sampling algorithm in our CRF method described in Chapter 3&4. That is, we can use the forward-backward algorithm to first sample labels (i.e., angle distribution) by probability estimated from our CNF model and then sample real-valued angles from the labels. See Chapter 3 for a detailed description of the algorithm.

## Replica exchange Monte Carlo simulation

The energy function we used for folding simulation consists of three items: DOPE (a pairwise statistical potential) (Fitzgerald et al., 2007; Shen and Sali, 2006), KMBhbond (hydrogen-bonding energy) (Morozov et al., 2004) and ESP (a simplified solvent accessibility potential) (Fernandez et al., 2002). We use the weight factors previously trained for the CRF model for these three energy items. Therefore, the energy function is not biased towards our CNF method. The weight factor for DOPE is always fixed to 1, so only two weight factors shall be determined. See Chapter 4 for a detailed description of weight determination.

Previously we employ a simulated annealing (SA) algorithm to minimize energy function, based upon the algorithm proposed by Aarts and Korst (Aarts and Korst, 1991). In this work, we employ a Replica Exchange Monte Carlo (REMC) method (Earl and Deem, 2005; Swendsen and Wang, 1986) to minimize energy function. By using REMC, we can minimize energy function to lower values and thus produce better decoys for most of our test proteins. Our REMC method employs 20 replicas and the highest temperature is set to 100. The temperature for replica $i$ ($i$=1, 2, ..., 20) is set to 5$i$. We have also tested other temperature assignment, but have not seen much difference in terms of folding performance. Each replica consists of 24,000 time steps. At each time step a new conformation is proposed and then accepted with probability $\min\left\{1, \exp\left(-\frac{\Delta E}{T_i}\right)\right\}$ where $\Delta E$ is the energy difference between the new and old conformations and $T_i$ is the temperature for this replica. The conformations between two neighboring replicas are exchanged every 30 time steps. Therefore, in total 800 conformation exchange events will happen between two neighboring replicas during the whole folding simulation. It will make our simulation process very inefficient if we yield only the decoy with the lowest energy at the end of the folding simulation. To generate more decoys from a single folding simulation, we output the final decoy of each replica as long as it has an energy value within 15% of the lowest energy we can achieve. Experimental results indicate that on average, each folding simulation can generate ~10 decoys.

## 5.3 Results

In Chapter 4, **we have demonstrated that our CRF method compares favorably with the popular fragment-based Robetta server in the CASP8 blind prediction**, in this chapter we will focus on the comparison between our CNF and CRF methods and show that our new method is indeed superior over our previous method.

We test our new method using two datasets and compare it with our previous method. These two datasets were used to evaluate our previous method before. The first dataset consists of 22 proteins: 1aa2, 1beo, 1ctfA, 1dktA, 1enhA, 1fc2C, 1fca, 1fgp, 1jer, 1nkl, 1pgb, 1sro, 1trlA, 2croA, 2gb1A, 4icbA, T052, T056, T059, T061, T064 and T074. These proteins have very different secondary structure type and their sizes range from 40 to 120 residues. Some proteins (e.g., T052, T056, T059, T061, T064 and T074) in this dataset are very old CASP targets. Therefore, we denote this dataset as "old testset". The second dataset contains 12 CASP8 free-modeling targets: T0397_D1, T0405_D1, T0405_D2, T0416, T0443_D1, T0443_D2, T0465, T0476, T0482, T0496_D1, T0510_D3 and T0513_D2. These proteins are called free-modeling targets because a structurally similar template cannot be identified for them using a template-based method. We denote this dataset as "CASP8 testset". To avoid bias, we removed all the proteins similar to the first dataset from our training set (see section Model parameter training). Since the training set was constructed before CASP8 started, there is no overlap between our training data and the CASP8 testset.

**Performance on the old test set**

As shown in Table 16, we evaluate our CNF and CRF methods in terms of their capability of generating good decoys. We run both methods on each test protein and generate similar number of decoys (5000-10,000). Each decoy is compared to its native structure and RMSD to the native is calculated for this decoy. Then we rank all the decoys of one test protein in an ascending order by RMSD. Finally we calculate the average RMSD of the top 1%, 2%, 5% and 10% decoys, respectively. We do not compare these two methods using the best decoys because they may be generated by chance and usually the more decoys are generated, the better the best decoys will be. In terms of the average RMSD of the top 5% or 10% decoys, our CNF method outperforms the CRF method on all test proteins except 1ctfA, 1dktA, 1fc2C and 1fgp. The CNF method reduces the average RMSD of top 10% decoys by at least 1Å for many proteins such as 1aa2, 1beo, 1fca, 1pgb, 1sro, 2gb1A, 4icbA, T052, T056, T059, T061 and T064. Furthermore, our CNF method dramatically reduces the average RMSD of top 10% decoys for some proteins. For example, our CNF method reduces the average RMSD of top 10% decoys for 4icbA from 8.0Å to 5.2Å, for T056 from 11.1Å to 7.2Å and for T061 from 7.6Å to 5.6Å. Even for some test proteins (e.g., 1enhA, 1pgb and 2gb1A) on which the CRF method has already performed well, our CNF method still improves a lot.

## Performance on the CASP8 test set.

To further compare our CRF and CNF methods, we also evaluate them on the 12 CASP8 free-modeling (FM) targets. During the CASP8 competition, structurally similar templates cannot be identified for these targets. Similarly, we evaluate both methods in terms of the average RMSD of the top 1%, 2%, 5% and 10% decoys, respectively. Compared to CRF, our CNF method does not significantly worsen the decoy quality of any of the 12 CASP8 targets. Instead, our CNF method outperforms the CRF method on 10 of the 12 targets and yields slightly worse performance on another two targets: T0397_D1 and T0482. In particular, our CNF method reduces the average RMSD of the top 10% decoys by at least 1Å for the following seven targets: T0405_D1, T0405_D2, T0416_D2, T0443_D2, T0476, T0496_D1, and T0510_D3.

Our CNF method reduces the average RMSD of top 10% decoys for T0510_D3 from 9.1 Å to 6.3 Å and for T0496_D1 from 10.1 Å to 8.1 Å. Even for T0416_D2, a target on which our CRF method performed well, our CNF method improves the average RMSD of the top 10% decoys by 1Å. We have also examined the average TM-score/GDT-TS of the top 10% decoys, on average our CNF method is better than the CRF method by ~10% (data not shown due to space limitation).

Table 16 Performance of the CNF and CRF methods on the old test set.

Note: Column "s/t" lists the size and secondary structure content of the test proteins. Column "M" indicates methods. "N" and "R" represent the CNF and CRF methods, respectively. Column "x%" lists the average RMSD (Å) of the decoys among the top x% of the generated decoys. Column "best" lists the RMSD of the best decoys.

|  | s/t | M | best | 1% | 2% | 5% | 10% |
|---|---|---|---|---|---|---|---|
| 1aa2 | 108 | N | 6.0 | 7.0 | 7.6 | 8.5 | 9.2 |
|  | 5α | R | 7.1 | 9.0 | 9.4 | 10.0 | 10.4 |
| 1beo | 98 | N | 5.5 | 6.1 | 6.5 | 7.4 | 8.3 |
|  | 5α | R | 5.6 | 7.2 | 7.8 | 8.7 | 9.3 |
| 1ctfA | 68 | N | 3.6 | 4.5 | 4.8 | 5.4 | 6.1 |
|  | 3α3β | R | 3.3 | 3.9 | 4.1 | 4.6 | 5.2 |
| 1dktA | 72 | N | 4.5 | 5.1 | 5.5 | 6.2 | 6.9 |
|  | 4β | R | 4.5 | 5.0 | 5.3 | 5.9 | 6.6 |
| 1enhA | 54 | N | 1.5 | 2.0 | 2.1 | 2.3 | 2.4 |
|  | 3α | R | 2.1 | 2.6 | 2.7 | 2.9 | 3.0 |
| 1fc2C | 43 | N | 2.0 | 2.3 | 2.4 | 2.5 | 2.6 |
|  | 2α | R | 2.1 | 2.3 | 2.3 | 2.4 | 2.4 |
| 1fca | 55 | N | 3.2 | 3.9 | 4.2 | 4.6 | 5.0 |
|  | 4β | R | 5.0 | 5.6 | 5.8 | 6.2 | 6.4 |
| 1fgp | 67 | N | 6.4 | 7.5 | 8.0 | 8.6 | 9.1 |
|  | 6β | R | 6.6 | 7.3 | 7.6 | 8.1 | 8.6 |
| 1jer | 110 | N | 9.6 | 10.8 | 11.1 | 11.6 | 12.1 |
|  | 2α6β | R | 10.0 | 11.5 | 11.9 | 12.4 | 12.8 |
| 1nkl | 78 | N | 1.8 | 2.5 | 2.6 | 2.8 | 3.0 |
|  | 5α | R | 2.3 | 2.8 | 2.9 | 3.2 | 3.4 |
| 1pgb | 56 | N | 1.4 | 1.9 | 2.0 | 2.3 | 2.6 |
|  | 1α4β | R | 2.2 | 3.0 | 3.2 | 3.5 | 3.7 |
| 1sro | 76 | N | 4.2 | 5.2 | 5.9 | 6.7 | 7.4 |
|  | 6β | R | 5.1 | 6.4 | 6.9 | 7.7 | 8.4 |
| 1trlA | 62 | N | 3.2 | 3.6 | 3.7 | 3.9 | 4.1 |
|  | 6α | R | 3.9 | 4.2 | 4.4 | 4.5 | 4.7 |
| 2croA | 65 | N | 1.8 | 2.2 | 2.3 | 2.4 | 2.5 |
|  | 5α | R | 2.2 | 2.5 | 2.6 | 2.7 | 2.8 |
| 2gb1A | 56 | N | 1.7 | 1.9 | 2.0 | 2.3 | 2.6 |
|  | 1α4β | R | 1.9 | 3.1 | 3.3 | 3.6 | 3.8 |
| 4icbA | 76 | N | 4.1 | 4.8 | 4.9 | 5.1 | 5.2 |

|  | 4α | R | 5.3 | 6.1 | 6.5 | 7.3 | 8.0 |
| --- | --- | --- | --- | --- | --- | --- | --- |
| T052 | 98 | N | 7.6 | 8.1 | 8.5 | 9.1 | 9.6 |
|  | 8β | R | 8.6 | 9.6 | 10.0 | 10.7 | 11.3 |
| T056 | 114 | N | 4.1 | 4.9 | 5.3 | 6.1 | 7.2 |
|  | 6α | R | 7.9 | 9.4 | 9.7 | 10.3 | 11.1 |
| T059 | 71 | N | 5.7 | 6.9 | 7.3 | 7.7 | 8.1 |
|  | 7β | R | 6.9 | 8.4 | 8.7 | 9.2 | 9.6 |
| T061 | 76 | N | 2.8 | 3.4 | 3.7 | 4.6 | 5.6 |
|  | 4α | R | 5.9 | 6.6 | 6.8 | 7.2 | 7.6 |
| T064 | 103 | N | 6.5 | 7.0 | 7.2 | 7.5 | 7.9 |
|  | 8α | R | 5.9 | 7.1 | 7.5 | 8.2 | 8.9 |
| T074 | 98 | N | 3.7 | 5.0 | 5.4 | 5.9 | 6.3 |
|  | 4α | R | 5.0 | 6.0 | 6.4 | 6.7 | 6.9 |

Table 17 Performance of our CNF and CRF methods on the CASP8 test set.
Column "s/t" lists the size and secondary structure content of the test proteins. Column "M" indicates methods. "N" and "R" represent the CNF and CRF methods, respectively. Column "x%" lists the average RMSD (Å) of the decoys among the top x% of the generated decoys. Column "best" lists the RMSD of the best decoys.

|  | s/t | M | best | 1% | 2% | 5% | 10% |
|---|---|---|---|---|---|---|---|
| T0397_D1 | 70 | N | 6.4 | 8.2 | 8.5 | 9.0 | 9.4 |
|  | 7β | R | 7.0 | 8.0 | 8.3 | 8.9 | 9.4 |
| T0405_D1 | 80 | N | 5.0 | 5.4 | 5.5 | 5.7 | 5.9 |
|  | 4α | R | 5.7 | 6.6 | 6.8 | 7.1 | 7.4 |
| T0405_D2 | 112 | N | 7.1 | 9.0 | 9.5 | 10.1 | 10.5 |
|  | 3α6β | R | 8.5 | 10.1 | 10.5 | 11.0 | 11.5 |
| T0416_D2 | 57 | N | 1.4 | 1.9 | 2.1 | 2.3 | 2.6 |
|  | 4α | R | 1.6 | 2.6 | 2.8 | 3.3 | 3.6 |
| T0443_D1 | 86 | N | 4.8 | 6.0 | 6.4 | 7.2 | 7.9 |
|  | 6α | R | 5.6 | 7.1 | 7.7 | 8.3 | 8.7 |
| T0443_D2 | 114 | N | 9.3 | 10.6 | 10.9 | 11.5 | 11.9 |
|  | 2α8β | R | 10.4 | 11.9 | 12.3 | 12.9 | 13.4 |
| T0465 | 157 | N | 11.0 | 11.8 | 12.2 | 12.9 | 13.5 |
|  | 5α8β | R | 10.2 | 12.2 | 12.7 | 13.4 | 13.9 |
| T0476 | 108 | N | 5.3 | 6.3 | 6.8 | 7.4 | 8.0 |
|  | 4α6β | R | 5.9 | 7.8 | 8.2 | 8.7 | 9.3 |
| T0482 | 120 | N | 10.7 | 11.9 | 12.2 | 12.8 | 13.2 |
|  | 3α5β | R | 8.8 | 10.9 | 11.5 | 12.3 | 13.0 |
| T0496_D1 | 110 | N | 5.7 | 6.2 | 6.6 | 7.3 | 8.1 |
|  | 3α6β | R | 6.3 | 8.2 | 8.7 | 9.5 | 10.1 |
| T0510_D3 | 44 | N | 3.0 | 4.0 | 4.5 | 5.3 | 6.3 |
|  | 1α3β | R | 4.7 | 7.2 | 7.7 | 8.6 | 9.1 |
| T0513_D2 | 77 | N | 7.5 | 8.4 | 8.7 | 9.1 | 9.5 |
|  | 2α4β | R | 8.0 | 9.3 | 9.6 | 10.0 | 10.4 |

We have also examined the relationship between RMSD and energy. Due to space limitation, here we only visualize the RMSD-energy relationship for several typical targets: T0397_D1, T0416_D2, T0476, T0482, T0496_D1 and T0510_D3, as shown in Figure 11. Note that in the figure, we normalize the energy of a decoy by the mean and standard deviation calculated from the energies of all the decoys of one target. By energy normalization, we can clearly see the energy difference between the decoys generated by the CNF/CRF methods. Figure 11 clearly demonstrates that our CNF method can generate decoys with much lower energy than the CRF method. However, decoys with lower energy might not have better quality if the correlation between RMSD and energy is very weak. For example, our CNF method can generate decoys for T0397_D1 and T0482 with much lower energy, but cannot improve decoy quality for them. To improve the decoy quality for T0397_D1 and T0482, we have to improve the energy function. By contrast, the correlation between RMSD and energy is positive for T0416_D2, T0476, T0496_D1 and T0510_D3. Therefore, we can improve decoys quality for these four targets by generating decoys with lower energy.

Our CNF method dramatically improves the decoy quality on T0416_D2 over the CRF method, as shown in Figure 11(b). The underlying reason is that our CNF method can estimate the backbone angle probability more accurately. Around half of the decoys generated by the CRF method for T0416_D2 are the mirror images of the other half. These mirror images are introduced by the non-native-like backbone angles around residue #31, as shown in Figure 12. We calculated the marginal probability of the 100 angle states at these residues and found out the native-like angle states have much higher marginal probability in the CNF model than in the CRF model. Thus, our CNF method can sample native-like angles at these residues more frequently than the CRF method and avoid generating a large number of mirror images. In addition to the CNF sampling method, our energy function also helps improve the occurring frequency of native-like angles at these residues.

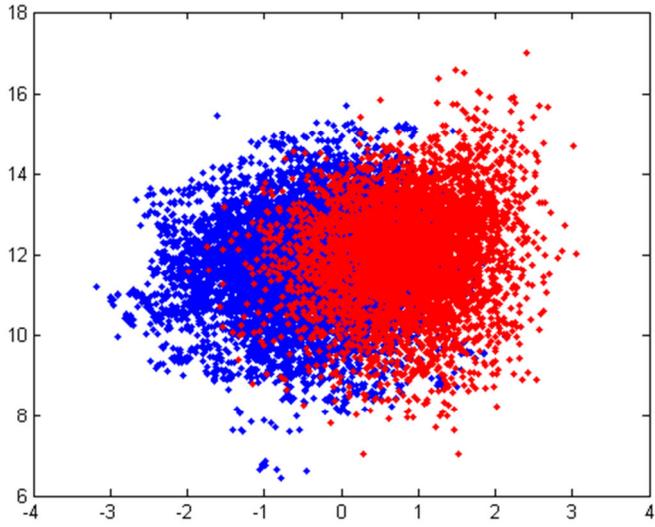
(a) T0397_D1

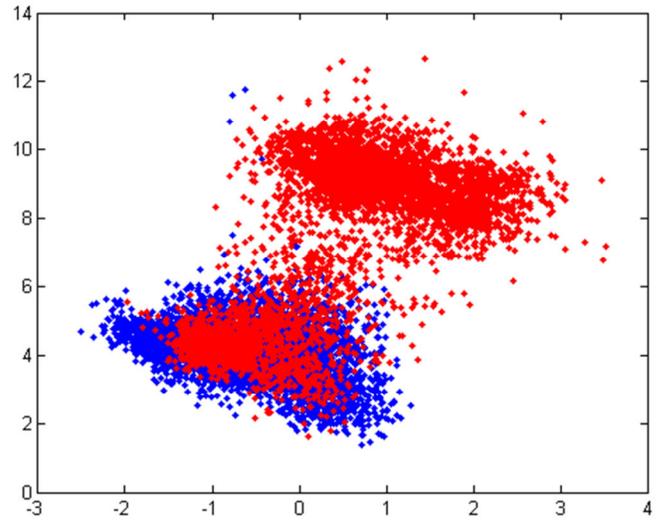
(b) T0416_D2

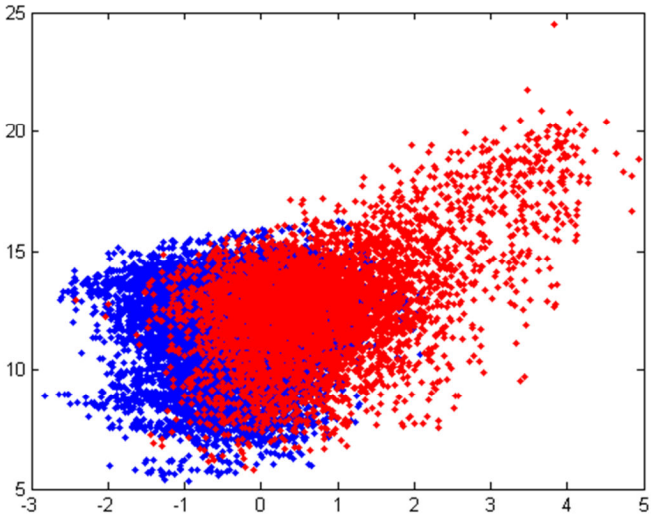
(c) T0476

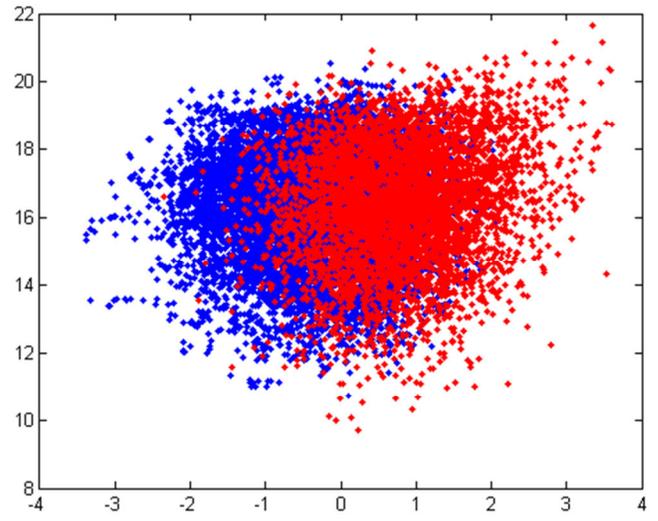
(d) T0482

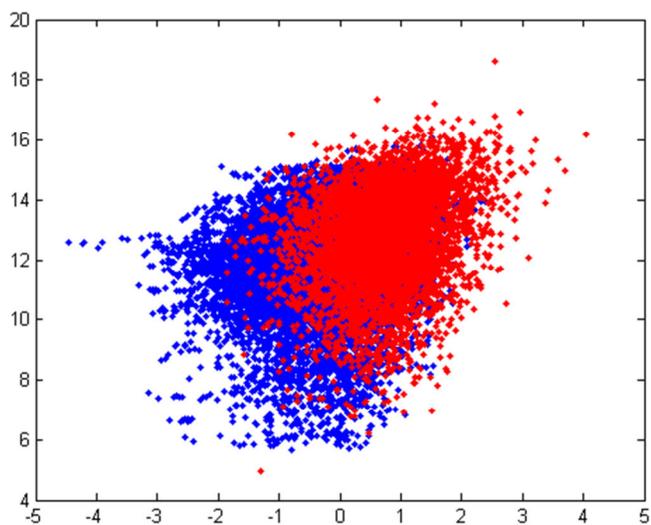 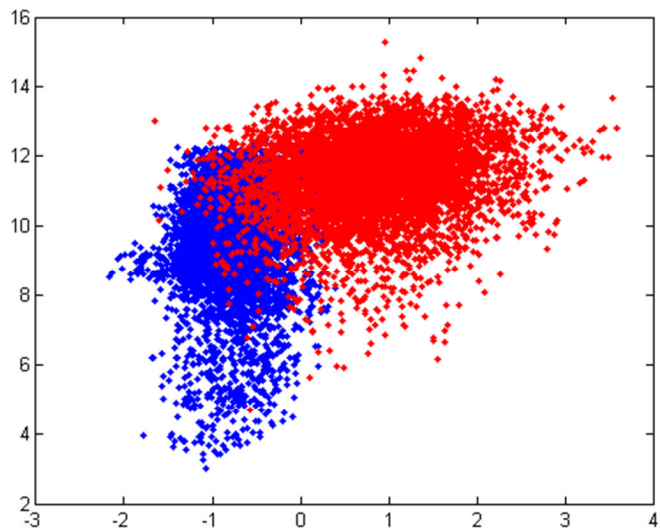

(e) T0496_D1  (f) T0510_D3

Figure 11 The relationship between RMSD (y-axis) and energy (x-axis) for T0397_D1, T0416_D2, T0476, T0482, T0496_D1 and T0510_D3.

The red and blue colors represent the CRF and CNF methods, respectively. See text for the energy normalization method.

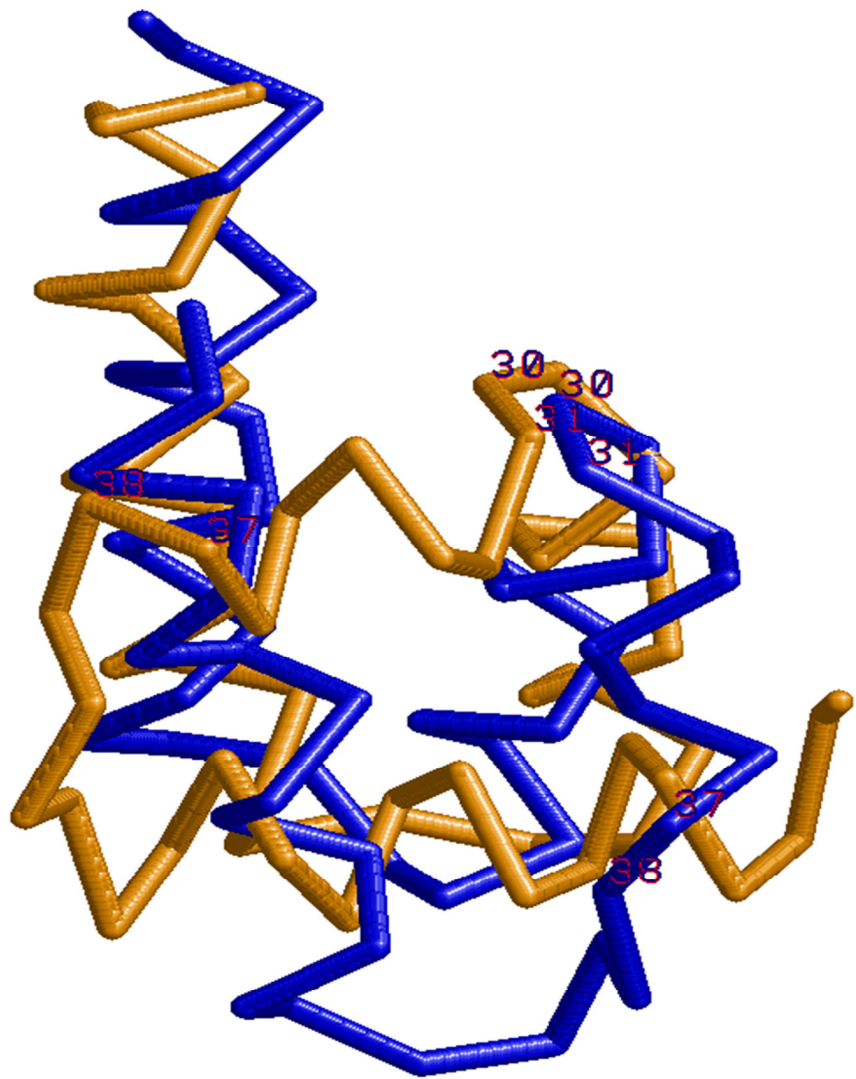

Figure 12 Two typical mirror images generated by the CRF method for T0416_D2. The decoys in blue and gold represent the lower and upper regions in Figure 11(b), respectively.

## Comparison with CASP8 models.

In order to compare our method with the CASP8 results, we cluster the decoys of the 12 CASP8 FM targets using MaxCluster (http://www.sbg.bio.ic.ac.uk/~maxcluster/index.html). We ran MaxCluster so that for a given target, the first cluster contains ~30% of all the decoys and the top 5 clusters in total cover ~70% of the decoys. We examine only the top 5 clusters because CASP8 evaluated at most 5 models for a FM target. As shown in Table 18, we list the GDT-TS of a cluster centroid, its rank among the CASP8 models and its percentile ranking among all the decoys we generated. As shown in this table, our method did pretty well on T0405_D1, T0416_D2, T0443_D1, T0476, T0496_D1, T0510_D3, and T0513_D2, reasonably well on T0397_D1, T0405_D2 and T0465 and badly on T0443_D2 and T0482. Roughly speaking our method can do well on mainly-alpha or small beta proteins, but not well on large beta proteins. This is expected since our CNF method can model well local sequence-structure relationship, but cannot model long-range hydrogen bonding.

Note that we generated decoys using domain definition we decided during the CASP8 season. Therefore, our domain definition may not be consistent with the CASP8 official definition. In this case we calculate the GDT-TS of a model using the native structure common to our domain definition and CASP8 definition. The GDT-TS of a model is calculated using the TM-score program and may be slightly different from the CASP8 official GDT-TS.

### Specific examples

In CASP8, we did prediction using the CRF method for T0476, T0496_D1 and T0510_D3, but not for T0416_D2 because our CRF method was not ready at the beginning of CASP8. The server model generated by our CRF method for T0510_D3 is among the best CASP8 server models (available at http://predictioncenter.org/casp8/results.cgi). Our CNF method further improves predictions for these four targets over the CRF method.

**T0416_D2.** The first and best cluster centroids have GDT-TS 69.3 and 76.8, respectively. As shown in Figure 13(a), the best cluster centroid is better than all the CASP8 server models. In fact the best cluster centroid is also better than all the CASP8 human models (data not shown). The best cluster centroid also has a small RMSD 2.7Å.

**T0476.** The first and best cluster centroids have GDT-TS 34.2 and 35.6, respectively. Our first and best cluster centroids for T0476 are ranked No.4 out of 66 and No. 15 out of 287 CASP8 server models, respectively. The best human model for T0476 has GDT-TS 48.3 and RMSD 7.8Å. Our best cluster centroid also has RMSD 7.8 Å.

**T0496_D1.** According to Grishin group, T0496_D1 is one of the only two CASP8 targets representing new folds (Shi et al., 2009). Our first and best cluster centroids have GDT-TS 30.5 and 49.1, respectively. As shown in Figure 13(c), the best cluster centroid is significantly better than all the CASP8 server models. In fact the best cluster centroid is also significantly better than all the CASP8 human models. The best CASP8 model has GDT-TS only 33.96. The smallest RMSD among the CASP8 models with 100% coverage is 11.34Å. Our best cluster centroid has a pretty good RMSD 6.2Å considering that this target has more than 100 residues. In summary, our CNF method can predict an almost correct fold for this target.

**T0510_D3.** The first and best cluster centroids have GDT-TS 47.7 and 51.7, respectively. The best cluster centroid has RMSD 6.9Å. As shown in Figure 13(d), our first cluster centroid is better than all the #1 models submitted by the CASP8 servers. If all the 321 CASP8 models are considered, our first cluster centroid is worse than only 3 of them (There are very few human predictions for T0510_D3) and our best centroid is ranked No. 2.

## 5.4 Conclusion

This chapter has presented a new fragment-free approach to protein ab initio folding by using a recently-invented probabilistic graphical model Conditional Neural Fields (CNF). Our fragment-free approach can overcome some limitations of the popular fragment assembly method. That is, this new method can sample protein conformations in a continuous space while the fragment-based methods cannot do so. This CNF method is also better than our previous CRF method in that 1) this method can easily model nonlinear relationship between protein sequence and structure; and 2) we can also minimize energy function to lower values. Experimental results indicate that our CNF method clearly outperforms the CRF method on most of the test proteins. Previously, we have compared our CRF method with the popular fragment-based Robetta server in the CASP8 blind prediction and shown that our CRF method is on average better than Robetta on mainly-alpha or small beta proteins (Zhao et al., 2009). This work further confirms our advantage on mainly-alpha or small beta proteins. Since CNF is better than CRF in modeling nonlinear sequence-structure relationship, we are going to incorporate more information (such as amino acid physical-chemical property profile) to our model so that we can improve sampling efficiency further.

Table 18. Clustering result of the 12 CASP8 free-modeling targets.

Column "GDT" lists the GDT-TS of the first and best cluster centroids. Column "CASP8 Rank" lists the rank of the #1 cluster centroid or the best cluster centroid among the first CASP8 server models or all the CASP8 server models, respectively. Column "Internal Rank" lists the percentile ranking (%) of a cluster centroid among all the decoys we generated for the target.

| Target | First Cluster | | | Best Cluster | | |
|---|---|---|---|---|---|---|
| | GDT | CASP8 Rank | Internal Rank(%) | GDT | CASP8 Rank | Internal Rank(%) |
| T0397_D1 | 25.7 | 12/60 | 50.6 | 28.6 | 28/262 | 18.8 |
| T0405_D1 | 39.2 | 6/63 | 41.6 | 48.4 | 14/285 | 6.5 |
| T0405_D2 | 27.0 | 10/62 | 72.3 | 34.6 | 19/280 | 5.1 |
| T0416_D2 | 69.3 | 1/53 | 5.4 | 76.8 | 1/242 | 3.5 |
| T0443_D1 | 46.9 | 3/64 | 38.2 | 49.2 | 6/253 | 19.7 |
| T0443_D2 | 24.8 | 26/59 | 35.3 | 27.9 | 73/252 | 12.1 |
| T0465 | 31.3 | 12/65 | 12.6 | 31.3 | 34/286 | 12.6 |
| T0476 | 34.2 | 4/66 | 17.5 | 35.6 | 15/287 | 10.0 |
| T0482 | 34.2 | 34/65 | 4.3 | 34.2 | 132/279 | 4.3 |
| T0496_D1 | 30.5 | 1/59 | 30.3 | 49.1 | 1/266 | 0.4 |
| T0510_D3 | 47.7 | 1/54 | 15.7 | 51.7 | 2/244 | 3.3 |
| T0513_D2 | 57.7 | 5/50 | 3.8 | 57.7 | 17/225 | 3.8 |

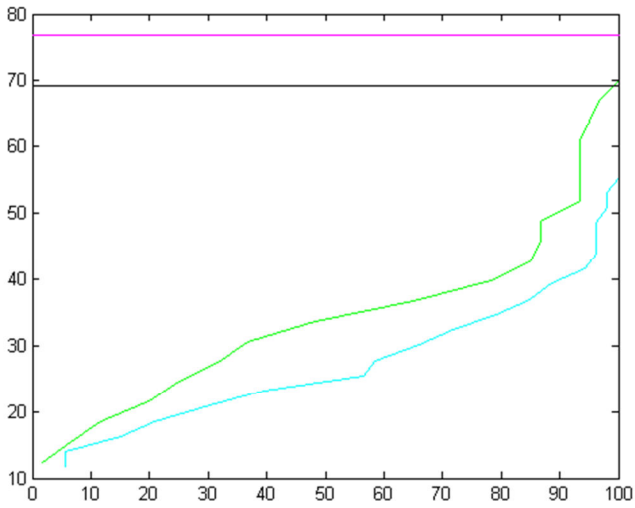

(a) T0416_D2

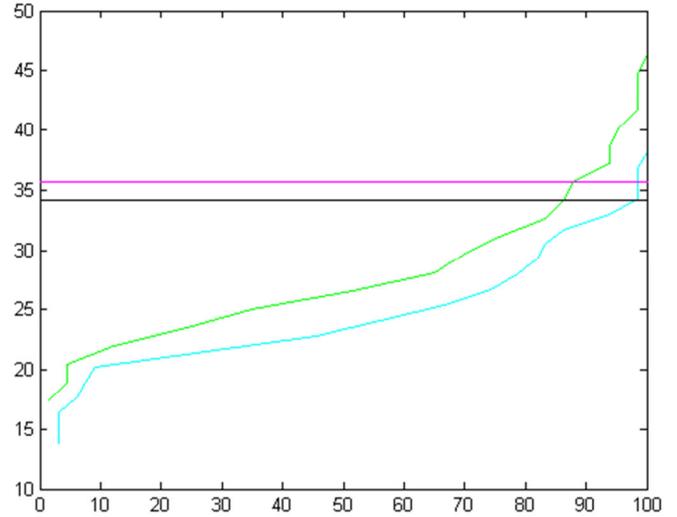

(b) T0476

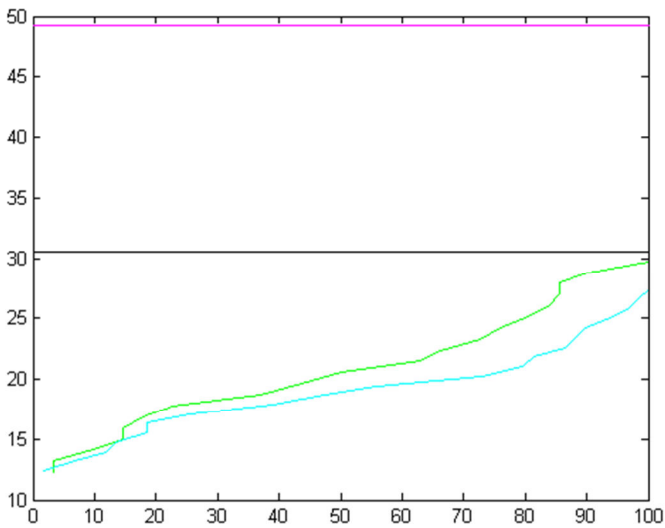

(c) T0496_D1

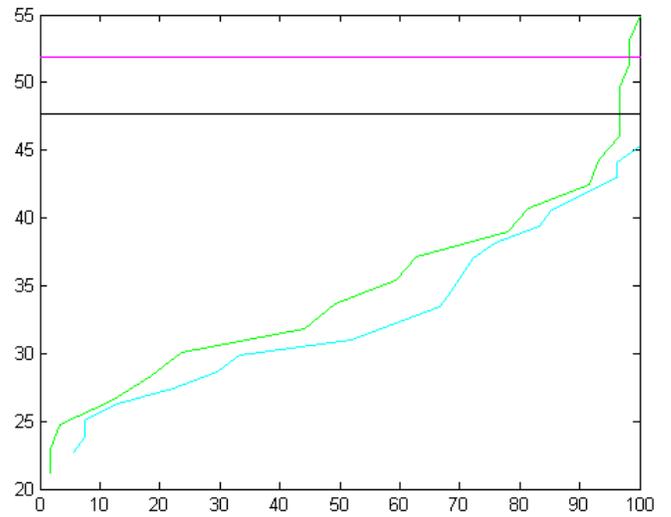

(d) T0510_D3

Figure 13 Ranking of our CNF predictions for T0416_D2, T0476, T0496_D1 and T0510_D3.

The x-axis is percentile ranking and y-axis GDT-TS. Our first and best cluster centroids are plotted in black and magenta lines, respectively. The #1 models submitted by the CASP8 server are ordered by their GDT-TS and their percentile ranking is displayed as a cyan curve, so are the best models from each server but as a green curve.

# Part 2.
# Statistical Potentials Using Machine Learning Methods

# Chapter 6. EPAD: A position-specific pairwise distance potential

## 6.1 Introduction

A lot of statistical potentials are derived from the inverse of the Boltzmann law. In the traditional position-independent distance-dependent statistical potentials (e.g., DOPE and DFIRE), the interaction potential of two atom types a and b can be estimated as follows.

$$U(d\,|\text{a},\text{b}) = -kT\,ln\,\frac{P(d\,|\,a,\,b\,)}{q(d)} \tag{6.1}$$

Where k is the Boltzmann constant, T is the temperature, and $d$ represents the inter-atom distance shell $[d, d + \Delta d]$. Meanwhile, $P(d\,|\text{a},\text{b})$ is the observed probability of two atoms interacting within the distance shell and $q(d)$ is the reference state (i.e., the expected probability of two non-interacting atoms within the distance shell). The reference state is used to rule out the average and generic correlation of two atoms not due to atomic interactions. Most statistical potentials parameterize the observed atomic interacting probability by (residue-specific) atom types and use a simple counting method to estimate it. For example, $P(d\,|\text{a},\text{b})$ in Eq. (6.1) is often calculated by $\frac{count(d,a,b)}{\sum_d count(d,a,b)}$ where $count(d,a,b)$ is the number of observed occurrences of two atoms a and b within a distance shell $[d, d + \Delta d]$. The distance-dependent statistical potentials developed so far mainly differ from one another in estimating the reference state (Shen and Sali, 2006; Wu et al., 2007b; Zhang and Zhang, 2010; Zhou and Zhou, 2002). Some (e.g., DFIRE and DOPE) use analytical methods to estimate the reference state while others use statistical methods (e.g., KBP (Lu and Skolnick, 2001) and RAPDF (Samudrala and Moult, 1998)). Although using different reference states, these potentials do not have very different energy curves (see Figure 2 in the RW paper (Zhang and Zhang, 2010) and Figure 4 in the DOPE paper (Shen and Sali, 2006)). These traditional position-independent potentials share a couple of common properties: 1) once the atom distance and types are given, the atomic interaction potential is fixed across all proteins and residues; and 2) the atomic interaction potentials approach to 0 when the distance is larger than 8Å.

This chapter presents a novel protein-specific and position-specific statistical potential EPAD. We parameterize the observed probability in EPAD by the evolutionary information and radius of gyration of the protein under consideration, in addition to atom types. EPAD distinguishes itself from others in that it may have different energy profiles for two atoms of given types, depending on the protein under consideration and the sequence profile context of the atoms (i.e., evolutionary information). Evolutionary information has been extensively used in protein secondary structure prediction(Jones, 1999; Wang et al., 2011), fold recognition (Maiorov and Crippen, 1992; Panchenko et al., 2000; Peng and Xu, 2009, 2010; Sippl and Weitckus, 1992; Skolnick et al., 2000), protein alignment (Notredame et al., 2000; Pei et al., 2008; Wu and Zhang, 2008b; Xu, 2005; Zhang and Skolnick, 2005b), model quality assessment (Jones and Thornton, 1996; Panchenko et al., 2000; Peng and Xu, 2010; Reva et al., 1997; Sippl, 1993) and even protein conformation sampling (Bystroff et al., 2000; Simons et al., 1997; Zhao et al., 2008; Zhao et al., 2010). However, evolutionary information is rarely used to design a statistical potential suitable for *ab initio* protein folding. Panjkovich et al have developed a structure-specific statistical potential using evolutionary information for the assessment of comparative models (Panjkovich et al., 2008). Nevertheless, this potential is not position-specific and subject to a couple of restrictions: 1) it requires the knowledge of at least one native structure in a protein family, so it cannot be applied to *ab initio* folding a protein with novel fold or to the assessment of models built from distantly-related templates; and 2) it requires at least 50 sequence homologs for sufficient statistics. By contrast, our statistical potential is not subject to such restrictions and thus, is more widely applicable. We term our statistical potential as **e**volutionary **pa**irwise **d**istance-dependent potential (EPAD).

Experimental results show that our position-specific statistical potential outperforms currently many popular ones in several decoy discrimination tests. These results imply that in addition to reference state, the observed atomic interacting probability is also critical to statistical potentials and can be estimated much more accurately using context-specific evolutionary information.

## 6.2 Methods and Experimental Procedures

Let $a_i$ and $a_j$ denote two atoms of two residues at positions $i$ and j, respectively.

Let $S$ denote the sequence profile of the protein under consideration. It is generated by running PSI-BLAST on the NR database with at most 8 iterations and E-value 0.001. $S$ is a position-specific scoring matrix with dimension $20 \times N$ where $N$ is the sequence length. Each column in $S$ is a vector of 20 elements, containing the mutation potential to the 20 amino acids at the corresponding sequence position. The sequence profile context of the residue at sequence position $i$ is a 20×15 submatrix of $S$, consisting of 15 columns $i-7, i-6,\ldots, i, i+1,\ldots, i+7$. In case that one column does not exist in $S$ (when $i \leq 7$ or $i+7 > N$), the zero vector is used.

Let $S_i$ and $S_j$ denote position-specific sequence profile contexts at positions $i$ and $j$, respectively. Our distance-dependent statistical potential is defined as follows.

$$U(d \mid a_i, a_j, S_i, S_j, r_g) = -kT \ln \frac{P(d \mid a_i, a_j, S_i, S_j, r_g)}{q(d \mid r_g)} \tag{6.2}$$

where k is the Boltzmann constant and T is the temperature, $q(d|r_g)$ is the reference state, and $P(d|a_i, a_j, S_i, S_j, r_g)$ is the observed probability of two atoms $a_i$ and $a_j$ interacting within a distance shell $[d, d + \Delta d]$ conditioned on atom types, residue sequence profile contexts and $r_g$ (the estimated radius of gyration of the protein under consideration). We use $r_g = 2.2\ N^{0.38}$ to estimate the radius of gyration where N is the protein sequence length. Comparing to Eq. (6.1), our statistical potential differs from the traditional position-independent potentials (e.g., DOPE and DFIRE) in a couple of aspects. First, the interaction potential of two atoms is protein-specific since it depends on the evolutionary information and radius of gyration of the protein under consideration. Second, our potential is position-specific since it is parameterized by sequence profile contexts in addition to atom types. We use the same reference state as DOPE (Shen and Sali, 2006), which is a finite sphere of uniform density with appropriate radius. That is, the reference state depends on only the size of a sample protein structure (see the section of **Estimating the reference state** for more details).

We cannot use the simple counting method to calculate $P(d|a_i, a_j, S_i, S_j, r_g)$ since there are insufficient number of solved protein structures in PDB for reliable simple counting of sequence profile contexts $S_i$ and $S_j$. Instead, we apply a probabilistic neural network (PNN) (Specht, 1990) to estimating $P(d|a_i, a_j, S_i, S_j, r_g)$ when both $a_i$ and $a_j$ are $C_\alpha$ atoms. PNN will effectively learn the sophisticated relationship between inter-atom distance and sequence profiles and yield accurate distance probability distribution. We then estimate $P(d|a_i, a_j, S_i, S_j, r_g)$ for non-$C_\alpha$ atoms conditioned upon $C_\alpha$ distance distribution.

**Distribution of $C_\alpha$ distances in a representative set of protein structures.**

Simple statistics on the PDB25 data set indicates that nearly 90% of residue pairs have $C_\alpha$ distance larger than 15Å and only 1% of them have $C_\alpha$ distance less than 4Å (see Figure 14).

**Estimating pairwise $C_\alpha$ distance distribution using probabilistic neural network (PNN).**

We discretize all the $C_\alpha - C_\alpha$ distance into 13 bins (3~4Å, 4-5Å, 5-6Å, …, 14-15Å, and >15Å). Given a protein and its $k^{th}$ residue pair of two residues $i$ and $j$, let $d_k$ denote the bin into which the distance of the $k^{th}$ residue pair falls into, and $x_k$ the position-specific feature vector, which contains sequence profile contexts $S_i$ and $S_j$ centered at the two residues $i$ and $j$ under consideration and the estimated radius of gyration of the protein under consideration.

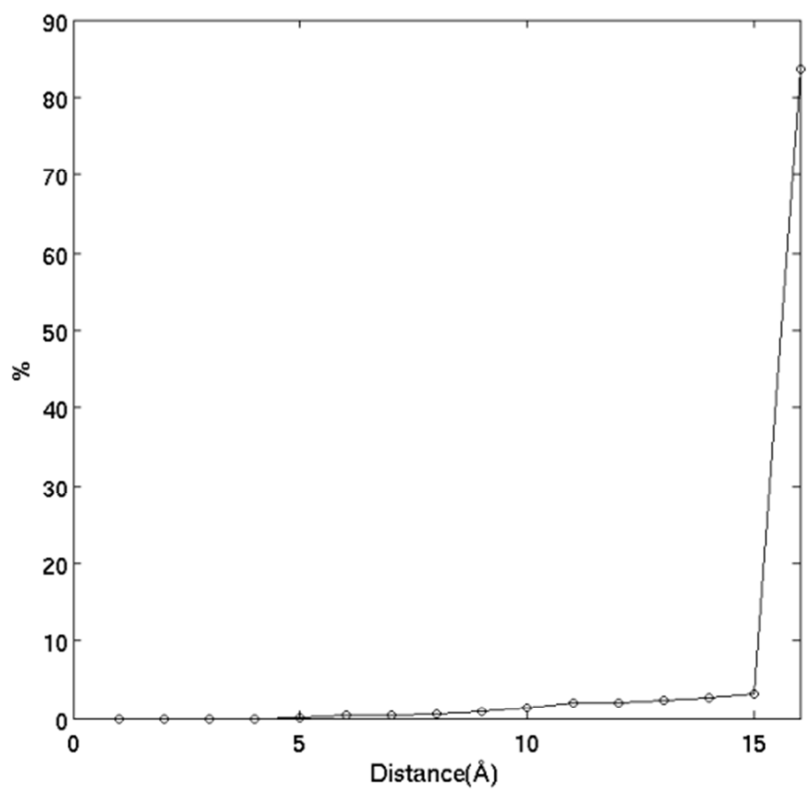

Figure 14 Distribution of pairwise $C_\alpha$ distances in a representative set of protein structures.

We always use $r_g = 2.2 N^{0.38}$ to estimate the radius of gyration for one protein where $N$ is the protein sequence length. That is, $r_g$ is independent of any 3D models including the native structure. We do not use $r_g$ specific to a decoy because our training set does not contain any decoys. We do not use $r_g$ calculated from the native structures either because in the realistic settings they are unavailable.

Let $p_\theta(d_k|x_k)$ be the probability of the distance label $d_k$ conditioned on the feature vector $x_k$. Meanwhile, $\theta$ is the model parameter vector. We estimate $p_\theta(d_k|x_k)$ as follows.

$$p_\theta(d_k|x_k) = \frac{\exp(\phi_\theta(x_k, d_k))}{Z_\theta(x_k)} \tag{6.3}$$

where $Z_\theta(x_k) = \sum_d \exp(\phi_\theta(x_k, d))$ is the partition function and $\phi_\theta(x, d)$ is a two-layer neural network. Figure 15 shows an example of the neural network with three and five neurons in the first and second hidden layers, respectively. Each neuron represents a sigmoid function $h(x) = 1/(1 + \exp(x))$. Therefore, we have

$$\phi_\theta(x_k, d_k) = \sum_{g_1=1}^{G_1} \theta^0_{d_k,g_1} h\left(\sum_{g_2=1}^{G_2} \theta^1_{g_1,g_2} h(<\theta^2_{g_2}, x_k>)\right) \tag{6.4}$$

Where $G_1$ and $G_2$ are the number of gates in the two hidden layers, $<.,.>$ denotes the inner product of two vectors, $\theta^2_{g_2}$ is the weight vector of the $g_2^{th}$ neuron (also known as gate) in the 2nd layer; $\theta^1_{g_1,g_2}$ is the weight connecting the $g_2^{th}$ neuron in the 2nd layer to the $g_1^{th}$ neuron in the 1st layer; and $\theta^0_{d_i,g_1}$ is the weight connecting the $g_1^{th}$ neuron in the 1st layer to the label $d_k$.

In the implementation, our neural network consists of two hidden layers. The first hidden layer (i.e., the layer connecting to the input layer) contains 100 neurons and the second hidden layer (i.e., the layer connecting to the output layer) has 40 neurons. This neural network is similar to what is used by the Zhou group for inter-residue contact prediction (Xue et al., 2009), which uses 100 and 30 neurons in the two hidden layers, respectively. The Zhou group has shown that using two hidden layers can obtain slightly better performance than using a single hidden layer. The input layer of our network has about 600 neurons, so in total our neural network has between 60,000 and 70,000 parameters to be trained.

**Model parameter training.**

We use the maximum likelihood method to train the model parameter $\theta$ and to determine the window size and the number of neurons in each hidden layer, by maximizing the occurring probability of the native $C_\alpha - C_\alpha$ distance in a set of training proteins. Given a training protein $t$ with solved experimental structure, let $D^t$ denote the set of pairwise residue distances and $X^t$ the set of all feature vectors. By assuming any two residue pairs to be independent of one another, we have

$$p_\theta(D^t|X^t) = \prod_{k=1}^{m_t} p_\theta(d_k^t|x_k^t) \tag{6.5}$$

where $m_t$ is the number of residue pairs in the protein $t$.

Given T training proteins, we need to maximize $\prod_{t=1}^{T} p_\theta(D^t|X^t)$, which is equivalent to the following optimization problem.

$$\begin{aligned}
\min_\theta &\sum_{t=1}^{T} -\log p_\theta(D^t|X^t) + \lambda\|\theta\|_2^2 \\
&= \min_\theta \sum_{t=1}^{T}\sum_{i=1}^{m^t} \left(-\log Z_\theta(x_k^t) + \phi_\theta(x_k^t, d_k^t)\right) + \lambda\|\theta\|_2^2
\end{aligned} \tag{6.6}$$

Meanwhile, $\lambda\|\theta\|_2^2$ is a $L_2$-norm regularization item to avoid overfitting and $\lambda$ is a hyper-parameter to be determined. This optimization problem can be solved by LBFGS(Liu and Nocedal, 1989).

It is very challenging to solve this non-convex optimization problem due to the huge amount of training data. We generate an initial solution randomly and then run the training algorithm on a supercomputer for about a couple of weeks. Our training algorithm terminates when the probability of either the training set or the validation set does not improve any more. Note that all the model parameters are learned from the training set, but not the validation set. The validation set, combined with the training set, is only used to determine when our training algorithm shall terminate. Our training algorithm usually terminates after 3000 iterations. We also reran our training algorithm starting from 9 initial solutions and did not observe explicit performance difference among these runs.

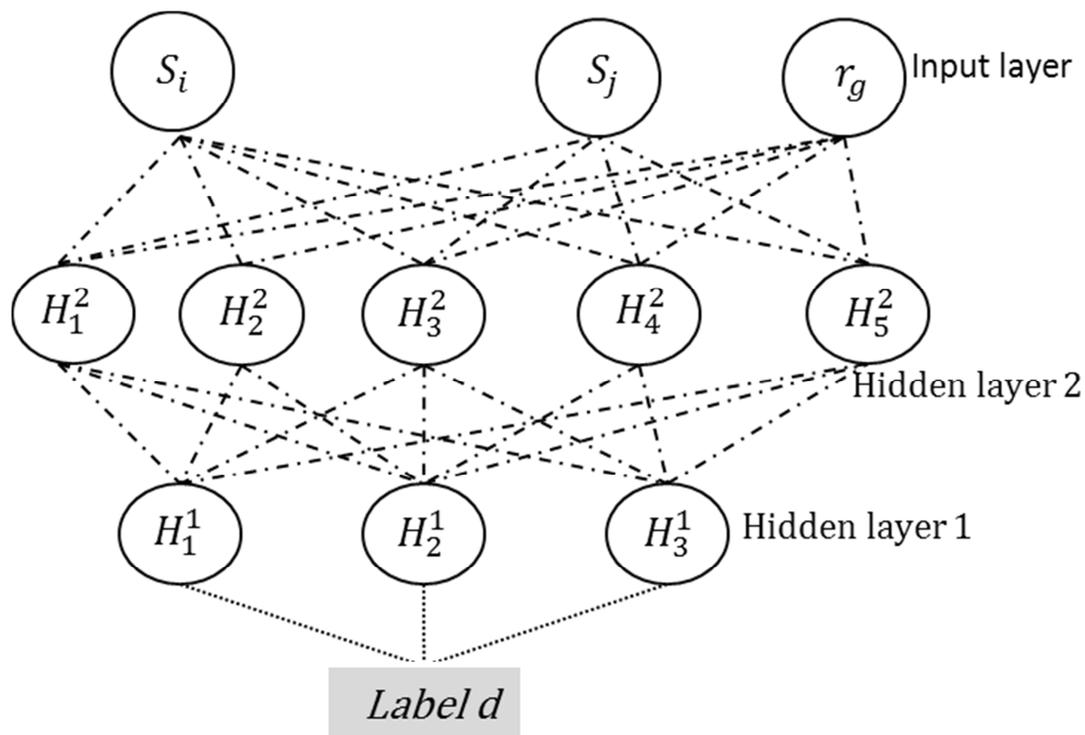

Figure 15 An example probabilistic neural network, in which $S_i$ and $S_j$ are the sequence profile contexts centered at the $i^{th}$ and $j^{th}$ residues, respectively. $H_q^1$ and $H_p^2$ are the neurons in the 1st and 2nd hidden layers.

## Training and validation data.

We use the PDB25 set of the PISCES server (Wang and Dunbrack, 2003) early in 2011 as the training and validation data. Any two proteins in PDB25 share no more than 25% sequence identity. Such a set in total includes more than 6000 proteins. We randomly chose about 5000 proteins from this PDB25 set as the training and validation proteins and also make sure that they have no overlap (i.e., > 25% sequence identity) with the Rosetta set (Qian et al., 2007) and the Decoy 'R' Us set (Samudrala and Levitt, 2000). We randomly choose 3/4 of the 5000 proteins as the training data and the remaining 1/4 as the validation data, which contain ~73 million training and ~19 million validation residue pairs, respectively. It is challenging to train our neural network model because 1) the number of training residue pairs is huge; and 2) the distance distribution is extremely unbalanced. As shown in Figure 14, 90% of residue pairs have $C_\alpha$ distance larger than 15Å and only 1% of them have $C_\alpha$ distance less than 4Å. It takes a couple of weeks to train a single neural network model using 1296 CPUs on a Cray supercomputer.

## Estimating inter-atom distance distribution for non-$C_\alpha$ main chain atoms.

We discretize the inter-atom distance of non-$C_\alpha$ atoms into 26 equal-width bins, each with 0.5Å. Due to limited computation resources, instead of training neural network models for each pair of atom types, which will take months or even a year to finish, we use a different approach to estimate the pairwise distance probability distribution for non-$C_\alpha$ main chain atoms. In particular, we calculate the inter-atom distance probability distribution for non-$C_\alpha$ main chain atoms conditioned upon $C_\alpha - C_\alpha$ distance probability distribution. Let $P_{\alpha\alpha}(d_{\alpha\alpha}|S_i, S_j, r_g)$ denote the $C_\alpha - C_\alpha$ distance probability distribution for residues $i$ and $j$, which can be estimated by our probabilistic neural network. Let $a$ and $b$ denote the amino acid types of the residues at $i$ and $j$, respectively. For the purpose of simplicity, we use $N$ and $O$ atoms as an example to show how to calculate the observed atomic interacting probability. Let $P(d|N, O, S_i, S_j, r_g)$ denote the distance probability distribution for the nitrogen atom in residue $i$ and the oxygen atom in residue $j$. We calculate $P(d|N, O, S_i, S_j, r_g)$ as follows.

$$P(d|N,O,S_i,S_j,r_g) = \sum_{d_{\alpha\alpha}} P_{NO}^{a,b}(d \mid d_{\alpha\alpha}) P_{\alpha\alpha}(d_{\alpha\alpha}|S_i, S_j, r_g) \tag{6.7}$$

where $P_{NO}^{a,b}(d \mid d_{\alpha\alpha})$ is the conditional distance probability distribution for atom $N$ in amino acid $a$ and $O$ in amino acid $b$ when the $C_\alpha$ distance of these two amino acids is $d_{\alpha\alpha}$. Since $P_{NO}^{a,b}(d \mid d_{\alpha\alpha})$ is position-independent, it can be estimated by simple counting.

## Estimating the reference state.

We calculate the reference state of a distance $d$ as the probability of two random points at the distance $d$ within a 3D ball of uniform density. Such a reference state has been discussed in detail before (Deltheli, 1919; Garcia-Pelayo, 2005; Shen and Sali, 2006). Here we briefly introduce it for completeness. Let $r_g$ be the radius of gyration of a protein. The estimated radius of the 3D sphere, denoted as $S^3$, corresponding to such a protein is $a = \sqrt{\frac{5}{3}} r_g$. The probability density $q(d|r_g)$ for the distance $d$ between two randomly chosen points $y_1$ and $y_2$ inside such a 3D sphere can be calculated as follows.

$$q(d|r_g) = \frac{\int_{S^3} d\vec{y_2} \int_{S^3} d\vec{y_1} \delta(|\vec{y_1} - \vec{y_2}| - d)}{\int_{S^3} d\vec{y_2} \int_{S^3} d\vec{y_1}} \tag{6.8}$$

where the delta function is defined by $\delta(y) = \begin{cases} 1, if\ y = 0 \\ 0, otherwise \end{cases}$.

To calculate $q(d|r_g)$, we fix point $y_1$ and move $y_2$ around $y_1$ within the sphere and then consider the spherical area formed by $y_2$, denoted as $O(h, d)$ where $h$ is the distance between the center of the 3D ball and point $y_1$. See (Garcia-Pelayo, 2005) for detailed deduction of $O(h, d)$. We only give the final result as follows.

$$O(h, d) = \begin{cases} 4\pi d^2, d < a - h \\ \pi d(d + a - h)\left(1 + \frac{a - d}{h}\right), a - h \leq d \leq a + h \\ 0, d > a + h \end{cases}$$

Since $\int_{S^3} d\vec{y_2} = \int_{S^3} d\vec{y_1} = \frac{4\pi a^3}{3}$, we have

$$q(d|r_g) = \frac{4\pi}{\left(\frac{4\pi a^3}{3}\right)^2} \int_0^a dh\ h^2 O(h, d) = \frac{3d^2(d - 2a)^2(d + 4a)}{16a^6} \tag{6.9}$$

**Window size and the number of neurons in the hidden layers.**

The window size for a sequence profile context and the number of neurons in the hidden layers are important hyper-parameters of our probabilistic neural network. Because it is time-consuming to train even a single neural network model for the estimation of distance probability distribution, we determine these hyper-parameters by training a neural network for inter-residue contact prediction, which obtains the best performance when the window size is 15 and the numbers of neurons in the first and second hidden layers are 40 and 100, respectively. The window size used by us is consistent with what is used by the Zhang group (Wu and Zhang, 2008a) and the numbers of hidden neurons are not very different from what is used by the Zhou group (Xue et al., 2009).

A small window size or number of neurons may result in a neural network with suboptimal performance while a very large window size or number of neurons may cause overfitting. Because it is time-consuming to train even a single PNN model for the estimation of distance probability distribution, we determine these hyper-parameters by training a neural network for inter-residue contact prediction. This takes much less time since there are fewer labels and we also use less training data. Our neural network obtains the best performance when the window size is 15 and the numbers of neurons in the first and second hidden layers are 40 and 100, respectively.

Following CASP definition, there is a contact between two residues if their $C_\beta - C_\beta$ distance is no more than 8Å. A pair of residues is treated as a positive example if they are in contact; otherwise a negative example. To compare our prediction results with CASP8 and CASP9 (human and server group) submissions, we use the PDB25 set generated by the PISCES server before CASP8 started to train our neural network. This PDB25 set contains ~3000 proteins and any two of them share no more than 25% sequence identity. We randomly chose ¾ of the set as the training data and the remaining ¼ as the validation data and use PSI-BLAST sequence profiles and PSIPRED predicted secondary structures as the input features. To avoid bias, we also used PSIPRED, PSI-BLAST and the NR database released before CASP8 in 2008. Our method achieves the best performance when the window size is 15 and the numbers of neurons in the first and second hidden layers are 40 and 100, respectively.

We test our neural network models on the 13 CASP8 and 28 CASP9 contact prediction targets, all of which are template-free modeling targets. Our method obtains an $L/5$ average accuracy of 57.87% on the CASP8 targets and of 36.63% on the CASP9 targets, respectively. As shown in Figure 16, our method exceeds all the CASP8 top groups and the 10 CASP9 server groups (represented by blue bars). Our method works particularly well for the beta proteins and alpha-beta proteins, obtaining an accuracy of 49.30% on the CASP9 beta proteins, significantly better than the accuracy (25.65%) on the alpha or mainly-alpha proteins.

After determining the number of neurons in the hidden layers, we also examine the $C_\alpha - C_\alpha$ distance prediction accuracy of our PNN with respect to the window size. As shown in Table 19, the neural network model yields almost the best performance when the window size is 15, which is consistent what is used by the Zhang group for contact prediction(Wu and Zhang, 2008a). The results in this table are obtained from the EPAD validation data.

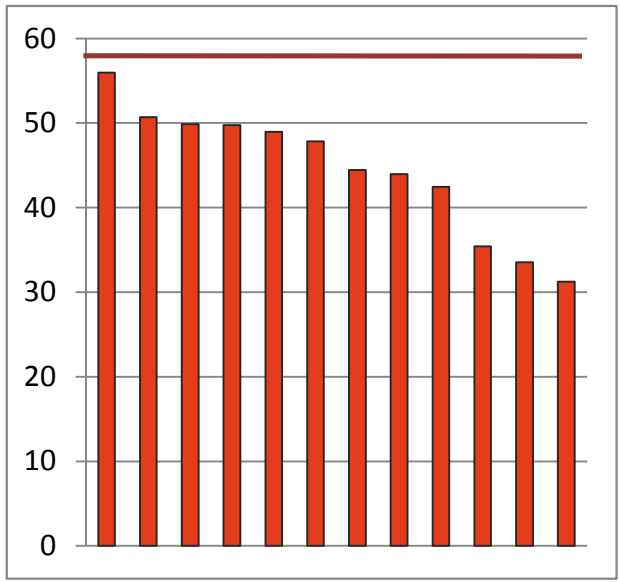 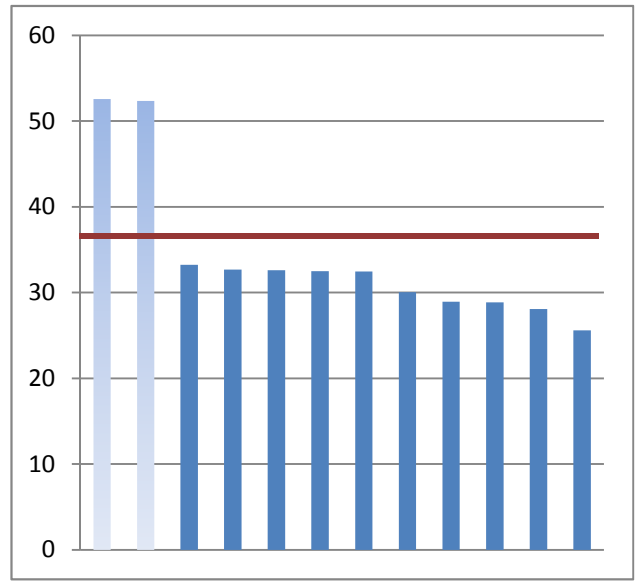

(A)                                                    (B)

Figure 16 Inter-residue contact prediction accuracy of our PNN and the top 12 (human and server) groups in CASP8 and CASP9.

The accuracy of our PNN is displayed by two brown lines. The predictions of the 12 groups are downloaded from the CASP website (see http://www.predictioncenter.org/download_area/). (A) The performance on the CASP8 targets. (B) The performance on the CASP9 targets and the top 2 groups are human groups and the the remainings are the server groups.

Table 19 $C_\alpha - C_\alpha$ distance prediction accuracy with respect to the window size

| window size | 9 | | 15 | | 19 | | 23 | |
|---|---|---|---|---|---|---|---|---|
| distance | precision | recall | precision | recall | precision | recall | precision | recall |
| <8 Å | 26.79 | 5.97 | 38.04 | 20.1 | 36.84 | 19.6 | 36.23 | 20.17 |
| 8-9 Å | 10.23 | 0.01 | 0 | 0 | 0 | 0 | 0 | 0 |
| 9 -10Å | 16.25 | 0.01 | 25.79 | 0.66 | 23.93 | 0.49 | 22.29 | 0.65 |
| 10 -11Å | 15.79 | 0.17 | 48.14 | 9.56 | 49.84 | 9.3 | 50.35 | 9.36 |
| 11 -12Å | 24.03 | 0.01 | 17.92 | 0.22 | 17.58 | 0.19 | 18.37 | 0.2 |
| 12 -13Å | 58.13 | 0.87 | 58.59 | 5.38 | 58.76 | 5.38 | 54.58 | 5.63 |
| 13 -14Å | 11.11 | 0.01 | 25.09 | 0.05 | 22.75 | 0.04 | 22.81 | 0.04 |
| >14 Å | 83.65 | 99.9 | 81.38 | 99.7 | 81.36 | 99.7 | 81.44 | 99.63 |

## 6.3 Results and Discussion

**Distance dependence of the statistical potentials.**

To examine the difference between our potential EPAD and the popular DOPE, we plot the potentials as a function of inter-atom distance for two atom pairs, as shown in Figure 17. Figure 17(A) shows the DOPE interaction potential for the atom pair ALA $C_\alpha$ and LEU $C_\alpha$ and also the EPAD interaction potential for this pair in three different positions of protein 1gvp. DOPE has the same energy curve for this atom pair regardless of its sequence positions. In particular, DOPE always has a favorable potential when the distance of this pair of atoms is between 5 and 7Å and has an interaction potential close to 0 when the distance is larger than 8.0Å. By contrast, EPAD has one unique energy curve for this atom pair for each position. The figure legend indicates the corresponding native distances (Å) between atom ALA $C_\alpha$ and atom LEU $C_\alpha$ at the three different sequence positions. For example, the bottom curve in Figure 17(A) visualizes the EPAD interaction potential for the ALA $C_\alpha$ and LEU $C_\alpha$ pair with native distance 8.31Å. This curve shows that when the distance between ALA $C_\alpha$ and LEU $C_\alpha$ is close to the native, their EPAD interaction potential is favorable. In fact, EPAD always has a favorable potential for these three ALA $C_\alpha$ and LEU $C_\alpha$ pairs when their distances are close to the natives.

Figure 17(B) compares the EPAD and DOPE interaction potentials for another atom pair Cys N and Trp O in three different proteins of 1b3a, 1bkr and 1ptq. Similar to Figure 17(A), EPAD has different interaction potentials for the same atom pair in three different proteins while DOPE has the same potential across all proteins. In particular, EPAD has a favorable potential when the distance between Cys N and Trp O is close to the native. Nevertheless, DOPE has a favorable potential when their distance is between 2 and 4Å and a potential close to 0 when the distance is >8.0Å.

In summary, our statistical potential EPAD is significantly different from currently popular potentials such as DOPE and DFIRE. DOPE, DFIRE, RAPDF and RW have more or less similar energy profiles for atom pairs of the same type. The difference between EPAD and any of DOPE, DFIRE, RAPDF and RW is much larger than that among DOPE (Shen and Sali, 2006), DFIRE (Zhou and Zhou, 2002), RAPDF (Samudrala and Moult, 1998) and RW (Zhang and Zhang, 2010).

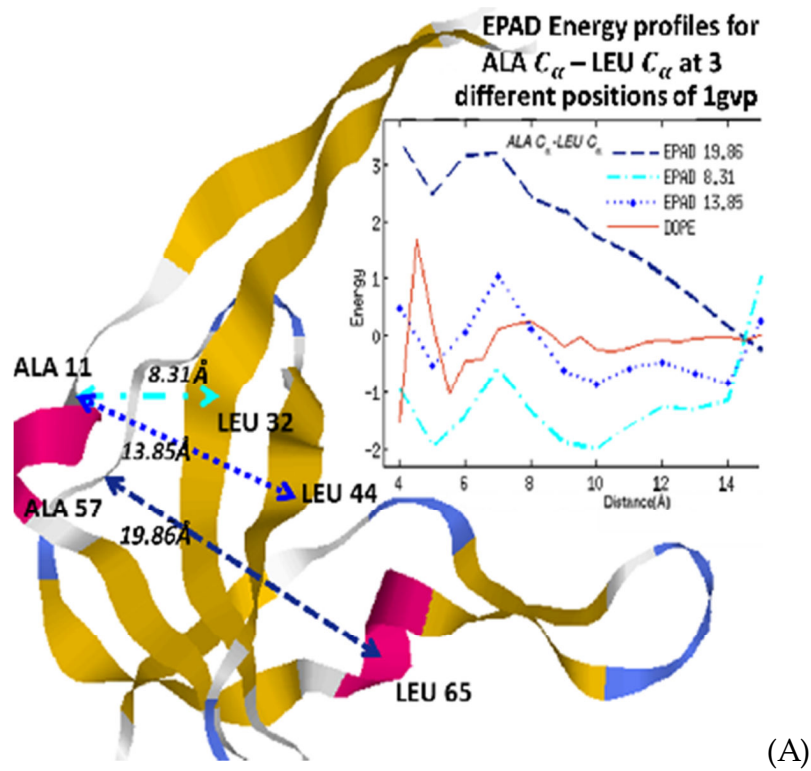

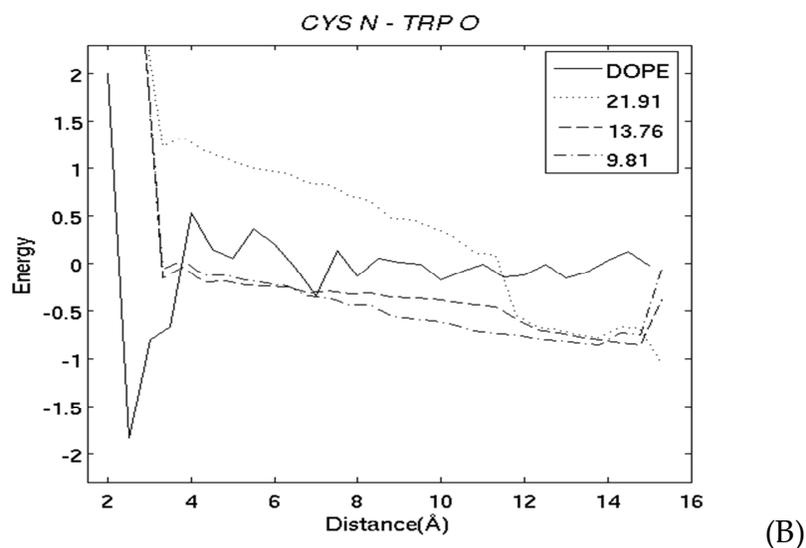

Figure 17 **Distance dependence of DOPE and our potential EPAD.**

(A) The solid curve shows the DOPE interaction potential for atom $C_\alpha$ in ALA and atom $C_\alpha$ in LEU. The other 3 curves show the EPAD potentials for the same atom pair in three different positions of protein 1gvp. The legend shows the native distances of this atom pair in these positions. (B) The curves show the DOPE and EPAD potential for atom N in Cys and atom O in Trp in three different proteins of 1b3a, 1bkr and 1ptq.

## Performance on decoy discrimination.

We test our backbone-based potential EPAD on several decoy sets including the Rosetta set (Qian et al., 2007), the CASP9 models, the I-TASSER dataset (Zhang and Zhang, 2010), the CASP5-8 dataset (Rykunov and Fiser, 2010) and the Decoy 'R' Us (Samudrala and Levitt, 2000) set as well as an in-house large set of template-based models. We evaluate EPAD and several others DOPE, DFIRE, OPUS (Lu et al., 2008) and RW (Zhang and Zhang, 2010) using five performance metrics including the number of correctly-identified natives, the ranking of the native structures, the Z-score of the native energy, the model quality of the first-ranked decoys and the Pearson correlation coefficient (Pearson CC) between the energy and the model quality. The first three metrics evaluate how well a statistical potential can differentiate natives from decoys. The Pearson CC is more important when we want to apply the potentials to folding a protein sequence. We evaluate the model quality of a decoy using the widely-used GDT (Zemla, 2003; Zemla et al., 1999, 2001), which compares a decoy with its native and generates a quality value between 0 and 100. The higher the GDT, the better quality the decoy has.

## Performance on the 2007 Rosetta dataset.

The set contains decoys generated and refined by the popular fragment assembly *ab initio* folding program Rosetta (Qian et al., 2007) for 58 proteins. To evaluate our potential in a more realistic setting, for each protein we use only the 100 low-quality decoys in the set, excluding the high-quality decoys. The average GDT of the best decoys is about 60. As shown in Table 20, our EPAD, which currently considers only backbone atoms, correctly identifies 34 native structures with the lowest Z-score (-2.46), while two full-atom potentials DOPE and DFIRE can identify only 21 natives. EPAD also exceeds DFIRE and DOPE in terms of the average ranking of the native structures (15.70 vs. 23.71 and 21.59). In terms of the average per-target Pearson correlation coefficient (CC) between the energy and GDT, EPAD (-0.42) is significantly better than DOPE (-0.32) and DFIRE (-0.25). EPAD also exceeds RW by all the 5 metrics.
EPAD compares favorably to OPUS-PSP, a full-atom statistical potential. OPUS can correctly identify many more native structures than EPAD, but it has a very low Pearson CC to decoy quality, which indicates that OPUS-PSP may not be good for *ab initio* folding. Since EPAD does not contain side-chain atoms, we simply build a full-atom potential EPAD2 by linearly combining EPAD with the side-chain component in OPUS-PSP (with equal weight). EPAD2 significantly outperforms DOPE, DFIRE, RW and OPUS-PSP by all the 5 metrics. EPAD2 greatly outperforms EPAD in correctly recognizing the native structures, which may imply that side-chain information is very helpful for the identification of the native structures. This trend is also observed on other datasets (e.g., I-TASSER and CASP5-8).

Table 21 compares the performance of several statistical potentials when only $C_\alpha$ atoms are considered in scoring a decoy. Again, EPAD significantly outperforms DOPE, DFIRE and OPUS. EPAD even performs as well as the full-atom potentials DOPE, DFIRE and RW. To excluding the impact of the different datasets used to build EPAD and DOPE, we rebuild a DOPE using the EPAD training data, denoted as MyDope. MyDope performs slightly worse than DOPE, possibly because we do not fine-tune MyDope. However, EPAD performs significantly better than both DOPE and MyDope. This indicates that EPAD outperforms DOPE not due to the training data, but due to the novel methodology.

**Performance on template-based models.**

To examine the performance of EPAD on template-based models, we constructed a large set of 3600 protein pairs, denoted as *InHouse*, with the following properties: 1) any two proteins in a pair share less than 25% sequence identity; 2) all protein classes (i.e., alpha, beta and alpha-beta proteins) are covered by this set; 3) the protein lengths are widely distributed; 4) the structure similarity of two proteins in a pair, measured by TM-score, ranges from 0.5 to 0.8 and is almost uniformly distributed; and 5) within each protein pair, one protein is designated as the target protein and the other as the template protein. Any two target proteins share less than 25% sequence identity. We generated four different target-template alignments for each protein pair using our in-house protein structure alignment program DeepAlign, our two threading programs BoostThreader (Peng and Xu, 2009) and CNFpred and HHpred (Soding, 2005). CNFpred is an improved version of BoostThreader, replacing regression trees in the latter with neural networks and making use of more protein features. Then we used MODELLER (N. Eswar, 2006) to build four different 3D models for each target protein based upon the four alignments, respectively. MODELLER also builds models for the unaligned regions. To remove overlap with the training data, we constructed 4 subsets *Set1*, *Set2*, *Set3* and *Set4* of *InHouse* as follows. *Set1* contains no target proteins sharing >25% sequence identity with the EPAD training proteins. *Set2* is a subset of *Set1*, containing no target proteins sharing >25% sequence identity with the EPAD validation proteins. *Set3* contains no target proteins with a BLAST E-value<0.01 with the EPAD training proteins. *Set4* is a subset of *Set3*, containing no target proteins with a BLAST E-value<0.01 with the EPAD training and proteins. In total, *Set1, Set2, Set3* and *Set4* contain 1139, 331, 965, and 266 protein pairs, respectively.
Table 22 lists the performance of several statistical potentials in identifying the 3D models with the best GDT in the five datasets: *InHouse*, *Set1*, *Set2*, *Set3* and *Set4*. As shown in Table 22, EPAD is able to recognize many more the best template-based models than the others, which are no better than random guess. Further, EPAD has similar performance on the 5 sets, which confirms that our probabilistic neural network (PNN) model is not over-trained. For over 95% of protein pairs in *InHouse*, the 3D models built from the structure alignments have the best GDT. This implies that except EPAD, the other potentials are not able to differentiate structure alignments from threading-generated alignments.

Table 20 Performance of EPAD and several popular full-atom statistical potentials on the Rosetta decoy set.

Numbers in bold indicate the best performance. The per-target Pearson CC is calculated between the energy and GDT and then the average value is reported in the table.

|  | EPAD | DOPE | DFIRE | OPUS | RW | EPAD2 |
|---|---|---|---|---|---|---|
| #natives identified | 34 | 21 | 21 | 39 | 21 | **46** |
| ranking of native | 15.7 | 23.7 | 21.6 | 9.8 | 23.9 | **13.4** |
| first-ranked GDT | 51.6 | 49.7 | 49.4 | 49.7 | 48.5 | **52.4** |
| Pearson CC | **-0.42** | -0.32 | -0.25 | -0.20 | -0.32 | -0.39 |
| Z-score | -2.46 | -1.61 | -1.67 | **-3.27** | -1.51 | **-3.28** |

Table 21 Performance of EPAD and several popular statistical potentials on the Rosetta decoy sets when only $C_\alpha$ atoms are considered.

Numbers in bold indicate the best performance. MyDope is a re-compiled DOPE using the EPAD training data.

|  | EPAD | DOPE | DFIRE | MyDope | OPUS |
|---|---|---|---|---|---|
| #natives identified | **33** | 11 | 12 | 10 | 6 |
| Ranking of native | **15.8** | 18.7 | 30.7 | 21.7 | 55.3 |
| first-ranked GDT | **51.2** | 47.0 | 47.8 | 48.2 | 45.9 |
| Pearson CC | **-0.40** | -0.24 | -0.20 | -0.21 | -0.15 |
| Z-score | **-2.45** | -1.51 | -0.66 | -1.23 | 0.25 |

Table 22 Performance of EPAD and several popular statistical potentials on the template-based models.

Only EPAD is a backbone-based potential while the others are full-atom potentials. In each cell, the numbers in and out parenthesis are the number and percentage of correctly-identified models (i.e., models with the lowest energy and the best GDT). Bold numbers indicate the best performance.

|       | InHouse | Set1 | Set2 | Set3 | Set4 |
|-------|---------|------|------|------|------|
| EPAD  | **1903** (53%) | **617** (54%) | **178** (54%) | **514** (53%) | **143** (54%) |
| DOPE  | 900 (25%) | 288 (25%) | 82 (25%) | 252 (26%) | 74 (28%) |
| DFIRE | 936 (26%) | 286 (25%) | 86 (26%) | 253 (26%) | 74 (28%) |
| OPUS  | 900 (25%) | 289 (25%) | 73 (22%) | 251 (26%) | 69 (26%) |
| RW    | 762 (21%) | 248 (22%) | 68 (21%) | 218 (23%) | 60 (22%) |

## Performance on the CASP9 models.

To further examine the performance of EPAD, we compile a test set from the CASP9 models submitted by the top 18 servers. We exclude the CASP9 targets with many domains since some servers do not place the models of all the domains in a single coordinate system. These 18 servers are BAKER-ROSETTASERVER (Raman et al., 2009), chunk-TASSER (Zhou et al., 2009), chuo-fams (Kanou et al., 2009), CLEF-Server (Shao et al., 2011), FAMSD (Kanou et al., 2009), gws (Joo et al., 2009), HHpredA (Hildebrand et al., 2009), Jiang_Assembly (Hu et al., 2011), MULTICOM-CLUSTER (Tegge et al., 2009), MULTICOM-NOVEL (Tegge et al., 2009), Pcomb (Larsson et al., 2008), Phyre2 (Kelley and Sternberg, 2009), pro-sp3-TASSER (Zhou and Skolnick, 2009), QUARK (Xu et al., 2011), RaptorX (Peng and Xu, 2011), Seok-server (Lee et al., 2010), Zhang-Server (Xu et al., 2011), ZHOU-SPARKS-X (Yang et al., 2011). We do not include the models from RaptorX-MSA, RaptorX-Boost, HHpredB, HHpredC, MULTICOM-REFINE, MULTICOM-CONSTRUCT and Jiang_THREADER since they are not very different from some of the 18 servers. In summary, this CASP9 dataset contains the first models submitted by 18 servers for 92 targets. This set is very challenging for any energy potentials because the models submitted by these top servers have similar quality especially for those not-so-hard targets. The first-ranked models by EPAD, DOPE, DFIRE and RW have an average GDT of 58.6, 55.7, 56.0, and 57.4, respectively. The average Pearson correlation coefficient (between GDT and energy values) for EPAD is -0.364, which is significantly better than DOPE (-0.25), DFIRE (-0.23) and RW (-0.28). Note that RW parameters are fine-tuned using the CASP8 and CASP9 models while EPAD, DOPE and DFIRE are independent of any decoy sets. In addition, EPAD is only a backbone-based potential while the other three are full-atom potentials.

Table 23 shows the performance of EPAD, DOPE, DFIRE and RW with respect to the hardness of the targets, which is judged based upon the average GDT of all the models of this target. We divide the targets into four groups according to the average GDT: <30, 30-50, 50-70, >70. EPAD performs very well across all difficulty levels and has particularly good correlation coefficient for the targets with average GDT less than 30. Even for easy targets EPAD also outperforms the others although it is believed that sequence profiles are not very effective in dealing with easy targets. The only exception is that EPAD has a worse average GDT of the first-ranked models than RW for the targets with average GDT between 30 and 50. This is because RW performs exceptionally well on a single target T0576. The best model identified by RW has GDT 53.3 while EPAD, DOPE and DFIRE can only identify a model with GDT 17.0.

## Performance on the Decoy 'R' Us dataset.

The set is taken from http://dd.compbio.washington.edu/, containing decoys for some very small proteins. In terms of the average rank of the native structures EPAD significantly exceeds the others, but EPAD correctly identifies slightly fewer native structures than DOPE and OPUS_PSP, in part because EPAD does not include side-chain atoms.

The results of DOPE and DFIRE in Table 24 are taken from (Shen and Sali, 2006), that of OPUS_PSP is from (Lu et al., 2008), and that of RW is calculated by the program in (Zhang and Zhang, 2010). As shown in the table, all the energy functions are able to correctly differentiate the native structures from decoys for the proteins in Lattice_ssfit. EPAD identifies more native structures for the proteins in Lmds, while DOPE and OPUS_PSP do so for 4state_reduced and Fisa_casp3. In terms of the average rank of the native structures EPAD significantly exceeds the others, but EPAD correctly identifies slightly fewer native structures than DOPE and OPUS_PSP, in part because EPAD does not contain side-chain information.

## Performance on the I-TASSER dataset.

This set contains decoys for 56 proteins generated by I-TASSER (http://zhanglab.ccmb.med.umich.edu/). The average TMscore of the decoys in this set ranges from 0.346 to 0.678. EPAD outperforms DFIRE and DOPE by 5 measures. EPAD is slightly better than RW in terms of the first-ranked TMscore and the correlation, but slightly worse than RW in terms of the Z-score of the natives. EPAD2 (i.e., the combination of the OPUS-PSP side-chain potential and EPAD) can obtain much better Z-score of the natives although the correlation is slightly decreased. This is consistent with what is observed on the Rosetta set.

As shown in Table 25, our statistical potential EPAD outperforms DFIRE and DOPE no matter which measure is used. EPAD is slightly better than RW in terms of the first-ranked TMscore and the correlation, but slightly worse than RW in terms of the Z-score of the natives. EPAD2 (i.e., the combination of the OPUS-PSP side-chain potential and EPAD) can obtain much better Z-score of the natives although the correlation is slightly decreased. This is consistent with what is observed on the Rosetta set.

Table 23 Performance of statistical potentials with respect to the hardness of the CASP9 targets.

To save space, DOPE and DFIRE are denoted as "DP" and "DF", respectively, and we also omit the negative sign of the correlation coefficient. The first column indicates the hardness of the targets, judged by the average GDT of all the models of the target.

|  | GDT of the first-ranked models ||||Correlation Coefficient||||
|---|---|---|---|---|---|---|---|---|
|  | EPAD | DP | DF | RW | EPAD | DP | DF | RW |
| <30 | **27.4** | 23.1 | 24.1 | 25.8 | **0.44** | 0.28 | 0.23 | 0.33 |
| 30-50 | 42.0 | 40.0 | 40.6 | **42.7** | **0.31** | 0.26 | 0.24 | 0.27 |
| 50-70 | **64.4** | 61.6 | 61.5 | 63.4 | **0.37** | 0.24 | 0.22 | 0.27 |
| >70 | **80.0** | 77.1 | 77.1 | 77.1 | **0.35** | 0.26 | 0.26 | 0.26 |

Table 24 Performance of several statistical potentials on the Decoy 'R' Us dataset.

In each entry of columns 2-6, the numbers in and out parenthesis are the numbers of correctly-identified native structures and the average rankings of the natives, respectively. Numbers in bold indicate the best performance.

| Decoy Set | EPAD | DFIRE | DOPE | OPUS_PSP | RW | #targets |
|---|---|---|---|---|---|---|
| 4state_reduced | 5/1.4 | 6/1.4 | **7/1** | **7/1** | 6/1.4 | 7 |
| Fisa | 3/65.5 | 3/64.3 | 3/94.5 | 3/78.8 | **3**/17.0 | 4 |
| Fisa_casp3 | 2/1.3 | 3/1 | 3/1 | 3/1 | 3/1 | 3 |
| Lattice_ssfit | 8/1 | 8/1 | 8/1 | 8/1 | 8/1 | 8 |
| Lmds | **9**/51.0 | 7/143.9 | 7/101.8 | 8/91.8 | 7/141.6 | 10 |
| All | 27/**24.8** | 27/55.4 | 28/45.6 | **29**/39.1 | 27/47.0 | 32 |

Table 25 Performance of several statistical potentials on the I-TASSER dataset.

CC-TM is the correlation coefficient between energy and the quality, measured by TMscore, of the decoys. Numbers in bold indicate the best performance.

|  | EPAD | DFIRE | DOPE | RW | EPAD2 |
|---|---|---|---|---|---|
| #natives identified | 53 | 47 | 30 | 53 | 53 |
| first-ranked TMscore | **0.579** | 0.558 | 0.560 | 0.569 | 0.564 |
| CC-TM | **-0.532** | -0.492 | -0.317 | -0.500 | -0.512 |
| Z-score of natives | -3.61 | -3.58 | -2.18 | -4.42 | **-5.94** |

## Performance on the CASP5-8 dataset.

EPAD is only second to QMEAN6, RW and RWplus in ranking the best models in the absence of the native structures. When the native structures are included, EPAD does not perform as well as when the native structures are not included. EPAD2 outperforms all the others in terms of the average ranking of the best models in the absence of the native structures or the average ranking of the native structures. EPAD2 also performs very well in terms of the number of correctly-identified models (or native structures). These results may further indicate that side-chain information is needed for the accurate identification of the native structures.

Table 26 shows the performance of a variety of potentials on the CASP5-8 models selected by Rykunov and Fiser (Rykunov and Fiser, 2010). As shown in this table, EPAD2, which is the combination of our potential EPAD and the side-chain potential in OPUS, outperforms all the others in terms of the average ranking of the best models in the absence of the native structures or the average ranking of the native structures. EPAD2 also performs very well in terms of the number of correctly-identified models (or native structures). EPAD is only second to QMEAN6, RW and RWplus in ranking the best models in the absence of the native structures. When the native structures are included, EPAD does not perform as well as when the native structures are not included. All these results may indicate that the side-chain component helps a lot in recognizing the native structures.

## Is our probabilistic neural network (PNN) model over-trained?

Our PNN model has 60,000-70,000 parameters to be trained. A natural question to ask is if our PNN model is biased towards some specific patterns in the training data? Can our PNN model be generalized well to proteins of novel folds or sequences? According to our experimental results on contact prediction (see "**Window size and the number of neurons in the hidden layers**" section in **6.2 Methods and Experimental procedures**), our PNN model is not over-trained. In this experiment, we used a training set built before CASP8 started, which are unlikely to have similar folds and sequence profiles with our test set (i.e., the CASP8 and CASP9 free modeling targets). Experimental results indicate that our PNN method compares favorably to the best CASP8 and CASP9 server predictors, which implies that our PNN model is not biased towards the training data.

Table 26 Performance of a variety of potentials on the selected CASP5-8 models.

Note: The results of EPAD and EPAD2 are shown in bold and the results of all the other potentials are taken from (Zhang and Zhang, 2010).

a. The average rank of lowest energy decoy according to GDT score (over 143 decoy sets) in the absence of native structures.

b. The number of sets when the best model was ranked as first, in the absence of native structures.

c. The average rank of the lowest energy decoy in GDT when native structures are present.

d. The number of sets when the best model was ranked as first when native structures are present.

e. Expected random values were generated by picking a wining model from the decoy sets randomly. Average values over 1000 random trials are shown in the last row.

| Scoring function | models only | | native structures included | |
|---|---|---|---|---|
| | Average [a] | Ranking [b] | Average [c] | Ranking [d] |
| **EPAD2** | **2.79** | **69** | **1.24** | **127** |
| QMEAN6 | 2.87 | 85 | 1.71 | 113 |
| RWplus | 2.97 | 57 | 1.78 | 106 |
| RW | 3.08 | 51 | 1.71 | 110 |
| **EPAD** | **3.29** | **62** | **2.71** | **74** |
| QMEANall_atom | 3.59 | 74 | 1.71 | 119 |
| QMEANSSE_agree | 3.74 | 62 | 3.72 | 39 |
| QMEANACC_agree | 4.04 | 40 | 3.78 | 48 |
| RF_CB_SRS_OD | 4.16 | 61 | 2.08 | 110 |
| RF_CB_OD | 4.62 | 62 | 2.00 | 111 |
| RF_HA_SRS | 4.65 | 49 | 1.38 | 137 |
| RF_CB_SRS | 4.72 | 56 | 2.18 | 114 |
| OPUS_CA | 4.72 | 79 | 5.13 | 55 |
| VSCOREcombined | 4.79 | 53 | 2.2 | 117 |
| QMEAN-pairwise | 4.8 | 54 | 3.15 | 85 |
| Rosetta | 5.01 | 57 | 4.09 | 68 |
| Dong-pair | 5.01 | 58 | 6.32 | 4 |
| RF_CB | 5.06 | 52 | 2.46 | 106 |
| VSCORE-pair | 5.08 | 54 | 1.85 | 128 |
| PROSAcombined | 5.11 | 57 | 3.38 | 87 |
| OPUS_PSP | 5.39 | 54 | 2.99 | 118 |
| RF_HA | 5.44 | 62 | 2.78 | 112 |
| DOPE | 5.77 | 54 | 3.27 | 95 |
| DFIRE | 6.03 | 50 | 5.69 | 33 |
| PROSA-pair | 6.03 | 56 | 3.54 | 95 |
| QMEAN-torsion | 6.71 | 45 | 3.24 | 114 |
| Shortle2006 | 6.85 | 35 | 1.79 | 129 |
| Liang_geometric | 6.88 | 44 | 2.48 | 114 |
| QMEANsolvation | 7.32 | 33 | 6.27 | 54 |
| Shortle2005 | 7.73 | 42 | 3.39 | 109 |
| Floudas-CM | 7.75 | 38 | 7.05 | 42 |
| Floudas-Ca | 7.79 | 33 | 8.36 | 10 |
| NAMD_1000 | 8.06 | 24 | 4.96 | 78 |
| Melo-ANOLEA | 9.62 | 19 | 5.19 | 86 |
| PC2CA | 9.75 | 19 | 5.06 | 85 |
| Melo-NL | 9.99 | 14 | 5.85 | 80 |
| NAMD_1 | 11.91 | 5 | 10.98 | 24 |
| Random [e] | 9.72 | 13.9 | 10.1 | 8.3 |

Note that our PNN model for statistical potential uses exactly the same architecture (2 hidden layers with 100 and 40 hidden neurons, respectively) as our PNN model for contact prediction. Considering that much more training data (~73 millions of residue pairs) is used for the derivation of our statistical potential than for contact prediction, it is less likely that our PNN model for statistical potential is biased towards some specific patterns in the training set. The result in Table 22 further confirms this. We use the 25% sequence identify or an E-value 0.01 as the cutoff to exclude proteins in InHouse with similar sequences to the training set and generate two subsets Set2 and Set4. Even if Set2 (Set4) contains some sequence profiles similar to the training set, the similarity between the whole InHouse set and the training set is still much larger than that between Set2 (Set4) and the training set, but the performance on the whole InHouse set is even slightly worse than that on Set2 (Set4).

## Folding Results with EPAD in the blind CASP 10 experiment

One of the most important functions of a good potential is its support in the folding process. We tested EPAD using CNF-Folder (see Chapter 5) by replacing the DOPE term in the energy function (see Chapter 4 for more details) with EPAD. The Energy guiding the folding simulation is the linear combination of 4 potential terms: $U_{all} = w_1 U_{EPAD} + w_2 U_{DOPE} + w_3 U_{BMK} + w_4 U_{ESP}$. The weight parameters ($w_i$) are learned using grid search through a series of folding experiments over a set of proteins containing all the targets from Table 18. Base on the observation from the experiments, our CNF-Folder adopts the strategy to apply DOPE with more weight on those low Neff targets or targets with short sequence length (≤65), while keeping $w_2$ to zero for other targets.

We have tested this new CNF-Folder with the new energy function on CASP9 Free modeling targets. Table 27 compares the performance of CNF-Folder and Rosetta on CASP9 Free modeling targets. We have downloaded the server prediction of Rosetta from CASP9 official web site (http://predictioncenter.org/download_area/CASP9/server_predictions/) and calculated the GDT scores using TM-score (Zhang and Skolnick, 2005b). In the previous CASP experiments, the GDT scores of our best clusters have been almost always within the top 7% decoys. Hence, for simplicity, we do not generate clusters as we normally do in the CASP blind experiments, but instead use the average GDT score of the top 10% decoys for each target in the comparison. As can be seen from Table 27, with the EPAD term, the new CNF-Folder performs favorably comparing to the most up to date Rosetta server.

Backed with this encouraging result, we have participated in the CASP 10 blind experiment using the new CNF-Folder-EPAD program. The server we registered for CASP 10 FM targets is named RaptorX-roll.

There are 16 template free modeling targets in CASP 10 experiment. We used a standalone classification program to pick out the FM targets from where 10 targets have overlap with CASP 10 official classifications. Table 28 compares the performance of our FM method RaptorX-roll on the FM targets with RaptorX-ZY, one of the top threading servers in CASP 10. It can be seen that RaptorX-roll yields better decoys on most of the FM targets than those from RaptorX-ZY.

Figure 18 visualizes RaptorX-roll's predictions on targets T0734 and T0740. The native structures are plotted in blue and decoy structures are in pink color. The Percentage of Residues comparison figures are also listed beside each target structure where RaptorX-roll's distributions are marked in pink color. The more a decoy's distribution curve stands to the right hand side, the higher quality a decoy structure is. T0734 had two domains and T0740 has one domain. It can be seen that RaptorX-roll gave much better prediction than any of the other predictors. One thing to notice is that we have given better predictions on most of the alpha targets while mediocre on the beta proteins. It emphasizes the need of more focus on the hydrogen bond potential in the future works.

Table 27 Folding comparison with Rosetta (GDT) on CASP9 FM targets

Note: The Rosetta data is obtained from CASP9 official web site. Rosetta's Best denotes the GDT score of the best decoy submitted by Rosetta. We list the average GDT of top 10% decoys generated by CNF-Folder as well as the best GDT. For convenience, those low Neff targets are listed below the other targets.

| Target | Type | Neff | Len | Rosetta Best | CNF-Folder Top 10% | CNF-Folder Best |
|---|---|---|---|---|---|---|
| T0531 | 1α3β | 2.9 | *61* | 36.2 | *38.6* | 46.2 |
| T0534_1 | 5α | 5.6 | 166 | *30.2* | 19.1 | 20.5 |
| T0534_2 | 5α2β | 7 | 175 | 22.4 | *34.8* | 40.3 |
| T0544_1 | 3α | 4.4 | *61* | 68.0 | *70.1* | 77.9 |
| T0544_2 | 3α | 4.3 | 78 | 46.6 | *55.0* | 60.5 |
| T0547_3 | 1α8β | 5.4 | 148 | 40.8 | *68.7* | 74.7 |
| T0550_1 | 11β | 8 | 159 | 14.7 | *20.0* | 24.1 |
| T0553 | 4α | 4.6 | 141 | 28.4 | *36.1* | 44.7 |
| T0571_1 | 15β | 7.4 | 168 | *37.0* | 14.7 | 18.8 |
| T0571_2 | 8β | 3.1 | 133 | *35.2* | 22.8 | 29.6 |
| T0578 | 3α6β | 3.3 | 163 | 23.3 | *23.5* | 29.0 |
| T0604_1 | 2α6β | 5.8 | 84 | 34.1 | *48.6* | 56.7 |
| T0604_3 | 3α10β | 4.8 | 205 | 13.2 | *21.7* | 26.9 |
| T0616 | 2α3β | 5.2 | 93 | 34.8 | *35.5* | 40.9 |
| T0618 | 6α2β | 4.5 | 170 | *35.4* | 31.2 | 38.8 |
| T0624 | 7β | 2.5 | 69 | *44.6* | 35.9 | 41.7 |
| T0637 | 7α | 7.3 | 135 | *39.6* | 30.1 | 35.6 |
| T0561 | 6α | *1.7* | 161 | 31.5 | *34.4* | 41.2 |
| T0581 | 3α6β | *1.8* | 112 | *67.9* | 33.8 | 42.9 |
| T0608_1 | 5α | *1.6* | 89 | 19.1 | *36.3* | 46.9 |
| T0621 | 2α10β | *1.2* | 169 | *18.3* | 15.9 | 19.7 |
| T0639 | 4α | *1.5* | 123 | *46.8* | 30.9 | 35.9 |

Table 28 Comparison with threading method RaptorX-ZY on CASP10 FM targets

Note: RaptorX-ZY is one of the top threading servers in CASP 10. The scores listed are GDT score of the best decoys submitted by each server. The CaspBest column lists the GDT scores of the best decoys out of all the CASP10 servers.

| Target | Type | Neff | length | RaptorX-roll | RaptorX-ZY | **CaspBest** |
|---|---|---|---|---|---|---|
| T0653 | $40\beta$ | 12.2 | 414 | 4.2 | **21.9** | 31.8 |
| T0658D1 | $9\beta$ | 3.6 | 185 | **16.1** | 13.1 | 24.7 |
| T0666 | $6\alpha$ | 4.3 | 195 | **26.3** | 23.2 | 33.8 |
| T0684D2 | $6\alpha4\beta$ | 1.2 | 167 | **17.4** | 16.5 | 21.3 |
| T0693D1 | $4\alpha4\beta$ | 4.6 | 100 | **25.3** | 24.0 | 35.0 |
| T0734 | $8\alpha2\beta$ | 5.5 | 214 | **20.3** | 13.9 | 21.0 |
| T0735D2 | $4\alpha$ | 4.6 | 117 | **39.8** | 31.0 | 41.5 |
| T0737D1 | $5\alpha2\beta$ | 1 | 116 | 27.4 | 27.8 | 36.8 |
| T0740 | $7\alpha2\beta$ | 1 | 165 | **38.9** | 25.3 | 38.9 |
| T0741 | $14\beta$ | 1 | 181 | 13.6 | **15.2** | 17.2 |

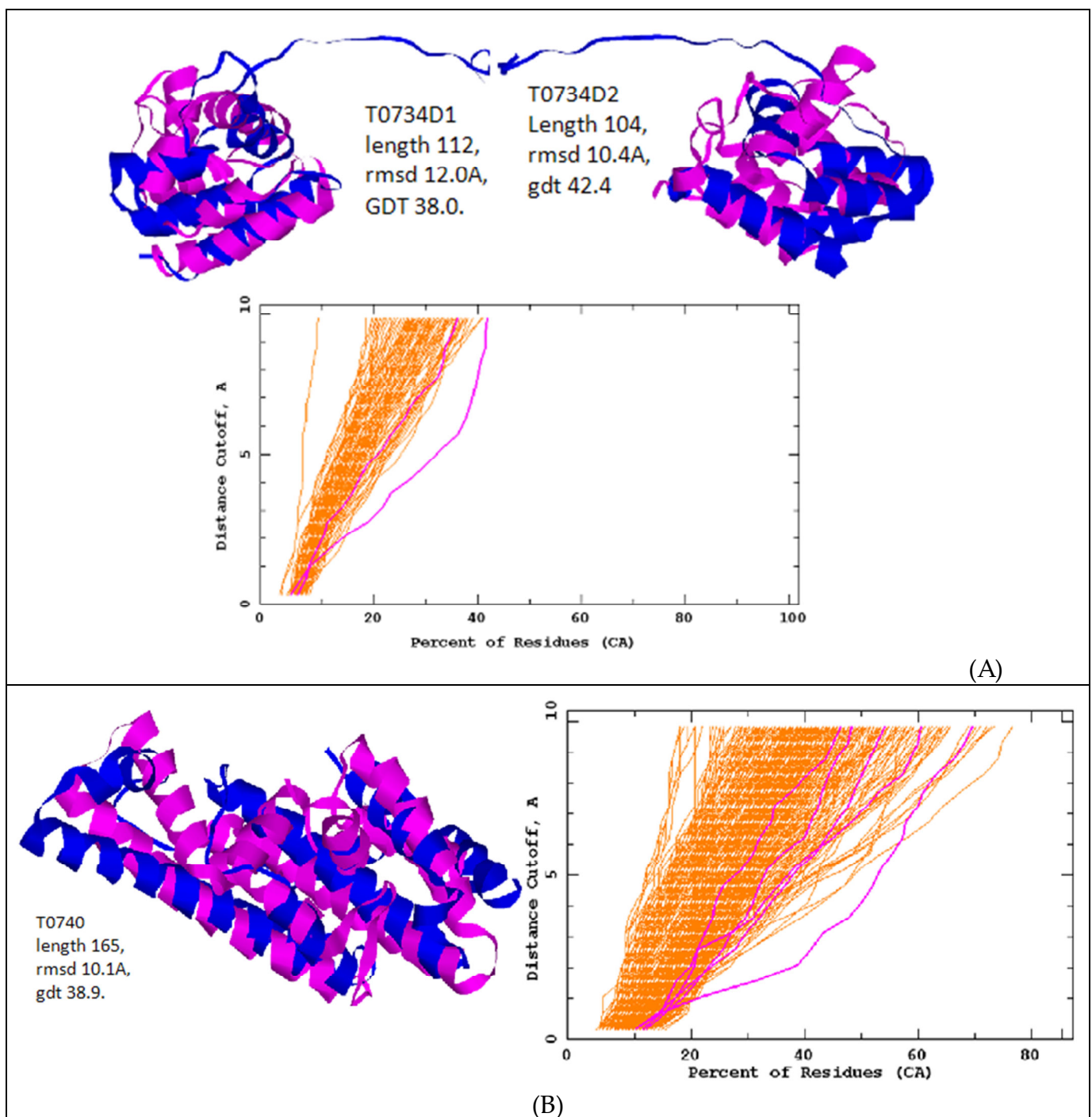

Figure 18 . **RaptorX-roll's Result on targets T0734 (A) and T0740 (B).**

The native structures are plotted in blue and decoy structures are in pink color. In the Percentage of Residues comparison figures, RaptorX-roll's distributions are marked in pink color. The more a decoy's distribution curve stands to the right hand side, the higher quality a decoy structure is.

## 6.4 Conclusion

This chapter has presented a novel protein-specific and position-specific knowledge-based statistical potential EPAD for protein structure and functional study. EPAD is unique in that it may have different energy profiles for two atoms of given types, depending on the protein under consideration and the sequence profile contexts of the residues containing them, while other potentials have the same energy profile for a given atom pair across all proteins. We achieved this by parameterizing EPAD using evolutionary information and radius of gyration of the protein under consideration in addition to atom types, which enables us to obtain a much more accurate statistical potential.

We also made a novel technical contribution to estimating the observed atomic interacting probability by introducing a probabilistic neural network to calculate the inter-atom distance probability distribution from sequence profiles and the radius of gyration. This is very different from the simple counting method widely used to derive the position-independent statistical potentials such as DOPE and DFIRE. The simple counting method does not work for our potential simply because there is not enough number of solved protein structures in PDB for reliable counting of sequence profile contexts.

Experimental results indicated that EPAD significantly outperforms several popular higher-resolution full-atom potentials in several decoy discrimination tests even if only backbone atoms are considered in EPAD. If we combine EPAD with the side-chain component in OPUS-PSP, we can achieve much better decoy discrimination performance especially in the presence of native structures. As opposed to the RW potential and many others, EPAD is not trained by any decoys, so in principal it is not restricted to any decoy generation method. Currently EPAD uses only 1Å resolution for the $C_\alpha - C_\alpha$ distance discretization. We will further improve EPAD by using a 0.5Å resolution, but this will take a very long time to train a neural network model for accurate estimation of the extremely unbalanced distance probability distribution.

# Chapter 7. Conclusion and Future works

## 7.1 Conclusions

The thesis is aimed to solve the template-free protein folding problem by tackling two important components: efficient sampling in vast conformation space, and design of knowledge-based potentials with high accuracy.

The CRF-Samplers are proposed to sample structures from the continuous local dihedral angles space by modeling the lower and higher order conditional dependency between neighboring dihedral angles given the primary sequence information. A framework combining the Conditional Random Fields and the energy function is introduced to guide the local conformation sampling using long range constraints with the energy function. In order to model the complex nonlinear relationship between the sequence profile and the local dihedral angle distribution, we further built the CNF-Folder by applying a novel machine learning model Conditional Neural Fields which utilizes the structural graphical model with the neural network. CRF-Samplers and CNF-Folder perform very well in CASP8 and CASP9.

On the other track, we have designed a novel pairwise distance statistical potential (EPAD) to capture the dependency of the energy profile on the positions of the interacting amino acids as well as the types of those amino acids, opposing the common assumption that this energy profile depends only on the types of amino acids. EPAD has also been successfully applied in the CASP 10 Free Modeling experiment with CNF-Folder, especially outstanding on some uncommon structured targets.

## 7.2 Future Works

### Conformation Sampling

Although the primary purpose of the free modeling folders including the CRF-Samplers and CNF-Folder is for template-free protein folding, they can be applied to other important applications. One direct example is to refine template-based models. Some preliminary results on CASP blind experiments can be found in (Xu et al., 2009) and (Ma et al., 2013). The method can also be applied to protein loop modeling, model refinement, and even RNA tertiary structure prediction (Wang and Xu, 2011).

The free modeling folders have some immediate limitations to relax.

By extracting distance constraints from template-based models, the conformational space of a target is dramatically reduced and thus we can afford to search this reduced space, which may search conformational space more thoroughly and lead to better prediction accuracy.

They can also be further extended to model the long-range hydrogen bonding effect to significantly enhance the performance on beta proteins.

## Statistical Potentials

We may extend our statistical potential as follows.

Currently EPAD considers only backbone atoms and is also orientation-independent. We can extend it to side-chain atoms and also make it orientation-dependent.

Second, in estimating the distance probability distribution of two positions, we use only sequence profile contexts relevant to only these two positions. We shall also use information in the sequence segment connecting the two residues, which contains important information in determining the relative orientation of the two residues.

Thirdly, we may also estimate the distance probability distribution more accurately by adding some physical constraints. For example, given any three atoms in a protein, their pairwise distances must satisfy the triangle inequality.

Further, for any three residues which are close to one another along the primary sequence, their $C_\alpha$ distances are also subject to the restriction of local atomic interaction. If we assume that there is a contact between two residues if their $C_\alpha$ or $C_\beta$ atoms are within 8Å, then the number of contacts for any given residue is limited by a constant (~13) due to geometric restraint. By enforcing these constraints, we shall be able to estimate the inter-atom distance probability distribution much more accurately and thus, design a much better statistical potential.

Other two important potentials to probe are hydrogen bonding potential and environmental potential.

We have some preliminary results on the latter category that a contact number potential is designed and experimental results support the hypothesis that the wrapping of each residue by surrounding amino acids is guided by maximizing the local static electric field. We will keep working on it in the near future.

Kim, H., and Park, H. (2003). Protein secondary structure prediction based on an improved support vector machines approach. Protein Eng *16*, 553-560.

Kolodny, R., Koehl, P., Guibas, L., and Levitt, M. (2002). Small libraries of protein fragments model native protein structures accurately. J Mol Biol *323*, 297-307.

Kortemme, T., and Baker, D. (2002). A simple physical model for binding energy hot spots in protein-protein complexes. P Natl Acad Sci USA *99*, 14116-14121.

Krogh, A., Brown, M., Mian, I.S., Sjolander, K., and Haussler, D. (1994). Hidden Markov-Models in Computational Biology - Applications to Protein Modeling. J Mol Biol *235*, 1501-1531.

Labesse, G., Colloc'h, N., Pothier, J., and Mornon, J.P. (1997). P-SEA: a new efficient assignment of secondary structure from C alpha trace of proteins. Comput Appl Biosci *13*, 291-295.

Lafferty, J., McCallum, A., and Pereira, F.C.N. (2001). Conditional random fields: Probabilistic models for segmenting and labeling sequence data.

Larsson, P., Wallner, B., Lindahl, E., and Elofsson, A. (2008). Using multiple templates to improve quality of homology models in automated homology modeling. Protein Science *17*, 990-1002.

Laurie, A.T.R., and Jackson, R.M. (2005). Q-SiteFinder: an energy-based method for the prediction of protein-ligand binding sites. Bioinformatics *21*, 1908-1916.

Lee, J., Lee, D., Park, H., Coutsias, E.A., and Seok, C. (2010). Protein loop modeling by using fragment assembly and analytical loop closure. Proteins-Structure Function and Bioinformatics *78*, 3428-3436.

Levitt, M. (1976). A simplified representation of protein conformations for rapid simulation of protein folding. J Mol Biol *104*, 59-107.

Levitt, M. (1992). Accurate Modeling of Protein Conformation by Automatic Segment Matching. J Mol Biol *226*, 507-533.

Li, S.C., Bu, D., Xu, J., and Li, M. (2008). Fragment-HMM: a new approach to protein structure prediction. Protein Science *17*, 1925-1934.

Li, X., and Liang, J. (2007). Knowledge-based energy functions for computational studies of proteins. In Computational methods for protein structure prediction and modeling (Springer), pp. 71-123.

Liu, D., and Nocedal, J. (1989). On the Limited Memory Bfgs Method for Large-Scale Optimization. Math Program *45*, 503-528.

Lu, H., and Skolnick, J. (2001). A distance-dependent atomic knowledge-based potential for improved protein structure selection. Proteins *44*, 223-232.

Lu, M., Dousis, A., and Ma, J. (2008). OPUS-PSP: An orientation-dependent statistical all-atom potential derived from side-chain packing. J Mol Biol *376*, 288-301.

Ma, J., Wang, S., Zhao, F., and Xu, J. (2013). Protein threading using context-specific alignment potential. Bioinformatics *29*, i257-i265.

Maiorov, V.N., and Crippen, G.M. (1992). Contact Potential That Recognizes the Correct Folding of Globular-Proteins. J Mol Biol *227*, 876-888.

Shao, M.F., Wang, S., Wang, C., Yuan, X.Y., Li, S.C., Zheng, W.M., and Bu, D.B. (2011). Incorporating Ab Initio energy into threading approaches for protein structure prediction. BMC Bioinformatics *12*.

Shen, M., and Sali, A. (2006). Statistical potential for assessment and prediction of protein structures. Protein Science *15*, 2507-2524.

Shi, S., Pei, J., Sadreyev, R., Kinch, L., Majumdar, I., Tong, J., Cheng, H., Kim, B.-H., and Grishin, N. (2009). Analysis of casp8 targets, predictions and assessment methods, . Database *2009:bap003*.

Simon, I., Glasser, L., and Scheraga, H.A. (1991). Calculation of Protein Conformation as an Assembly of Stable Overlapping Segments: Application to Bovine Pancreatic Trypsin Inhibitor. Proceedings of National Academy Sciences, USA *88*, 3661-3665.

Simons, K.T., Kooperberg, C., Huang, E., and Baker, D. (1997). Assembly of protein tertiary structures from fragments with similar local sequences using simulated annealing and Bayesian scoring functions. J Mol Biol *268*, 209-225.

Simons, K.T., Ruczinski, I., Kooperberg, C., Fox, B.A., Bystroff, C., and Baker, D. (1999). Improved recognition of native-like protein structures using a combination of sequence-dependent and sequence-independent features of proteins. Proteins *34*, 82-95.

Singh, H., Hnizdo, V., and Demchuk, E. (2002). Probabilistic model for two dependent circular variables. Biometrika *89*, 719-723.

Sippl, M. (1993). Recognition of errors in three-dimensional structures of proteins. Proteins: Structure, Function, and Bioinformatics *17*, 355-362.

Sippl, M.J. (1990). Calculation of Conformational Ensembles from Potentials of Mean Force - an Approach to the Knowledge-Based Prediction of Local Structures in Globular-Proteins. J Mol Biol *213*, 859-883.

Sippl, M.J., and Weitckus, S. (1992). Detection of native-like models for amino acid sequences of unknown three-dimensional structure in a data base of known protein conformations. Proteins-Structure Function and Bioinformatics *13*, 258-271.

Skolnick, J. (2006). In quest of an empirical potential for protein structure prediction. Curr Opin Struc Biol *16*, 166-171.

Skolnick, J., Kihara, D., and Zhang, Y. (2004). Development and large scale benchmark testing of the PROSPECTOR_3 threading algorithm. Proteins *56*, 502-518.

Skolnick, J., Kolinski, A., and Ortiz, A. (2000). Derivation of protein-specific pair potentials based on weak sequence fragment similarity. Proteins *38*, 3-16.

Soding, J. (2005). Protein homology detection by HMM-HMM comparison. Bioinformatics *21*, 951-960.

Specht, D.F. (1990). Probabilistic Neural Networks. Neural Networks *3*, 109-118.

Swendsen, R.H., and Wang, J.S. (1986). Replica Monte-Carlo Simulation of Spin-Glasses. Physical Review Letters *57*, 2607-2609.

Tanaka, S., and Scheraga, H.A. (1976). Medium- and long-range interaction parameters between amino acids for predicting three-dimensional structures of proteins. Macromolecules *9*, 945-950.